\documentclass[lettersize,journal, twoside]{IEEEtran}
\usepackage{amsmath,amsfonts}
\usepackage{array}
\usepackage{textcomp}
\usepackage{stfloats}
\usepackage{url}
\usepackage{verbatim}
\usepackage{graphicx}
\usepackage{cite}
\usepackage{tikz}
\usepackage{censor}
\def\censorcolor{gray!50} \let\svcensorrule\censorrule \renewcommand\censorrule[1]{ \textcolor{\censorcolor}{\svcensorrule{#1}}}

\usepackage{makecell, longtable}
\usepackage{stfloats}

\usepackage{caption}
\usepackage{booktabs}
\usetikzlibrary{shadows,arrows.meta,shapes,positioning, fit, calc}
\usepackage{comment}
\usepackage[edges]{forest}
\usepackage{amssymb}
\usepackage{pifont}
\usepackage{multirow}
\usepackage{colortbl}
\usepackage{xcolor}

\usepackage[ruled,vlined,linesnumbered]{algorithm2e}

\SetCommentSty{mycommfont}
\SetKwComment{Comment}{$\triangleright$\ }{}
\let\oldnl\nl%
\newcommand{\nonl}{\renewcommand{\nl}{\let\nl\oldnl}}%

\usepackage[flushleft]{threeparttable}
\newcommand{\cmark}{\ding{51}}
\newcommand{\xmark}{\ding{55}}

\DeclareMathOperator*{\argmin}{arg\,min}
\newcolumntype{C}[1]{>{\centering\arraybackslash}p{#1}}
\newcolumntype{P}[1]{>{\raggedright\arraybackslash\noindent}p{#1}}

\definecolor{hidden-draw}{RGB}{205, 44, 36}
\definecolor{lightgray}{gray}{0.9}

\usepackage{soul}

\definecolor{color1}{HTML}{E0F7FA}
\definecolor{color2}{HTML}{80DEEA}
\definecolor{color3}{HTML}{00ACC1}

\definecolor{color1}{gray}{0.9}
\definecolor{color2}{gray}{0.8}
\definecolor{color3}{gray}{0.7}

\usepackage{tcolorbox}

\usepackage[hidelinks]{hyperref}
\hypersetup{
   linkcolor=red,
   breaklinks=true,   
   colorlinks=true,   
   citecolor=red, 
}

\setstcolor{red}
\hypersetup{colorlinks,linkcolor={blue},citecolor={blue},urlcolor={blue}}  

\begin{document}

\title{\huge{Toward Generalist Neural Motion Planners for Robotic Manipulators: Challenges and Opportunities}}
\author{Davood Soleymanzadeh$^{1}$, Ivan Lopez-Sanchez$^{2}$, Hao Su$^{2}$, Yunzhu Li$^{3}$, Xiao Liang$^{4}$, and Minghui Zheng$^{*,1}$ \\
\thanks{This work was supported in part by U.S. National Science Foundation under
Grant 2422826 and Grant 2524088, and in part by National Institutes of Health
under Grant 1R01EB035404-01 and Grant R01NS141171.}
\thanks{$^{1}$Davood Soleymanzadeh and Minghui Zheng are with the J. Mike Walker '66 Department of Mechanical Engineering, Texas A\&M University, College Station, TX, USA (\tt\footnotesize e-mail: davoodso@tamu.edu; mhzheng@tamu.edu).}
\thanks{$^{2}$Ivan Lopez-Sanchez and Hao Su are with the Lab of Biomechatronics and Intelligent Robotics, Department of Biomedical Engineering, Tandon School of Engineering, New York University, NY, USA (\tt\footnotesize e-mail: hao.su@nyu.edu).}
\thanks{$^{3}$Yunzhu Li is with the Department of Computer Science, Columbia University, NY, USA (\tt\footnotesize e-mail: yunzhu.li@columbia.edu).}
\thanks{$^{4}$Xiao Liang is with the Zachry Department of Civil and Environmental Engineering, Texas A\&M University, College Station, TX, USA (\tt\footnotesize e-mail: xliang@tamu.edu).}
\thanks{$^*$ Corresponding Author.}}
\maketitle

\begin{abstract} \label{sec:abstract}
State-of-the-art generalist manipulation policies have enabled the deployment of robotic manipulators in unstructured human environments. However, these frameworks struggle in cluttered environments primarily because they utilize auxiliary modules for low-level motion planning and control. Motion planning remains challenging due to the high dimensionality of the robot's configuration space and the presence of workspace obstacles. Neural motion planners have enhanced motion planning efficiency by offering fast inference and effectively handling the inherent multi-modality of the motion planning problem. Despite such benefits, current neural motion planners often struggle to generalize to unseen, out-of-distribution planning settings. This paper reviews and analyzes the state-of-the-art neural motion planners, highlighting both their benefits and limitations. It also outlines a path toward establishing generalist neural motion planners capable of handling domain-specific challenges. For a list of the reviewed papers, please refer to \href{https://davoodsz.github.io/planning-manip-survey.github.io/}{ https://davoodsz.github.io/planning-manip-survey.github.io/}.

\end{abstract}

\section{Introduction} \label{sec:intro}
\begin{figure}[t] 
\centering
\includegraphics[scale=0.4]{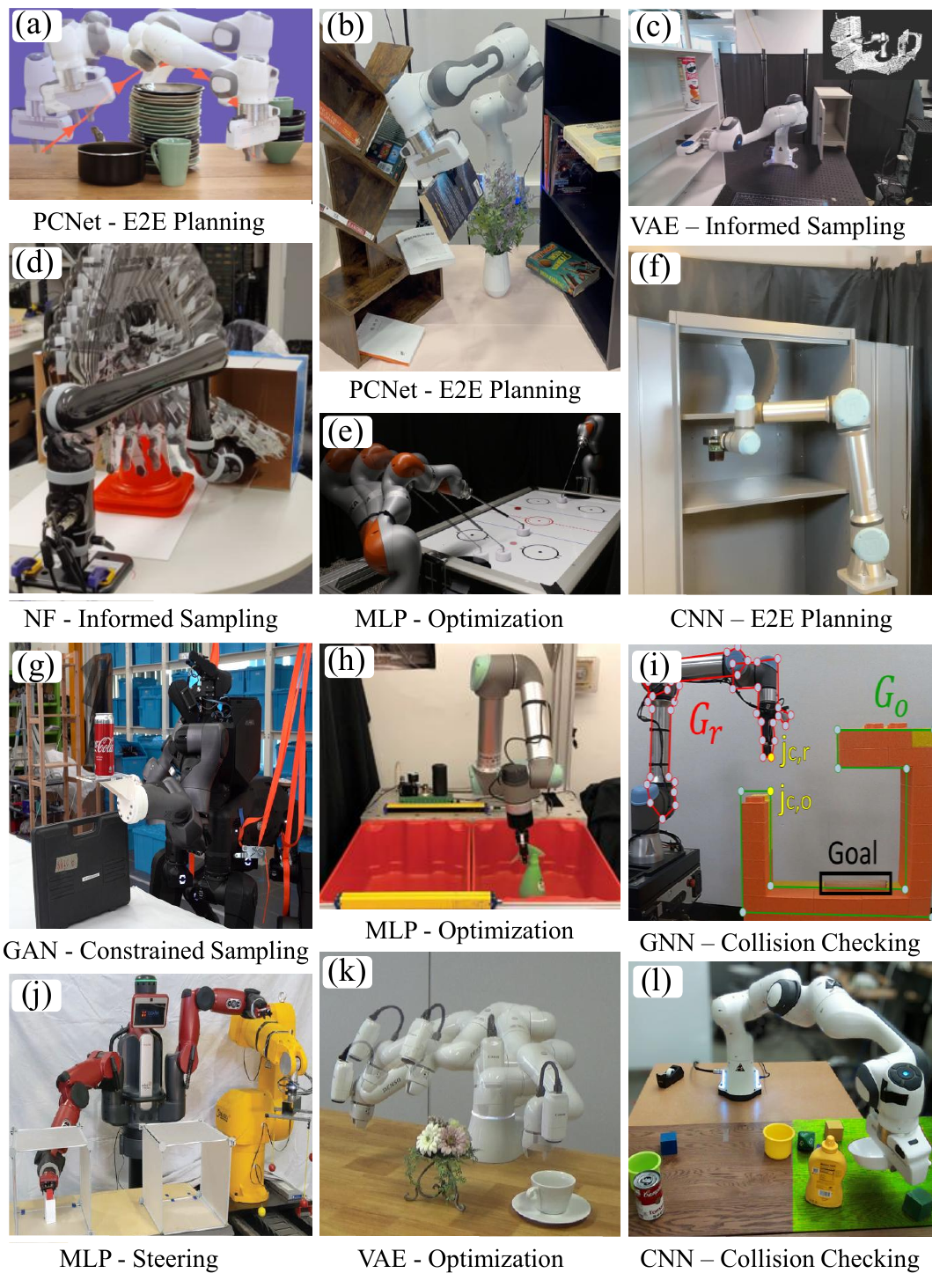}
\caption{Deep learning for robotic manipulator motion planning. (a, b) Point cloud networks (PCNets) for End-to-end (E2E) planning \cite{fishman2023motion, dalal2024neural} (Section \ref{subsec:DL-endtoend}). (c) Variational autoencoders (VAEs) \cite{johnson2023learning}, and (d) normalizing flows \cite{lai2021plannerflows} for informed sampling (Section \ref{subsec:DL-constrainedsampling}). (e) Multilayer perceptions (MLPs) for trajectory optimization \cite{ichnowski2020deep} (Section \ref{subsec:DL-optimization}). (f) Convolutional neural networks (CNNs) for E2E planning \cite{ni2024physics} (Section \ref{subsec:DL-endtoend}). (g) Generative adversarial networks (GANs) for constraint manifold learning \cite{acar2021approximating} (Section \ref{subsec:DL-constrainedsampling}). (h) MLPs for trajectory optimization \cite{kicki2023fast} (Section \ref{subsec:DL-optimization}). (i) Graph neural networks (GNNs) for collision checking \cite{kim2022graphdistnet} (Section \ref{subsec:DL-collisionchecking}). (j) MLPs for steering \cite{chiang2021fast} (Section \ref{subsubsec: DL-sampling-based-steering}). (k) VAEs for trajectory optimization \cite{osa2022motion} (Section \ref{subsec:DL-optimization}). (l) CNNs for collision checking \cite{danielczuk2021object} (Section \ref{subsec:DL-collisionchecking}).}
\label{fig_1-opening_image}
\end{figure}

\IEEEPARstart{T}{he} emergence of robotics for humans and society \cite{proia2021control}, has positioned robotic manipulators as fundamental elements across various applications, such as the evolution of medical services \cite{moustris2011evolution}, and manufacturing settings \cite{tamizi2023review, lasi2014industry}. A ubiquitous challenge in the deployment of robotic manipulators is planning motions for seamless operation in dynamic, cluttered environments \cite{tamizi2023review, orthey2023sampling, McMahon_2022}.

\begin{figure*}[htbp] 
\centering
\includegraphics[width=\textwidth]{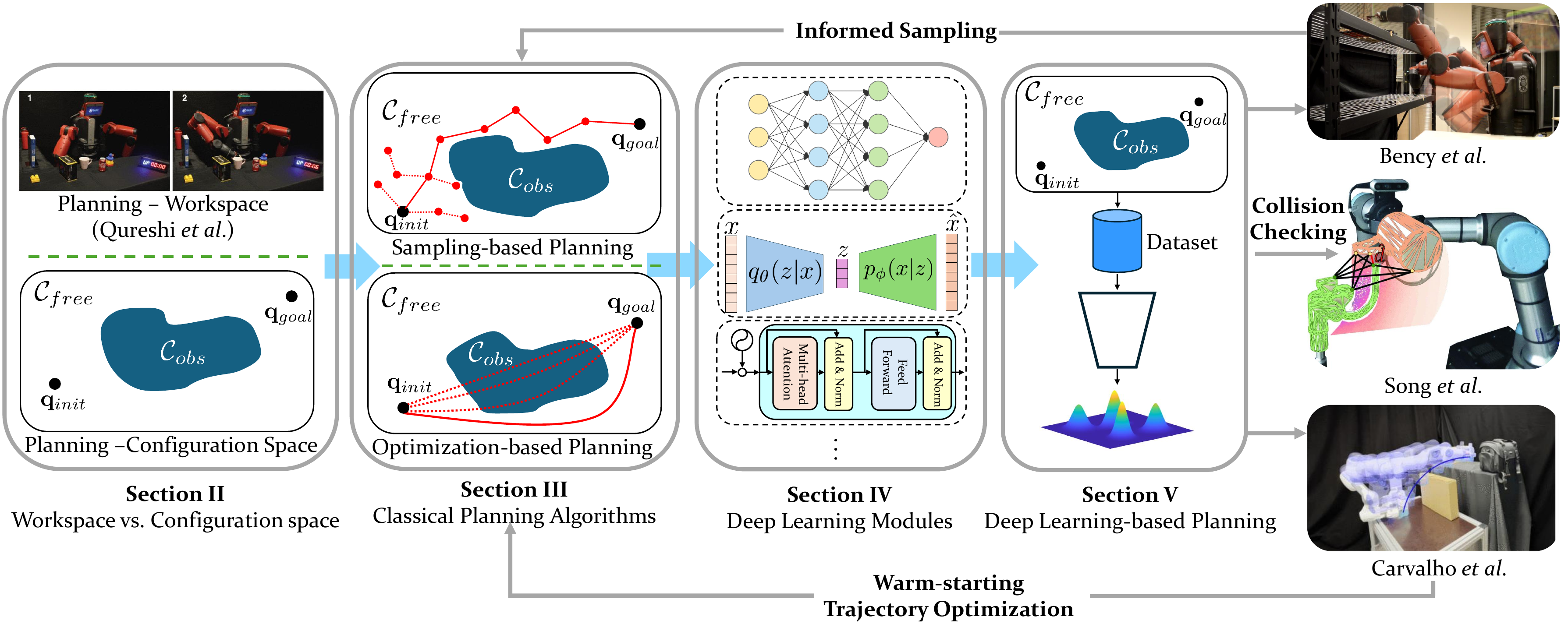}
\caption{
Overview of this survey paper. Figures are adopted from Carvalho \textit{et al.} \cite{carvalho2023motion}, Qureshi \textit{et al.} \cite{qureshi2019motion}, Bency \textit{et al.} \cite{bency2019neural}, and Song \textit{et al.} \cite{song2023graph}}
\label{fig:main}
\end{figure*}

Planning for a robotic system involves finding a feasible collision-free path between a pre-defined start and goal within its configuration space. For robotic manipulators, the process is difficult due to high dimensionality and expensive collision checking. Robotic manipulators that operate in dynamic everyday environments need fast planning algorithms. Classical planning algorithms are often slow, computationally expensive, and require extensive knowledge about the manipulator's environment and its operational capabilities \cite{noroozi2023conventional, wang2021survey}.
Deep learning methods have been leveraged to address these challenges by encoding the underlying similarities between planning problems for efficient planning, thanks to their fast inference, and ease of implementation \cite{pierson2017deep}.

\begin{figure}[htbp] 
\centering
\vspace{-5pt}
\includegraphics[width=0.45\textwidth]{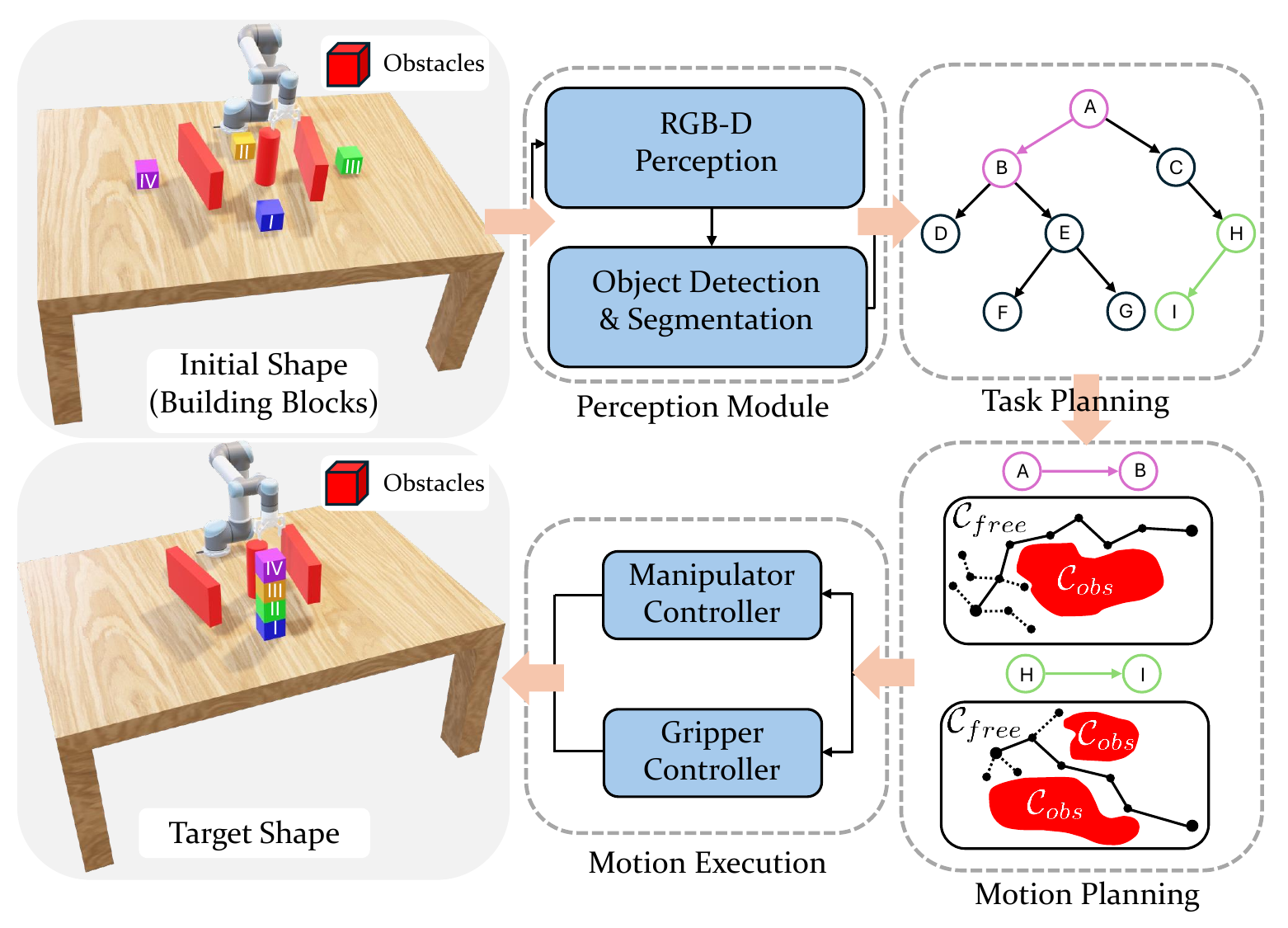}
\caption{An example of a robotic manipulator's autonomy stack.}
\label{fig_2- autonomy-stack}
\end{figure}

Deep learning techniques have significantly enhanced various components of the manipulator's autonomy stack \cite{firoozi2023foundation}. Figure \ref{fig_2- autonomy-stack} demonstrates various components of the manipulator's autonomy stack. Deep learning methods for vision \cite{dehghani2023scaling} and language processing \cite{achiam2023gpt} have been utilized to convert high-dimensional environmental sensory data (e.g., images, videos, voices, texts) into low-dimensional interpretable embeddings \cite{iosifidis2022deep}. Furthermore, deep learning methods have been leveraged for task specifications, transforming high-level input instructions to low-level sequences of actions that robotic manipulators can execute while imposing temporal specifications \cite{ahn2022can}. Additionally, deep learning methods have been efficiently utilized to solve the inverse kinematics problem for redundant robotic manipulators \cite{kang2021rcik}.

Deep learning methods have also enhanced the motion planning sub-stack, either by improving specific components of classical motion planners or by functioning as end-to-end planners \cite{McMahon_2022,wang2021survey}, as illustrated in Figure \ref{fig_1-opening_image}. Despite vast utilization, there are challenges associated with the utilization of deep learning methods for manipulator motion planning. These challenges include:

\begin{table*}
\begin{threeparttable}
\centering
\caption{Overview of existing survey papers on robotic motion planning, their scope, focus, and limitations compared to this paper.}
\label{tab: existing-surveys}
\small
\begin{tabular}{>{\raggedright\arraybackslash}p{0.1\textwidth} >
{\raggedright\arraybackslash}p{0.15\textwidth} >{\raggedright\arraybackslash}p{0.35\textwidth} >{\raggedright\arraybackslash}p{0.25\textwidth} >{\raggedright\arraybackslash}p{0.25\textwidth}}
\arrayrulecolor{gray!80}
		\hline
  \rowcolor{gray!20}
		\textbf{Reference} & \textbf{Robot Types} & \textbf{Scope and Focus} & \textbf{Notes} \\\hline
		\textbf{Wang \textit{et al.}, 2021 \cite{wang2021survey}} & {Not specific to manipulators}&Machine learning for improving/replacing classical planning algorithms & Up to 2021 \\\hline
            \textbf{McMahon \textit{et al.}, 2022 \cite{McMahon_2022}} & {Not specific to manipulators}& 
			Machine learning for improving classical sampling-based planning algorithms & {Not covering other planning algorithms}  \\\hline
            \textbf{Noroozi \textit{et al.}, 2023 \cite{noroozi2023conventional}}&{Not specific to manipulators}&{All types of planning algorithms (including both classical and learning-based)}& {Not specific to deep learning}\\\hline
\textbf{Tamizi \textit{et al.}, 2023 \cite{tamizi2023review}}&{Robotic manipulators}&{All types of planning algorithms (classical, learning-based)}&{Not specific to deep learning}\\
\hline
\textbf{Carvalho \textit{et al.}, 2025 \cite{de2025data}}&{Not specific to manipulators}&{Data-driven planning algorithms}&{Not specific to robotic manipulators} \\
\bottomrule
\rowcolor{blue!10}\textbf{Ours}&Robotic manipulators& Focused on deep learning for improving/replacing classical planning algorithms&Since 2018 \\
\hline
\end{tabular}
\begin{tablenotes}
      \item \textbf{Note}: The main criteria for selecting recent survey papers were twofold. First, the survey needed to contain planning for robotic manipulators. Second, it had to consider recent papers that utilize deep learning methods for planning.
    \end{tablenotes}
\end{threeparttable}
\end{table*}

\begin{itemize}
    \item \textbf{Data scarcity}: The first step in training and utilizing deep learning methods is dataset collection. Although foundation models \cite{yang2023foundation} are pre-trained on internet-scale datasets, they are not suitable for the motion planning sub-stack. In the literature, almost all the deep learning models utilized for planning are trained on limited datasets specifically gathered for that specific problem \cite{wang2021survey}. Creating internet-scale datasets for low-level planning poses significant challenges, including the need for high-fidelity physics simulators for large-scale data generation.

    \item \textbf{Generalization}: The main drawback of deep learning tools is their excellent performance on in-distribution problems during inference time, while struggling to generalize to out-of-distribution problems. This limitation is particularly pronounced in planning due to the inherent discontinuities in the planning problem \cite{farber2003topological}. Small changes in the planning problem (workspace) can lead to significant changes in configuration space that were not present in the training dataset \cite{chamzas2021learning}. Addressing these challenges is crucial for the efficient utilization of deep learning methods in motion planning.

    \item \textbf{Real-time applications}: Although more complex deep learning methods and deeper networks may excel at encoding complex similarities within planning problems, the relatively slow inference time of these models limits their deployment in real-world dynamic environments \cite{ratliff2018riemannian}. Further research is required to reduce the inference time of these planners for real-time deployment.

    \item \textbf{Safety guarantees}: It is challenging to analytically ensure the safety and stability of deep learning models. Additional considerations and constraints should be included to guarantee the required safety criteria \cite{rana2020learning}. Rigorously checking for the safety and stability of these methods is challenging and requires further exploration.
\end{itemize}

In this survey paper, we explored the state-of-the-art literature on the utilization of deep learning methods in robotic manipulator motion planning. We identified current challenges and provided future perspectives and research directions to address these challenges accordingly. Table \ref{tab: existing-surveys} lists related survey papers on robotic motion planning. In comparison with these papers, our focus is to provide the promises and limitations of using deep learning methods to enhance classical motion planning algorithms for robotic manipulators.

Our criteria for paper selection are as follows. (1) We focused on the-state-of-the-art literatures that utilize \textit{deep learning} for \textit{robotic manipulator planning} since \textit{2018}; (2) We focused solely on papers that apply \textit{deep learning} to enhance \textit{global planning algorithms}. (3) We \textit{excluded} papers that utilize deep learning for task planning within task and motion planning (TAMP) scenarios and papers that utilize deep learning for low-level motion control of robotic manipulators. (4) We also surveyed \textit{classical planning algorithms}-related research papers to identify their components that can be enhanced by deep learning methods.

The key contributions of this survey paper are as follows:

\begin{itemize}
    \item \textbf{Robotic manipulator motion planning:} We review the state-of-the-art literature that has utilized deep learning for robotic manipulator planning. Robotic manipulators are increasingly deployed within critical applications (e.g., healthcare, re-manufacturing, and agriculture), which necessitate safe and efficient motion planning algorithms. However, motion planning for robotic manipulators remains challenging due to their high DOF and the complexity of real-world environments.
    
    \item \textbf{Systematic mapping from deep learning frameworks to motion planning algorithmic primitives:} We provide a systematic mapping from various deep learning architectures (e.g., convolutional neural networks, deep generative models, large language models) to core algorithmic primitives of classical motion planning algorithms (e.g., sampling and steering primitives of sampling-based planning algorithms).

    \item \textbf{Road to generalist neural motion planners:} We outline a path toward generalist neural motion planners capable of end-to-end planning for robotic manipulators. We summarize the progress that has been made in this direction, identify how far the research community has advanced, and highlight key considerations necessary to achieve this goal. Particularly, we emphasize the need for standardized benchmarks, large-scale planning datasets, explicit handling of safety constraints, generalization to out-of-distribution scenarios, and robustness to planning uncertainties for reliable deployment within unstructured real-world environments. Additionally, we discuss how large-scale foundation models can be established and leveraged to facilitate traversing this path.
\end{itemize}

Given the recent advances in computational power and the emergence of new deep-learning methods, we believe it is essential to revisit the recent applications of these methods in motion planning for robotic manipulators. This paper can provide a unified, comprehensive overview of the challenges and promises of utilizing deep learning methods for robotic manipulator motion planning.

This survey paper is organized as follows. Section \ref{sec:planning_def} defines the planning preliminaries for robotic manipulators, explaining foundational concepts essential to the planning problem. Section \ref{sec:planning_algortihms} reviews classical planning algorithms widely used for robotic manipulators, summarizing their strengths and limitations. Section \ref{sec:DL-modules} introduces various deep learning methods and their potential application for robotic manipulator motion planning. Section \ref{sec:dl-based_planning} reviews the state-of-the-art literature on neural motion planners for robotic manipulators. Section \ref{sec:perspectives} outlines the current challenges and presents future perspectives associated with the application of neural motion planners. Section \ref{sec: domainspecific} provides domain-specific challenges and necessary considerations for effective deployment of neural motion planners. Finally, section \ref{sec: conclusion} summarizes our findings and conclusions based on the reviewed works. Figure \ref{fig:main} illustrates the structure of the survey paper.

\section{Robotic Manipulators Planning Preliminaries} \label{sec:planning_def}
In this section, we introduce commonly used terminologies and planning problems for robotic manipulators. Table \ref{tab: terminology} provides robotic manipulators' planning terminologies and their definitions used in this paper, and Table \ref{tab: planningnotation} summarizes the mathematical notations used throughout this section.

\setlength{\tabcolsep}{3pt}
\begin{table*}[htbp]
\centering
\caption{Robotic manipulator planning terminologies, and their definitions used in this manuscript.}
\label{tab: terminology}
\resizebox{\textwidth}{!}{%
\begin{tabular}
{>{\raggedright\arraybackslash}p{0.14\textwidth}>{\raggedright\arraybackslash}p{0.86\textwidth}} 
\arrayrulecolor{gray!80}
\hline
\rowcolor{gray!20} \textbf{Term}&\textbf{Definition}\\
\hline 
Configuration space & An N-dimensional space spanned by robotic manipulator joint values, where $N$ is the number of DOF. \\
\hline
Workspace& The 3-dimensional space in which the robotic manipulator's end-effector operates.\\
\hline
Planning constraints& Include geometric constraints; kinematic constraints; and dynamic constraints.\\
\hline
Global planning& Planning algorithms that first plan a path from a start to a goal configuration, and then execute it. \\
\bottomrule
\end{tabular}}
\end{table*}
\setlength{\tabcolsep}{3pt}
\begin{table}[htbp]
\begin{center}
\caption{Mathematical notations and definitions for motion planning of robotic manipulators utilized throughout this paper.}
\label{tab: planningnotation}
\resizebox{0.49\textwidth}{!}{%
\begin{tabular}
{>{\raggedright\arraybackslash}p{0.05\textwidth}>{\raggedright\arraybackslash}p{0.2\textwidth}>{\raggedright\arraybackslash}p{0.05\textwidth}>{\raggedright\arraybackslash}p{0.2\textwidth}}
\arrayrulecolor{gray!80}
\hline
\rowcolor{gray!20} \textbf{Symbol}&\textbf{Definition}&\textbf{Symbol}&\textbf{Definition} \\
\hline
$\mathcal{C}$&Configuration space&$\mathcal{C}_{\text{free}}$& Free configuration space \\
$\mathcal{C}_{\text{obs}}$&Obstacle configuration space&$\mathcal{X}$&Workspace \\
$\mathcal{X}_{\text{free}}$&Free workspace&$\mathcal{X}_{\text{obs}}$& Obstacle workspace \\
$\mathbf{q}$&Configuration&$\mathbf{x}$&Workspace pose \\
$f(.)$&Forward kinematics&$\mathbf{q}_{init}$&Initial configuration \\ 
$\mathbf{q}_{goal}$&Goal configuration&$\sigma$&Path \\
$c(.)$&Planning cost&$\mathbf{\tau}$&Trajectory \\
\bottomrule
\end{tabular}}
\end{center}
\end{table}

\subsection{Terminologies}
\begin{figure}[htbp] 
\centering
\includegraphics[scale=0.35]{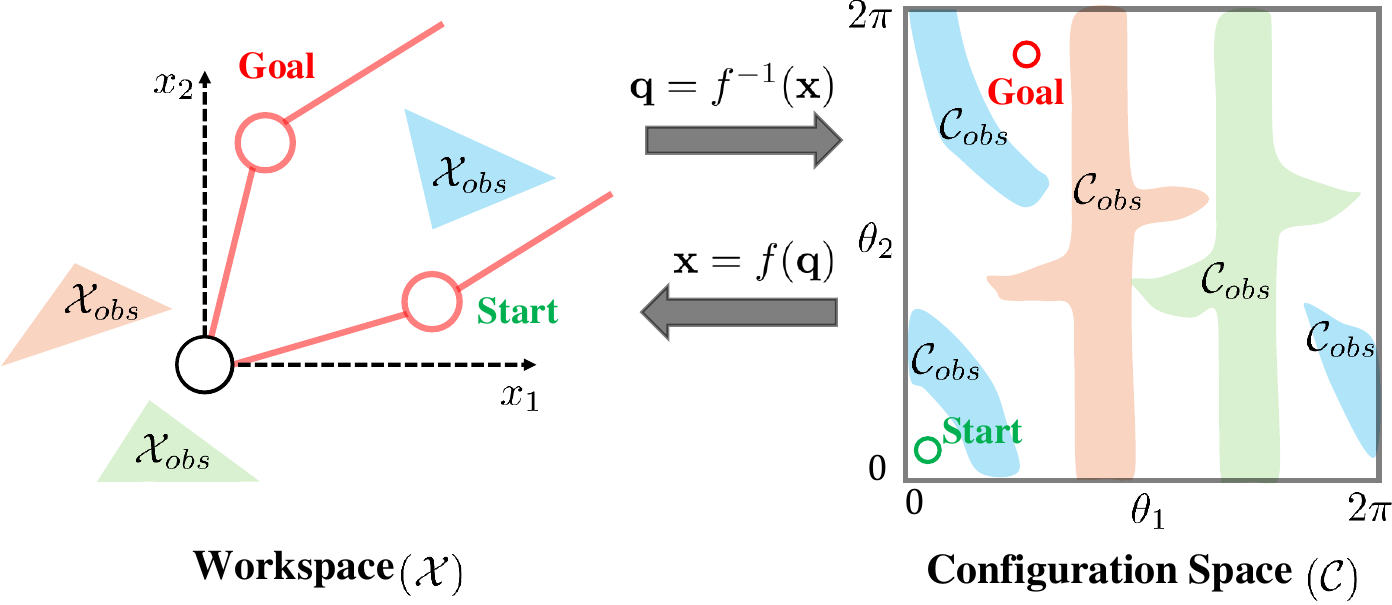}
\caption{An example of a 2-DOF planar manipulator: workspace vs. configuration space\cite{lynch2017modern}.}
\label{fig_3- planning-def}
\end{figure}

Common terms and terminologies related to planning for robotic manipulators \cite{sandakalum2022motion} are as follows. Figure \ref{fig_3- planning-def} illustrates these terms for a 2-DOF robotic manipulator.

\vspace{0.2cm}
\noindent
\textbf{Configuration Space:} The \textit{configuration} of a robotic manipulator is defined by its joint values, represented as $\mathbf{q}=(q_1, ..., q_N)$ where $N$ is the number of DOF of the manipulator \cite{malhan2022generation}. The \textit{configuration space} ($\mathcal{C}$) is an $N$-dimensional space spanned by these joint values, which describe the manipulator's configuration. Using configuration space simplifies the planning problem, as the manipulator is represented by a point (instead of bodies and volumes), and obstacles are represented as forbidden regions ($\mathcal{C}_{obs}$) in this space. The planning problem is then reduced to moving the point representing the manipulator's configuration from the initial configuration to the goal configuration while avoiding the forbidden regions (i.e., staying within the free space $\mathcal{C}_{free} = \mathcal{C}\backslash \mathcal{C}_{obs}$). However, computing these forbidden regions is nontrivial: inverse kinematics for high-DOF (especially redundant) manipulators generally lacks closed-form solutions except for special kinematic structures, and the exact boundary separating $\mathcal{C}_{\text{obs}}$ from $\mathcal{C}_{\text{free}}$ is typically high-dimensional and nonconvex, with no simple closed-form representation~\cite{das2020learning}.

\vspace{0.2cm}
\noindent
\textbf{Workspace:} The workspace $\mathcal{X}$ is the space in which a manipulator's end-effector operates. In this space, the end-effector position is specified by Cartesian coordinates in $\mathbb{R}^3$, while its orientation is represented by the rotation group $SO(3)$ ($\mathcal{X} \in \mathbb{R}^3 \times SO(3)$).

\vspace{0.2cm}
\noindent
\textbf{Forward \& Inverse Kinematics:} These operations enable mapping between the joint configuration and the Cartesian pose of all geometries attached to the manipulator. \textit{Forward kinematics} ($f$) maps a configuration ($\mathbf{q}$) from the configuration space to a workspace pose ($\mathbf{x}$), i.e., $\mathbf{x} = f(\mathbf{q})$. Conversely, \textit{inverse kinematics} maps a pose ($\mathbf{x}$) from the workspace ($\mathcal{X}$) back to a configuration ($\mathbf{q}$) in the configuration space ($\mathcal{C}$). i.e., $\mathbf{q} = f^{-1}(\mathbf{x})$.

\subsection{Planning Definitions}
Various approaches that have been utilized to solve the planning problem are as follows. Refer to Table \ref{tab: planningdef} for an overview of the characteristics of these approaches. This Table highlights key characteristics and differences among various planning methods for robotic manipulators.

\setlength{\tabcolsep}{3pt}
\begin{table}[htbp]
\begin{center}
\caption{Three main approaches to solve the planning problem for robotic manipulators, their distinguishing characteristics.}
\label{tab: planningdef}
\resizebox{0.49\textwidth}{!}{%
\begin{tabular}
{@{}P{0.1\textwidth}C{0.1\textwidth}C{0.1\textwidth}C{0.1\textwidth}C{0.1\textwidth}}
\arrayrulecolor{gray!80}
\hline
\rowcolor{gray!20} {}&\textbf{Geometric \newline constraints}&\textbf{Kinematics \newline constraints}&\textbf{Dynamic \newline constraints} & \textbf{Velocity \newline evolution} \\
\hline 
\textbf{Path \newline planning}&\cmark&\xmark&\xmark&\xmark \\
\hline
\textbf{Motion \newline planning}&\cmark&\cmark&\cmark&\xmark \\
\hline
\textbf{Trajectory \newline planning}&\cmark&\cmark&\cmark&\cmark \\
\bottomrule
\end{tabular}}
\begin{minipage}{0.49\textwidth}
\smallskip
\footnotesize
\textbf{Note}: In this manuscript, we use the term motion planning to broadly describe low-level planning within the robotic manipulation stack. Low-level planning can be further categorized into \textit{path planning}, \textit{motion planning}, and \textit{trajectory planning}, each with distinct meanings, although these terms are often used interchangeably in the literature.
\end{minipage}
\end{center}
\end{table}

\vspace{0.2cm}
\noindent
\textbf{Path Planning:} The path planning problem can be defined as $M = \{\mathbf{q}_{init}, \mathbf{q}_{goal}, \mathcal{C}_{free}\}$ where $\mathbf{q}_{init}$, and $\mathbf{q}_{goal}$ represent initial and goal configurations, respectively. The objective of path planning is to find a feasible path connecting the start and the goal configurations that lies entirely within the free space. The primary goal is to find a feasible path $\sigma = [\mathbf{q}_1, ..., \mathbf{q}_t, ..., \mathbf{q}_T]$ such that: 
\begin{equation} \label{ppdef}
\begin{aligned}
    \sigma(0) &= \mathbf{q}_{\text{init}}, \\
    \sigma(t) &\in \mathcal{C}_{\text{free}}, \\
    \sigma(T) &= \mathbf{q}_{\text{goal}},
\end{aligned}
\end{equation}
where the primary constraints for path planning are \textit{geometric constraints}. Therefore, path planning provides a geometric description of the robotic manipulator's movement by generating collision-free waypoints between the start and goal configurations.

\vspace{0.2cm}
\noindent
\textbf{Motion Planning:} Path planning, as defined in Eq. \ref{ppdef}, focuses solely on geometric constraints. However,  for effective planning, other types of constraints must also be considered, including \textit{kinematic constraints}, \textit{dynamic constraints}, and the robotic manipulator's evolution over time \cite{sandakalum2022motion}. The motion planning problem incorporates all these constraints.

To address the motion planning problem, a set of costs $c_i$ are considered. These costs include all constraints - geometric, kinematic, and dynamic - as soft constraints (i.e., costs to be optimized). The motion planning problem can be defined as follows:
\begin{equation}\label{motionpp}
\begin{aligned}
    \mathbf{\sigma}^* &= \argmin_{\mathbf{\sigma}} \sum_i \lambda_i c_i(\mathbf{\sigma}),
\end{aligned}
\end{equation}
where $\lambda_i > 0$ is a hyperparameter weight assigned to each cost function $c_i(\sigma)$, and $\mathbf{\sigma}$ and $\mathbf{\sigma}^*$  are the planned and optimal motion, respectively. Therefore, motion planning generates a collision-free path between the start and goal configuration while satisfying kinematics and dynamics constraints. 

\vspace{0.2cm}
\noindent
\textbf{Trajectory Planning:} In the literature, the motion planning problem and the trajectory planning problem are often treated as synonyms.
However, from a technical perspective, trajectory planning differs in that it also considers the time evolution of velocity as part of the trajectory \cite{sandakalum2022motion}. Let $\mathbf{s} = [\mathbf{q}^T, \dot{\mathbf{q}}^T]^T \in \mathbb{R}^{2d}$ represent the state of the robotic manipulator, where $\mathbf{q}$ is the robotic manipulator configuration, and $\dot{\mathbf{q}}$ is the velocity of robotic manipulator. A Trajectory $\mathbf{\tau} = [\mathbf{s_1}, ..., \mathbf{s}_t, ..., \mathbf{s}_T]$ is defined as a sequence of states over a time horizon $T$. The trajectory planning problem can be defined as follows \cite{urain2022learning}:
\begin{equation}\label{trajpp}
\begin{aligned}
    \mathbf{\tau}^* &= \argmin_{\mathbf{\tau}} \sum_i \lambda_i c_i(\mathbf{\tau}).
\end{aligned}
\end{equation}

The main difference between the motion planning problem (Eq. \ref{motionpp}) and the trajectory planning problem (Eq. \ref{trajpp}) lies in the inclusion of the robotic manipulator velocity evolution into the problem formulation. Therefore, trajectory planning generates a collision-free path between the start and goal configuration while satisfying kinematics and dynamics constraints, and includes the evolution of robotic manipulator velocity. 

\subsection{Metrics} \label{subsec:planning_def-metrics}
The common metrics to evaluate the performance of planning algorithms for robotic manipulators are: \textit{planning time}, \textit{planning cost}, and \textit{success rate}.

\begin{itemize}
    \item \textit{Planning Time}: Planning time ($T$) is the average planning time the planner takes to find a solution.
    \item \textit{Planning Cost}: Planning cost ($C$) refers to the average length of the planned paths in the configuration space or the workspace.
    \item \textit{Success Rate}: Success rate ($S$) represents the percentage of successfully planned paths.
\end{itemize}
\section{Classical Planning Algorithms} \label{sec:planning_algortihms}
In this section, we introduce classical motion planning algorithms for robotic manipulators. Over the years, a diverse array of algorithms has been developed to address the planning problem. These algorithms are broadly categorized into two groups: sampling-based, and optimization-based algorithms.

\subsection{Sampling-based Planning Algorithms} \label{subsec: classic-sampling-based}
Sampling-based planning algorithms utilize random sampling to create a tree or roadmap within the configuration space \cite{zhang2025motion}, and are broadly categorized into unconstrained and constrained algorithms.

\setlength{\tabcolsep}{3pt}
\begin{table*}[htbp]
\begin{center}
\caption{The two main categories of sampling-based planning algorithms for motion planning of robotic manipulators, their advanced variants, algorithmic primitives, and limitations.}
\label{tab: sbmpclassic}
\begin{tabular}
{>{\raggedright\arraybackslash}p{0.1\textwidth}>{\raggedright\arraybackslash}p{0.08\textwidth}>{\raggedright\arraybackslash}p{0.3\textwidth}>{\raggedright\arraybackslash}p{0.15\textwidth}>{\raggedright\arraybackslash}p{0.3\textwidth}}
\arrayrulecolor{gray!80}
\hline
\rowcolor{gray!20} \textbf{Category}&\textbf{Basic Algorithm}&\textbf{Variants}&\textbf{Algorithmic Primitives}&\textbf{Limitations} \\
\toprule
\textbf{Graph-based (multi-query)}&\scriptsize{PRM \cite{kavraki1996probabilistic}}&\scriptsize{LazyPRM \cite{bohlin2000path}, PRM* \cite{karaman2011sampling}, LazyPRM* \cite{karaman2011sampling}, SPARS \cite{dobson2013sparse}, Improved SPARS \cite{dobson2013improving}.}& $\bullet$ Graph Construction \newline $\bullet$ Path Finding& $\bullet$ Computationally complex graph construction. \\
\midrule
\textbf{Tree-based (single-query)}&\scriptsize{RRT \cite{lavalle1998rapidly}}&\scriptsize{Connect-RRT \cite{kuffner2000rrt}, RRT* \cite{karaman2011sampling}, LBT-RRT \cite{salzman2016asymptotically}, Stable-sparse RRT \cite{li2016asymptotically}, Transition-based RRT \cite{jaillet2010sampling}, Lazy-RRG* \cite{hauser2015lazy}, Quotient-space RRT (QRRT) \cite{orthey2019rapidly}, Quotient-space RRT* (QRRT*) \cite{orthey2020multilevel}, RRT\# \cite{arslan2012role}, RRTX \cite{otte2015rrt}, P-RRT* \cite{qureshi2013adaptive, qureshi2016potential}, Informed-RRT* \cite{gammell2014informed}, FMT* \cite{janson2015fast}, BFMT* \cite{starek2015asymptotically}, Hierarchical FMT* \cite{reid2020sampling}, BIT* \cite{gammell2020batch}, ABIT* \cite{strub2020advanced}, AIT* \cite{strub2020adaptively}.}&$\bullet$ Sampling \newline $\bullet$ Local Planning \newline $\bullet$ Collision Checking& $\bullet$ Struggle to scale up to high-dimensional configuration spaces. \newline $\bullet$ Non-smooth paths \newline $\bullet$ Low convergence rate \newline $\bullet$ Hardly applicable to dynamic environments \newline $\bullet$ Inefficient sampling primitive\\
\bottomrule
\end{tabular}
\end{center}
\end{table*}

\subsubsection{Unconstrained Planning}
Sampling-based motion planners can be divided into two main categories: multi-query probabilistic roadmaps (graph-based planners) \cite{kavraki1996probabilistic, karaman2011sampling, bohlin2000path, haghtalab2018provable, dobson2013sparse, dobson2013improving}, and single-query rapidly-exploring random trees (tree-based planners) \cite{lavalle1998rapidly, kuffner2000rrt, salzman2016asymptotically, li2016asymptotically, jaillet2010sampling, hauser2015lazy, orthey2019rapidly, orthey2020multilevel, arslan2012role, otte2015rrt, gammell2014informed, qureshi2013adaptive, qureshi2016potential}. Table \ref{tab: sbmpclassic} summarizes representative planners from each category with their primitives and limitations.

\textbf{Probabilistic Roadmaps (PRMs):} PRMs \cite{kavraki1996probabilistic} operates in two phases: graph construction and path-finding. During the graph construction phase, the algorithm randomly generates samples in the robotic manipulator's configuration space and adds them as a new node to the graph. Then the algorithm tries to connect the new node to the existing nodes of the graph using a specific distance metric and collision avoidance. This process will create a graph in the free configuration space ($\mathcal{C}_{free}$). In the path-finding phase, the algorithm inserts the start and goal configurations of the planning problem into the constructed graph and uses graph search algorithms to find a path connecting these two nodes.

\vspace{0.2cm}
\noindent
\textbf{Tree-based Planners:} In the context of probabilistic roadmaps, ``multi-query'' means that the graph constructed during the first phase can be used for various start and goal configurations to solve different path planning problems. This also can be achieved by simply solving a series of ``single-query'' problems for different start and goal configurations \cite{karaman2011sampling}. In ``single-query'' planners, a tree is incrementally built from the start configuration to the goal configuration, which reduces the planning time compared to ``multi-query'' planners. This results in a widespread application of single-query planners compared to probabilistic roadmaps.

\begin{figure}[htbp] 
\centering
\includegraphics[width=3 in]{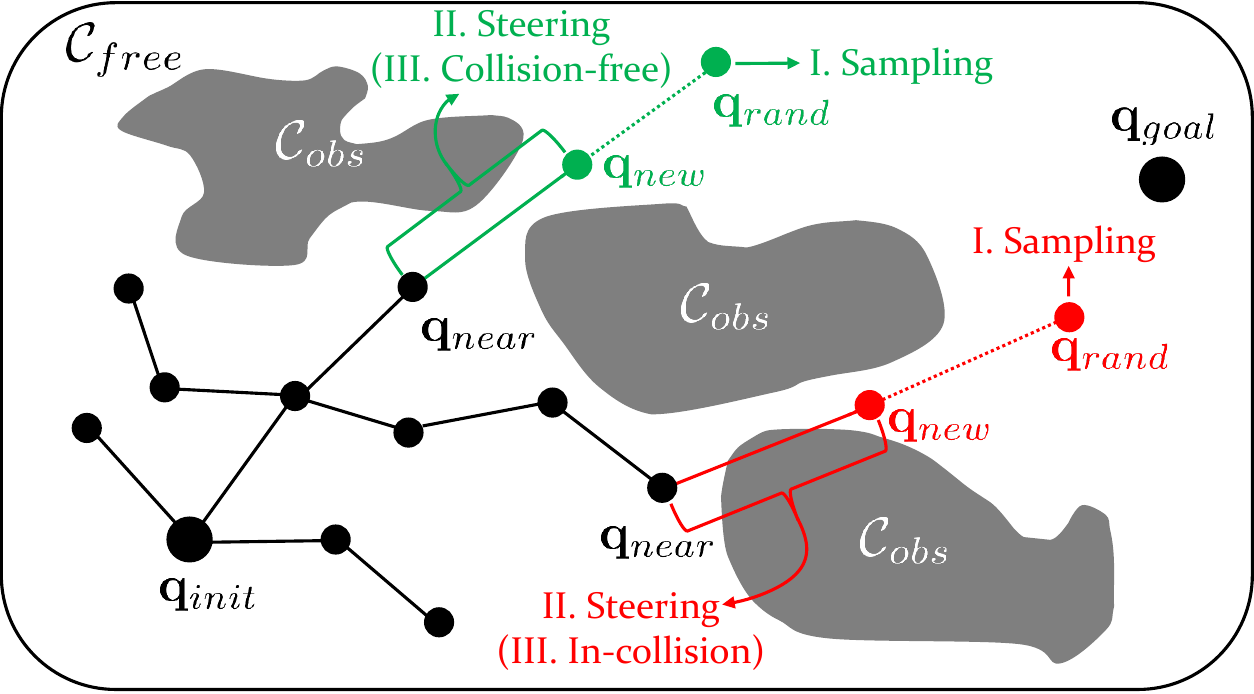}
\caption{Algorithmic primitives (I. Sampling, II. Steering, and III. Collision Checking) of sampling-based planning algorithms.}
\label{fig_4- primitives}
\end{figure}

Tree-based planners, such as Rapidly-exploring Random Tree (RRT) algorithm \cite{lavalle1998rapidly}, have three algorithmic primitives: First, random sampling within the configuration space (\textit{Sampling}). Second, steering from existing sampled configurations to the new sample (\textit{Local Planning}). Third, collision checking the steering connections (\textit{Collision Checking}). Figure \ref{fig_4- primitives} illustrates the algorithmic primitives of sampling-based planning algorithms. These algorithmic steps enable an implicit representation of the configuration space, making tree-based planners an effective choice for path-planning problems within the high-dimensional configuration space of robotic manipulators.

\vspace{0.2cm}
\noindent
\textbf{Sampling-based Planning Algorithms Variants and Extensions}: Given that basic sampling-based planning algorithms are effective in finding the feasible path within high-dimensional configuration spaces due to its probabilistic completeness, several variants and extensions of the sampling-based planning algorithms have been developed to further enhance their performance. Some of the variants and extensions are as follows:

\begin{itemize}
    \item \textbf{Informed sampling}: Sampling functions in sampling-based planners can be either random sampling or informed sampling. Random sampling functions indiscriminately select points from the entire configuration space to construct the tree. In contrast, informed sampling functions adaptively sample the configuration space. Informed samplers commonly utilize hand-crafted heuristics to direct the sampling process towards configuration space regions with high success rates and low planning costs \cite{gammell2014informed}.
    
    \item \textbf{Lazy collision checking}: This technique performs collision checking in the path construction phase, rather than during the initial graph or tree construction stage of the path planning algorithm. By postponing collision checking, the algorithm can focus on rapidly generating a graph or tree structure without being overwhelmed by the computational overhead of immediate collision evaluations \cite{hauser2015lazy, bohlin2000path}.
    
    \item \textbf{Bi-directionality}: In this module, two sampling-based planners are developed simultaneously: one starts from the start node, and the other from the goal node. This module enhances the efficiency of the path planning algorithm by concurrently expanding these planners towards each other \cite{kuffner2000rrt}. 
    
    \item \textbf{Optimality}: This module makes it possible for the sampling-based planner to find an asymptotically optimal path by utilizing tree rewiring operations after each sampling iteration. The rewiring process continually looks for ways to reduce the overall path length, which eventually leads to an asymptotically optimal solution \cite{karaman2011sampling}.
\end{itemize}

\vspace{0.2cm}
\noindent
\textbf{Advanced Variants:} There are several families of sampling-based planning algorithms that extend beyond the basic framework of probabilistic roadmaps and tree-based planners. The first family is Fast Marching Tree (FMT*) algorithms \cite{janson2015fast, starek2015asymptotically, reid2020sampling}. These algorithms combine features of both PRM and RRT algorithms by using a lazy dynamic programming approach on samples to grow a tree of paths. The FMT* algorithm leverages the strength of both PRM and RRT by incorporating elements of PRM's multi-query and RRT's single-query approach. Another family is Batch Informed Trees (BIT*) algorithms \cite{gammell2020batch, strub2020advanced, strub2020adaptively}, \cite{li2023biait}. The BIT* algorithm combines search-based planners (e.g., A* \cite{goto2003heuristics}), and sampling-based planners (e.g., RRT* \cite{karaman2011sampling}) to approximate and search the configuration space. The BIT* algorithm efficiently explores the configuration space by prioritizing regions with a higher probability of containing an optimal path.

\vspace{0.2cm}
\noindent
\textbf{Summary}: Sampling-based planning algorithms, have three algorithmic primitives, \textit{sampling}, \textit{steering}, and \textit{collision checking}.

\begin{itemize}
    \item \textbf{Sampling:} There are two types of sampling functions in sampling-based planners: uniform sampling and informed sampling. Uniform sampling indiscriminately samples from the configuration space, leading to computational complexity. Informed sampling functions sample within promising areas of the configuration space to reduce computational complexity. Classical planners commonly use hand-crafted heuristics for this purpose, which is challenging to implement for robotic manipulators. In Section \ref{subsubsec: DL-sampling-based-primitive} we explore how various deep learning frameworks can improve this part of sampling-based planning algorithms for robotic manipulators.

    \item \textbf{Steering}: Classical sampling-based algorithms steer toward newly sampled configurations using straight-line paths, which can violate motion constraints of robotic manipulators. Recently, deep learning frameworks have been employed to learn efficient steering functions to improve sampling-based planning algorithms (Section \ref{subsubsec: DL-sampling-based-steering}).

    \item \textbf{Collision Checking}: Classical sampling-based algorithms perform fine-grained geometric collision queries to validate each edge of the constructed tree, which is computationally intensive. Recently, deep learning frameworks have been utilized as proxy collision checkers to enable efficient, continuous collision checking (Section \ref{subsec:DL-collisionchecking}).
\end{itemize}

\subsubsection{Constrained Planning}
Geometric task constraints, such as maintaining a certain pose for a robotic manipulator end-effector, are types of constraints that require modifications to the sampling-based planning algorithms \cite{pettinger2024efficient}. These constraints are prevalent in real-world applications of robotic manipulators, which require modified sampling-based planning algorithms.

\vspace{0.2cm}
\noindent
\textbf{Definition:} In constrained planning, the planner must not only avoid collisions but also satisfy hard geometric task constraints. Thus, the constrained planning problem involves finding a path between any specified initial and goal configuration that lies entirely within the union of free configuration space and constraint manifold.

There are several methods to incorporate constraint adherence within sampling-based planning algorithms. The first approach is \textit{projection}, where a projection operator is utilized to project a given configuration onto the constraint manifold \cite{kingston2019exploring, berenson2011task}. Another method involves \textit{tangent spaces}, which calculate the tangent space of the constraint manifold, allowing for generation of nearby samples \cite{kim2016tangent}. These tangent spaces can be stored and integrated into an \textit{atlas}, providing a linear piecewise approximation of the constraint manifold \cite{jaillet2017path}. For more details on constrained sampling-based planning, refer to the survey paper by Kingston \textit{et al.} \cite{kingston2018sampling}.

\vspace{0.2cm}
\noindent
\textbf{Summary}: Classical constrained sampling-based planning algorithms utilize \textit{projection} and \textit{continuation} operators to enforce kinematic constraints. Recently, deep learning frameworks have been employed to learn the constraint manifold for direct constraint-aware sampling (Section \ref{subsec:DL-constrainedsampling}).

\subsubsection{Limitations of Sampling-based Planning Algorithms}
Although sampling-based motion planners are widely used for robotic manipulators' path planning, they come with several limitations and drawbacks, including:

\begin{itemize}
    \item \textbf{Curse of dimensionality}: The number of samples for path planning may increase exponentially with the dimension of the configuration space. Sometimes the sampling-based planners need to cover the configuration space with discrete samples.
    
    \item \textbf{Path smoothness}: The output of sampling-based planners often contains many unnecessary nodes which result in jerky motions, and takes longer to execute. The probabilistic nature of sampling-based planners results in paths with unnecessary motions and discontinuities, which necessitate post-processing steps to improve the quality of the generated path. For more details, refer to Section \ref{subsec:classic-optimization-based}.
    
    \item \textbf{Low convergence rate}: Sampling-based planning algorithms demonstrate low convergence rates in the high-dimensional environment, due to random sampling.
    
    \item \textbf{Hardly applicable to dynamic environments:} Planning in dynamic environments necessitates regular re-planning for collision avoidance \cite{pi2024omepp}. The sampling-based motion planning approaches incur a significant computational cost, rendering them unsuitable for real-time computation and reactive motion planning.

    \item \textbf{Sample inefficiency}: Due to the sampling primitives of these algorithms, these planners often struggle to provide enough samples in important areas within the configuration space (e.g., near obstacles and in narrow spaces). This reduces their capability to model the configuration space \cite{cheng2020learning} explicitly. In highly complex workspaces and narrow passages, these algorithms may reach a singularity such that the constructed graph becomes disconnected and planning fails.
\end{itemize}

\subsection{Optimization-based Planning Algorithms} \label{subsec:classic-optimization-based}
Sampling-based planners are effective and fast for path planning in high-dimensional configuration spaces. However, these planners can not guarantee the smoothness and local optimality of the planned path. Optimization-based planners address these drawbacks by post-processing the planned path to eliminate redundant or jerky motions \cite{ratliff2009chomp}. There are two primary methods for this post-processing: gradient-free and gradient-based optimization methods \cite{orthey2023sampling}. Table \ref{tab: trajoptclassic} summarizes trajectory optimization planners with their characteristics and limitations.

\vspace{0.2cm}
\noindent
\textbf{Gradient-free Optimization Methods:} A family of gradient-free methods aims to shorten the path by removing redundant nodes. These methods typically employ (partial) short-cutting \cite{kavraki1996probabilistic, chen1998sandros, geraerts2007creating}, which involves tree pruning, or hybridization \cite{raveh2011little}. Another family focuses on smoothing the path by utilizing smooth curves to interpolate between the path's waypoints. Techniques such as B-splines \cite{maekawa2010curvature}, Bezier curves \cite{yang2013continuous}, and cubic polynomials \cite{thrun2006stanley} are commonly utilized for this purpose.

\setlength{\tabcolsep}{3pt}
\begin{table*}[htbp]
\begin{center}
\caption{The two main categories of trajectory optimization planning algorithms for robotic manipulators, their constituent algorithms, characteristics, and limitations.}
\resizebox{\textwidth}{!}{%
\label{tab: trajoptclassic}
\begin{tabular}
{>{\raggedright\arraybackslash}p{0.1\textwidth}>{\raggedright\arraybackslash}p{0.3\textwidth}>{\raggedright\arraybackslash}p{0.3\textwidth}>{\raggedright\arraybackslash}p{0.3\textwidth}}
\arrayrulecolor{gray!80}
\hline
\rowcolor{gray!20} \textbf{Category}&\textbf{Algorithms}&\textbf{Characteristics}&\textbf{Limitations} \\
\toprule
\textbf{Gradient-free}&\scriptsize{Short-cutting \cite{kavraki1996probabilistic, chen1998sandros, geraerts2007creating}, Path Smoothing by Utilizing Smooth Curves (i.e., B-splines \cite{maekawa2010curvature}, Bezier Curves \cite{yang2013continuous}, Cubic Polynomials \cite{thrun2006stanley}}&$\bullet$ Post-processing the output of sampling-based planning algorithms (Removing redundant nodes, smoothing)&$\bullet$ Computationally complex \\
\midrule
\textbf{Gradient-based}&\scriptsize{Elastic Band \cite{quinlan1993elastic}, Elastic Strips \cite{brock1998elastic}, CHOMP \cite{ratliff2009chomp, zucker2013chomp}, TrajOpt \cite{schulman2013finding, schulman2014motion}, GPMP \cite{mukadam2016gaussian}, GPMP2 \cite{mukadam2018continuous}, STOMP \cite{kalakrishnan2011stomp}, GCS \cite{marcucci2023motion, cohn2023non, cohn2024constrained, garg2024planning}}& $\bullet$ Refining an initial path subject to kinematic and task-specific constraints.& $\bullet$ Getting stuck in local minima. \newline $\bullet$ Computationally complex. \newline $\bullet$ Hand-crafted cost functions. \newline $\bullet$ Requires collision gradient. \newline $\bullet$ Dependent on the initial guess \\
\bottomrule
\end{tabular}}
\end{center}
\end{table*}

\vspace{0.2cm}
\noindent
\textbf{Gradient-based Optimization Methods:} Gradient-based optimization methods leverage optimization techniques from the field of optimal control to iteratively refine the initial path subject to planning constraints. These methods formulate planning as an optimization problem subject to kinematic and task-specific constraints \cite{orthey2023sampling} (cf. motion planning definition in Section \ref{sec:planning_def}).

One approach to optimizing the initial path involves utilizing \textit{elastic band} \cite{quinlan1993elastic} or \textit{elastic strips} \cite{brock1998elastic, brock2002elastic, yang2010elastic} methods to locally deform the initial path for smoothness and collision avoidance. These methods model the initial path as an elastic band subjected to two types of artificial forces: an internal force that maintains the connectivity of the path, and an external repulsion force that steers the path away from obstacles.

Another gradient-based approach is global optimization, which also refines the initial trajectory utilizing a numerical optimization method, subject to planning and task-specific constraints \cite{warren1989global, ratliff2009chomp, dragan2011manipulation, zucker2013chomp, he2013multigrid, byravan2014space, marinho2016functional, osa2020multimodal, schulman2013finding, schulman2014motion, mukadam2016gaussian, lambert2021entropy, alwala2021joint, dong2016motion, huang2017motion, mukadam2018continuous, wang2022bpomp, yu2023gaussian, barcelos2024path, cohn2024constrained, power2024constrained, yu2023stochastic}. An early method proposed for this approach involves subjecting the initial trajectory to repulsive and attractive artificial potential fields \cite{khatib1986real} to improve its quality \cite{warren1989global}. 

The Covariant Hamiltonian Optimization for Motion Planning (CHOMP) \cite{ratliff2009chomp, zucker2013chomp}  utilizes covariant gradient descent \cite{bagnell2003covariant} as the numerical optimization technique to optimize the initial, in-collision trajectory. CHOMP discretizes the initial trajectory into waypoints and subjects them to trajectory smoothness and obstacle avoidance, modeled as pre-computed signed distance fields \cite{quinlan1995real}. In cluttered environments, a fine-grained trajectory discretization is needed for optimization, which increases computational complexity \cite{wang2022bpomp}. 

Trajectory Optimization (TrajOpt) employs continuous-time collision checking \cite{schulman2014motion} to ensure continuous-time safety, thereby reducing the need for a large number of trajectory states \cite{schulman2013finding, schulman2014motion}. TrajOpt utilizes sequential convex optimization \cite{werner2012optimization} for global optimization, which addresses the non-convexity problem by solving a series of convex optimization problems. Like CHOMP, TrajOpt also requires a densely parameterized trajectory in clutter environments.

The main drawback of CHOMP and TrajOpt is the need for a finely discretized trajectory for optimization. Continuous-trajectory representation, like radial basis functions (FBS) \cite{marinho2016functional}, and Gaussian process (GP) \cite{mukadam2016gaussian}, has helped alleviate this problem. Gaussian Process Motion Planning (GPMP) \cite{mukadam2016gaussian} addresses this issue by parameterizing the initial trajectory via a few states and utilizing a continuous Gaussian process for interpolation to query the initial trajectory.

GPMP2 \cite{mukadam2018continuous}, \cite{meng2022anisotropic} improves upon GPMP by considering the trajectory optimization problem as probabilistic inference \cite{toussaint2009robot, toussaint2010bayesian} on a factor graph \cite{kschischang2001factor}. It leverages Incremental Smoothing and Mapping (iSAM2) \cite{kaess2012isam2} optimization algorithm for trajectory optimization.

The main drawback of the above-mentioned gradient-based optimization methods is that both cost and constraint functions need to be differentiable. Stochastic trajectory optimization methods relax the requirement for differentiable constraints by utilizing sampling methods \cite{kalakrishnan2011stomp, park2012itomp, park2013real, osa2020multimodal, kim2015trajectory, petrovic2019stochastic, petrovic2020cross, petrovic2022mixtures, le2024accelerating, mainprice2013human}. The Stochastic Trajectory Optimization for Motion Planning (STOMP) \cite{kalakrishnan2011stomp, park2012itomp} improves upon the aforementioned trajectory optimization methods by employing a derivative-free stochastic optimization approach. This algorithm starts by generating a set of random trajectories around the initial candidate solution. Then, these trajectories are evaluated using planning cost functions to update the candidate solution \cite{theodorou2010reinforcement}.

Another drawback of optimization-based planning algorithms comes from the non-convexity of the planning problem. Graphs of Convex Sets (GCS) \cite{marcucci2023motion, cohn2023non, cohn2024constrained, garg2024planning} address this limitation by modeling non-convex obstacle-avoidance constraints as a collection of safe convex regions \cite{amice2022finding, petersen2023growing, dai2024certified}. This approach simplifies the planning process into two steps: selecting which convex region to traverse (the discrete component), and optimizing the parameterized robot trajectory within each region (the continuous component). This formulation reduces the planning problem to a Shortest-Path Problem (SPP) \cite{marcucci2024shortest} over the GCS.

Although optimization-based trajectory planners have been widely used in the research community for trajectory planning, there are several limitations and drawbacks associated with them, including:

\begin{figure}[htbp] 
\centering
\includegraphics[width=2.5 in]{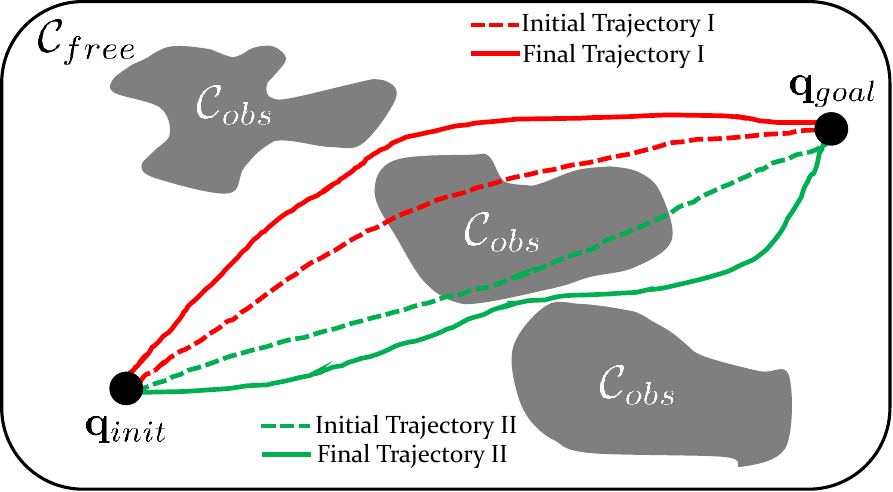}
\caption{An abstract illustration demonstrating how global trajectory optimization techniques rely on the initial trajectory to warm-start the optimization process. This highlights that even slightly different initializations can lead to distinct final trajectories.}
\label{fig_5- optimization-based}
\end{figure}

\begin{itemize}
    \item \textbf{Local minima - lack of formal completeness guarantees}: Optimization-based planners can easily get stuck in local minima due to the inherent non-convexity of the optimization problem.
    
    \item \textbf{Collision gradients}: The collision detection module in the optimization-based motion planning algorithms is required to provide collision gradients. The gradient helps the optimal trajectory steer away from the obstacles. The efficiency of optimization-based algorithms depends on the smoothness and continuity of the collision gradients. However, existing classical collision detection methods only provide numerical \cite{schulman2014motion} or stochastic gradients \cite{kalakrishnan2011stomp}.
    
    \item \textbf{Optimization warm-starting}: As demonstrated in Figure \ref{fig_5- optimization-based}, the output of optimization-based trajectory planning methods is dependent on the initial trajectory. Therefore, a feasible initialization is needed to avoid getting stuck in local minima.
    
    \item \textbf{Long-horizon, global optimization}: Trajectory optimization methods solve the optimization problem over a long temporal horizon, which is challenging for dynamic planning and real-time obstacle avoidance.
    
    \item \textbf{Requiring hand-crafted cost functions}: Trajectory optimization methods often require hand-crafted cost functions to encode task-specific constraints, and desired behaviors \cite{osa2017guiding}. Within complex planning environments, the cost function can become ill-conditioned and lead to exploding and vanishing gradients. In turn, this will lead to algorithmic singularity and the optimization problem cannot converge to a solution.
\end{itemize}

\vspace{0.2cm}
\noindent
\textbf{Summary:} One class of optimization-based algorithms uses gradient-free methods, such as short-cutting \cite{geraerts2007creating}, B-Splines \cite{maekawa2010curvature}, and Bezier Curves \cite{yang2013continuous}, to post-process the output of sampling-based algorithms such that the resulting path is smooth and jerk-free.

Another class of optimization-based algorithms, known as trajectory optimization (TO), uses gradient-based optimization techniques to refine an initial straight-line path such that it satisfies motion planning constraints. A key challenge with these methods is their dependence on the initial trajectory, which can lead to convergence to local minima. Recently, deep learning frameworks have been utilized to warm-start the optimization process to improve the efficiency of these algorithms (Section \ref{subsec:DL-optimization}).
\subsection{Collision Checking} \label{subsec: classic-collision-checking}
Collision avoidance is an important component of motion planning for robotic manipulators, which ensures the robot avoids self-collision and collisions with the environment for efficient plan execution \cite{sundaralingam2023curobo, thomason2024motions}.

\begin{figure}[!htbp] 
\centering
\includegraphics[width=0.98\linewidth]{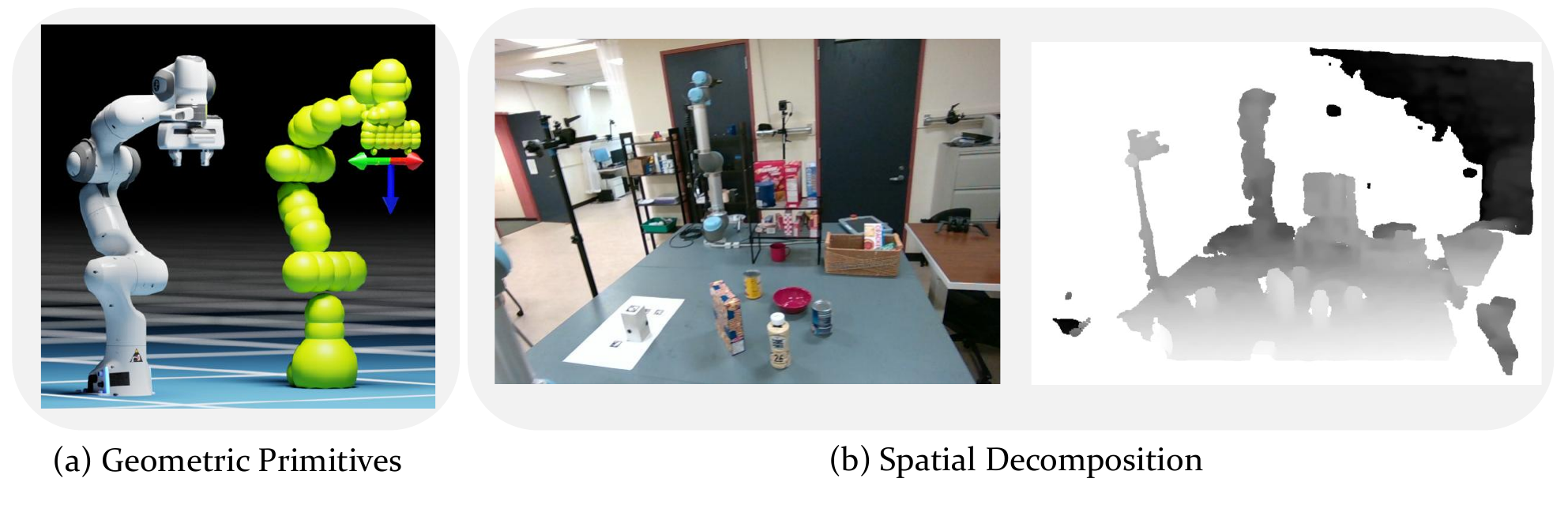}
\caption{Common practices for collision checking: (a) Convex geometric primitives \cite{sundaralingam2023curobo}. (b) Spatial decomposition \cite{ramsey2024collision}.}
\label{fig_6- collision-checking}
\end{figure}

Collision and proximity queries are important algorithmic primitives of sampling-based and optimization-based planning algorithms. In sampling-based planning algorithms, 90\% of the computation time is consumed for collision queries. The primary goal of collision avoidance is to predict potential collisions and redirect the planned path toward the free configuration space. 

\vspace{0.2cm}
\noindent
\textbf{Geometric Collision Checking:} Geometric collision checking methods utilize geometric primitives (e.g., spheres, ellipsoids), and employ the Gilbert-Johnson-Keerthi (GJK) algorithm and its variants for collision detection \cite{gilbert1988fast, bergen1999fast}. A major limitation of the geometric primitive method is its dependency on the number of obstacles within the configuration space; as the number of obstacles increases, so does the computational complexity of the collision-checking method. To address this issue, hierarchical representations are utilized \cite{lin2017collision}, where a coarse representation is utilized for initial fast collision checking, and a finer representation is utilized when necessary (e.g., near or in-collision scenarios).

There are generally two methods for hierarchical representation of the workspace, as illustrated in Figure \ref{fig_6- collision-checking}: bounding volume hierarchy (BVH), and spatial decomposition (SD) \cite{fan2022fast}. Bounding volume hierarchy is an object-centric method that utilizes bounding volumes, such as spheres, discrete oriented polytope (k-DOP) \cite{klosowski1998efficient}, and oriented bounding box (OBB) \cite{gottschalk1996obbtree}, for hierarchical representation of the environment \cite{pan2012fcl}. Spatial decomposition, on the other hand, is a space-centric method, that employs partitioning techniques such as K-d trees \cite{bentley1975multidimensional}, octrees \cite{meagher1982geometric}, and space-time bounds to decompose the workspace into cells. Cells are considered occupied if an object occupies that cell \cite{lin2017collision}. Both hierarchical representations are utilized for continuous \cite{redon2005fast} and point cloud collision detection \cite{ramsey2024collision}. Continuous collision checking is most suitable for local steering in sampling-based planning algorithms, while point cloud collision checking is capable of performing collision queries between point clouds and point clouds and other geometric primitives. Furthermore, these collision-checking algorithms can be parallelized to leverage the capabilities of GPUs \cite{pan2010gpu} and multi-core CPUs \cite{tang2008interactive} for enhanced parallel computations.

\vspace{0.2cm}
\noindent
\textbf{Signed Distance Fields (SDFs)}: The Signed distance function is an alternative methodology for representing collision distances. It computes the distance between a point in the workspace and a surface, assigning a zero value to points on the surface. other points receive a signed value indicating the distance and direction relative to the surface. The resulting Singed Distance Field (SDF) can be used as collision avoidance constraints in various motion planning algorithms \cite{ratliff2009chomp}. These methods provide a distance field and its gradient for proper collision checking within trajectory optimization methods \cite{oleynikova2017voxblox, pan2022voxfield, millane2024nvblox}.

\vspace{0.2cm}
\noindent 
\textbf{Swept Volume}: The swept volume between two manipulator configurations represents the total volume occupied by the manipulator as it moves from one configuration to another. It is commonly used for continuous collision detection. However, computing the exact swept volume for articulated objects, such as robotic manipulators, is computationally complex. To address this, several approximation methods have been proposed in the literature, including polyhedra-based \cite{gaschler2013robot}, occupation grid-based \cite{taubig2011real}, and boundary-based \cite{campen2010polygonal} algorithms.

\vspace{0.2cm}
\noindent
\textbf{Machine learning-based Proxy Collision Checkers:} Proxy collision checkers have been developed to improve the computational efficiency of motion planning algorithms for robotic manipulators by replacing conventional geometric collision-checking methods \cite{ das2020learning}. These proxy collision checkers are utilized during the tree construction phase, while final path validity is checked by an exact geometric collision checker. Proxy collision checkers must maintain high accuracy to avoid false negatives and scale well to high-dimensional configuration spaces. Various machine learning-based binary classifiers, including parametric methods like Support Vector Machines (SVMs), and K-Nearest Neighbors (KNNs), as well as non-parametric approaches like Gaussian Mixture Models (GMMs), and Gaussian Processes (GPs) have been utilized as proxy collision checkers \cite{verghese2022configuration, zhi2022diffco, pan2015efficient, huh2016learning, pan2016fast, huh2017adaptive, han2019configuration, das2020forward}.

\vspace{0.2cm}
\noindent
\textbf{Collision Detection for pHRI:} In addition to proactive collision avoidance for motion planning, another line of research focuses on utilizing sensors for contact detection \cite{kuccuktabak2024physical} in physical human-robot interaction (pHRI) scenarios where human and robotic manipulators operate in close proximity. In pHRI scenarios, proprioceptive sensors primarily detect collisions, as exteroceptive sensors may not adequately predict potential contacts between the robot and its environment. Collision detection involves monitoring changes in the electrical currents of manipulator drives \cite{yamada1997human}, the differences between measured torques and the nominal control law \cite{takakura1989approach}, \cite{zhang2020online}, and utilizing tactile sensors \cite{fan2022enabling}. After a contact is detected, the collision needs to be identified, and processed to determine the intentionality, locality, duration, and severity of it for efficient pHRI \cite{golz2015using}. For a detailed review of contact detection and collision management, refer to the survey paper by Haddadin \textit{et al.} \cite{haddadin2017robot}.

\vspace{0.2cm}
\noindent
\textbf{Summary:} Collision checking takes up to 90\% of the computation time in motion planning algorithms. Classical approaches rely on methods like geometric collision detection \cite{klosowski1998efficient, ramsey2024collision}, signed distance fields \cite{ratliff2009chomp}, swept volume estimation \cite{gaschler2013robot}, and machine learning-based proxy collision checkers \cite{ das2020learning}. In recent years, the parallelization, auto-differentiation, and fast inference capabilities of deep neural networks have been leveraged to improve the efficiency of collision checking algorithms (Section \ref{subsec:DL-collisionchecking}).

\section{Deep Learning Basics and Potential for Robotic Manipulators Motion Planning} \label{sec:DL-modules}
In this section, we introduce various building blocks of deep learning methods used for manipulator motion planning. 

\subsection{Basic Deep Learning Frameworks} \label{subsec: dlbasicframeworks}
This section provides a brief introduction to basic deep learning frameworks used in motion planning for robotic manipulators. Table \ref{tab: dlmodules1} provides an overview of these frameworks and their applications in robotic manipulator motion planning.

\vspace{0.2cm}
\noindent
\textbf{Multi-layer Perceptrons (MLPs):} Multi-layer perceptron is an architecture that consists of stacking several fully connected layers. Each layer in an MLP has two parts: an affine transformation and an activation function. The activation function introduces nonlinearity into the network such that MLPs are realized as universal function approximations \cite{zhang2023dive}.

The universal function approximation property of multilayer perceptrons (MLPs) has enabled their utilization for learning complex mappings in robotic manipulators' motion planning. When trained on an oracle planning dataset, these structures rely on hierarchical nonlinear transformations to capture global planning patterns. This allows MLPs to generate end-to-end plans, replace the sampling primitive in sampling-based algorithms for informed sampling, warm-start trajectory optimization with high-quality initial guesses, or serve as a proxy collision checker to accelerate planning.

\vspace{0.2cm}
\noindent
\textbf{Convolutional Neural Networks (CNNs):} Although MLPs are efficient in learning interactions between various features in tabular data \cite{borisov2110deep}, they are agnostic toward existing patterns within the data. In contrast, Convolutional Neural Networks leverage the translation invariance \cite{zhang1988shift} and locality properties of convolutional kernels to capture spatial hierarchies and patterns within grid-like topology data (e.g., images). Translation invariance ensures that all regions of an image are treated uniformly, while locality focuses on small neighborhoods to encode hidden representations. These properties make CNNs highly effective for image recognition, and object detection, with applications in robotic manipulator autonomy stack \cite{yu2020robotic}.

In real-world motion planning, the exact geometry of obstacles is often unknown, and planners often rely on partial or noisy sensed representations obtained from cameras or depth sensors. The locality and translation invariance of CNNs enable them to extract spatially consistent features from sensed representations such as occupancy grids, depth maps, or RGB images. CNNs can capture both local obstacle geometry and global workspace structure by hierarchically composing local features into higher-level representations. This allows CNNs to be integrated into motion planning for end-to-end planning, to guide sampling in sampling-based algorithms, or serve as a proxy collision checker to accelerate planning.

\vspace{0.2cm}
\noindent
\textbf{Recurrent Neural Networks (RNNs):} MLPs and CNNs are well-suited for handling fixed-length data, i.e., structured tabular data and grid-like data structures. However, these frameworks fall short for applications in domains such as time series prediction and language processing, where inputs are sequential and vary in length \cite{lu2019learning}.

Recurrent Neural Networks (RNNs) \cite{schuster1997bidirectional, sutskever2014sequence} feature recurrent connections that capture the dynamics within sequence data by maintaining a form of memory across the input. This memory is achieved by incorporating cycles into the network's architecture, which allows RNNs to process data sequences \cite{xu2024leto}.

Motion planning for robotic manipulators is inherently a long-horizon problem that requires reasoning over temporal dependencies between successive configurations. RNNs allow information from earlier steps in the planning to influence later decisions by maintaining a hidden state that is updated sequentially, which enables RNNs to capture temporal correlations. Therefore, RNNs can be applied to generate end-to-end motion plans by encoding sequential dependencies, or to bias samples in sampling-based algorithms toward consistent regions of the configuration space.

\vspace{0.2cm}
\noindent
\textbf{Graph Neural Networks (GNNs):} MPLs, CNNs, and RNNs are specialized in handling structured data, i.e., tabular, grid-like, and sequential. Graph Neural Networks (GNNs) extend beyond these limitations by being powerful tools for operating on both structured and unstructured data \cite{battaglia2018relational}.

Graphs and GNNs have been used for the representation and encoding of data structures in a variety of applications, including robotic tactile recognition \cite{zhang2025multi}, robotic grasping \cite{ding2025dual}, surface defect recognition \cite{wang2022graph}, and water quality prediction \cite{qiao2024attention}, showcasing their capability in representing complex data structures and the relationships between their elements.

\setlength{\tabcolsep}{3pt}
\begin{table*}[htbp]
\begin{center}
\caption{Schematic of basic Deep learning frameworks, their characteristics, and their potential for improving various components of classical planning algorithms for robotic manipulators.}
\label{tab: dlmodules1}
\begin{tabular}
{>{\raggedright\arraybackslash}p{1.0\textwidth}}
\includegraphics[width=\textwidth, trim={0.0cm 0.0cm 0.0cm 0.0cm},clip]{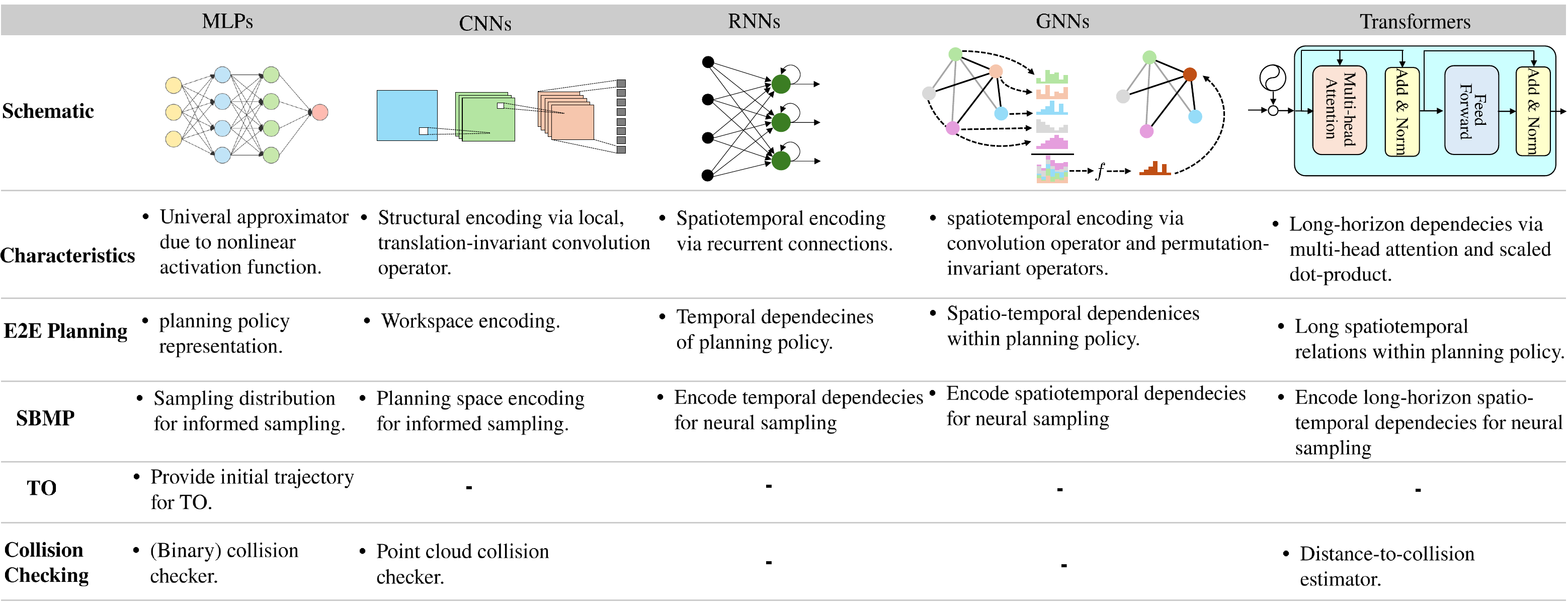}
\end{tabular}
\begin{minipage}{\textwidth}
\smallskip
\footnotesize
\textbf{Note:} ``E2E Planning'' denotes end-to-end planning, ``SBMP'' denotes sampling-based motion planning algorithms, and ``TO'' denotes trajectory optimization algorithms.
\end{minipage}
\end{center}
\end{table*}

The workspace and configuration space in robotic manipulator motion planning are inherently high-dimensional, unstructured, and exhibit spatiotemporal dependencies between workspace entities. GNNs have the potential to encode these dependencies by representing the planning problem as a graph, where nodes encode workspace entities such as robots or obstacles, and edges capture their relationships. The GNN aggregates local and global dependencies with iterative message passing while preserving permutation invariance. As a result, GNNs are capable of encoding both spatial correlations (e.g., robot-to-obstacle proximity) and temporal dependencies within planning waypoints. Given these properties, GNNs can be employed for end-to-end planning, to guiding the sampling primitive in sampling-based algorithms toward feasible regions, or serve as a proxy collision checker to accelerate planning.

\vspace{0.2cm}
\noindent
\textbf{Transformers:} The early boom in deep learning was powered by foundational architectures like MLPs, CNNs, and RNNs. However, the current wave of progress in deep learning is predominantly driven by the transformer architecture. Central to the transformer architecture is the multi-head attention mechanism (scaled-dot product), which was originally proposed by Vaswani \textit{et al.} \cite{vaswani2017attention}.

Transformers have become central to natural language processing tasks \cite{devlin2018bert}, and have also emerged as the default model for numerous vision tasks, including image recognition and object detection \cite{dosovitskiy2020image}. Moreover, transformers have demonstrated impressive performance in other domains such as reinforcement learning \cite{chen2021decision} and GNNs \cite{dwivedi2020generalization}.

The multi-head attention mechanism operates on a collection of embedding vectors that represent candidate waypoints or states along a planning trajectory. Within each head, the scaled dot-product attention computes similarity between a query vector (e.g., the current waypoint) and all key vectors (other waypoints), to produce attention coefficients. These coefficients weight the corresponding value vectors to selectively attend to past, present, or future waypoints that are most relevant to the current planning state. Subsequently, multi-head attention encodes diverse dependency patterns within the planning dataset (e.g., local smoothness vs. global feasibility).

Within the motion planning framework, the multi-head attention mechanism encodes long-range spatiotemporal dependencies across the planning trajectories, such that early waypoints can directly influence the prediction of later ones to improve the consistency of long-horizon trajectories. Consequently, this framework can be leveraged to learn and encode these dependencies for end-to-end planning, guiding sampling primitive of sampling-based algorithms, and performing collision checking.

\vspace{0.2cm}
\noindent
\textbf{Large Language Models (LLMs):} Large language models are built upon the transformer architecture and are extensively pre-trained on internet-scale datasets and can be fine-tuned for specific tasks \cite{zhou2023comprehensive}. These models hold the potential to significantly enhance various aspects of the robotics domain \cite{shah2023lm, song2025learning, li2025language}.

LLMs and vision-language models (VLMs) have been widely applied across the manipulation stack for robot policy learning, high-level task planning, and code generation \cite{firoozi2023foundation}. In policy learning, language-conditioned imitation learning leverages the semantic understanding of LLMs to enable vision-based manipulation guided by language instructions \cite{shridhar2023perceiver}. These models also help decompose complex manipulation tasks into simpler sub-tasks to assist reinforcement learning agents to interact with the environment \cite{di2023towards}. For task planning, LLMs have been utilized to generate high-level task plans for long-horizon, complex manipulation tasks \cite{ahn2022can}. Additionally, their code generation capabilities reduce the need for extensive domain knowledge in task planning and manipulation \cite{liang2022code}. These models can also be integrated with classical planning algorithms to form a modular framework for motion planning.

Although LLMs improve the generalizability of manipulation algorithms, they still face challenges such as distribution shift during policy deployment. Furthermore, their high computational demands and long inference times limit their deployment on real-world robotic systems. Additionally, these models are restricted to processing language without the capability to reason about the physical world.
\subsection{Generative Models} \label{subsec: DGMs}
Discriminative and generative models are two fundamental approaches for data modeling and prediction in deep learning. These models serve different purposes and are based on distinct principles for learning from data \cite{zhang2023dive}.

\begin{itemize}
    \item \textbf{Discriminative models}: A discriminative model is primarily applicable to tasks such as regression or classification, where the objective is to distinguish between different classes or predict a specific value. Essentially, a discriminative model aims to model the conditional probability of the output given the input, focusing on differences between classes.
    
    \item \textbf{Generative models}: A generative model learns to model the underlying distribution of the training data and uses this model for generating new data. For instance, an image generation algorithm can generate new images by learning the underlying statistical properties and patterns of an image dataset. 
\end{itemize}

\setlength{\tabcolsep}{3pt}
\begin{table*}[htbp]
\begin{center}
\caption{Schematic of various deep generative models, their characteristics, and their applications in improving various components of classical planning algorithms for robotic manipulators. The flow matching figure is adopted from \cite{lipman2024flow}.}
\label{tab: dlmodules2}
\begin{tabular}
{>{\raggedright\arraybackslash}p{1.0\textwidth}}
\includegraphics[width=\textwidth, trim={0.0cm 0.0cm 0.0cm 0.0cm},clip]{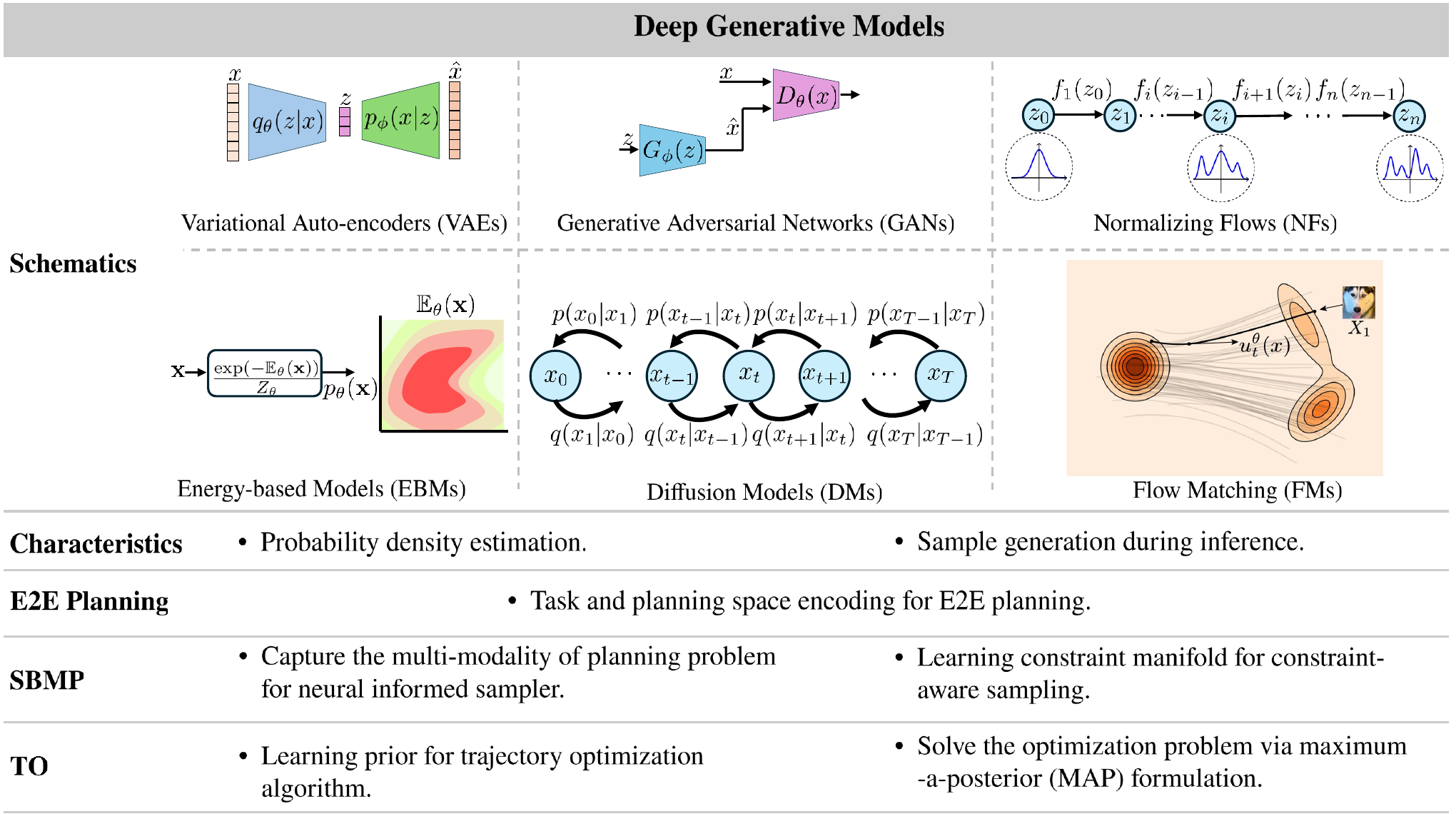}
\end{tabular}
\begin{minipage}{\textwidth}
\smallskip
\footnotesize
\textbf{Note:} ``E2E Planning'' denotes end-to-end planning, ``SBMP'' denotes sampling-based motion planning algorithms, and ``TO'' denotes trajectory optimization algorithms.
\end{minipage}
\end{center}
\end{table*}

Deep Generative Models (DGMs) \cite{bond2021deep} are neural networks designed to approximate complex, high-dimensional probability distributions within datasets. Once trained, a DGM can estimate the likelihood of observations and generate samples from the learned underlying distribution. These models have been widely used in diverse applications such as anomaly detection \cite{chai2024walk}, wind farm control \cite{huang2025wind}, and surface defect recognition \cite{tai2024defect}. Some of the most popular approaches in DGMs include Variational Auto-encoders (VAEs) \cite{zhai2018autoencoder}, Generative Adversarial Networks (GANs) \cite{pan2019recent}, Normalizing Flows \cite{kobyzev2020normalizing}, Energy-based Models (EBMs) \cite{song2021train}, Diffusion Models \cite{yang2023diffusion}, and Flow Matching \cite{lipman2022flow}. Table \ref{tab: dlmodules2} provides an overview of generative models, their characteristics, and their application in robotic manipulator motion planning.

Motion planning datasets are inherently multi-modal where multiple feasible trajectories may exist for a given planning problem. Deep generative models capture this inherent multi-modality by learning a distribution over trajectories and generate diverse candidates during inference. For instance, VAEs can encode trajectories into a latent space and decode them to generate new samples, GANs can learn to generate trajectories by matching the distribution of the planning dataset, and diffusion models can iteratively refine noisy trajectories via learned denoising steps. This generative capability allows deep generative models to be utilized to generate end-to-end plans, to generate informed samples within sampling-based planning algorithms, and to warm-starting trajectory optimization with high-quality initial guesses.

\vspace{0.2cm}
\noindent
\textbf{Variational Auto-encoders (VAEs):} VAEs are a class of deep generative models designed to encode the distribution within a dataset by mapping it into a latent space characterized by a Gaussian distribution \cite{kingma2013auto, rezende2014stochastic, sohn2015learning, rifai2011contractive, higgins2016beta, van2017neural, tolstikhin2017wasserstein}. The learned underlying distribution is then used to generate new data points.

VAEs consist of two components: an encoder and a decoder. The encoder is a neural network that maps a data point to a hidden representation in the latent space. The decoder is also a neural network that reconstructs the input data from the latent space representation.

\vspace{0.2cm}
\noindent
\textbf{Generative Adversarial Networks (GANs):} GANs consist of two components: a generator, and a discriminator \cite{goodfellow2020generative, mirza2014conditional}. These components engage in a competitive interaction, where the generator attempts to create data similar to the original data, while the discriminator tries to distinguish between the generated and the actual data. This competition allows GANs to generate samples that closely resemble the original training data.

\vspace{0.2cm}
\noindent
\textbf{Normalizing Flows (NFs):} NF principle \cite{kingma2018glow, berg2018sylvester, de2020block} is to transform a simple distribution into a more complex distribution that better represents the underlying characteristics of the dataset. This transformation is achieved through a sequence of invertible, and differentiable mappings \cite{rezende2015variational}.

\vspace{0.2cm}
\noindent
\textbf{Energy-Based Models (EBMs):} EBMs capture complex relationships within a dataset by defining a scalar energy function over the data space \cite{lecun2006tutorial}. The non-normalized probabilistic property of EBMs allows them to be compatible with various deep learning frameworks for energy function representation. However, this property makes training EBMs challenging. Additionally, in practical applications, selecting an appropriate energy function is non-trivial and requires careful consideration.

\vspace{0.2cm}
\noindent
\textbf{Diffusion Models:} Diffusion models operate by sequentially adding noise to the input data in a forward process and then removing this noise in a backward process known as the denoising process \cite{ho2020denoising}. The effectiveness of diffusion models relies on their ability to predict the noise added during the forward process.

\vspace{0.2cm}
\noindent
\textbf{Flow Matching:} Flow matching is an alternative to diffusion models for generative modeling. Unlike diffusion models, which rely on an iterative denoising process, flow matching \cite{lipman2022flow, lipman2024flow} directly learns a time-dependent vector field that transports the base distribution to the target distribution.
\subsection{Point Cloud Neural Networks} \label{subsec: PCNets}
3D point clouds, derived from 3D LIDAR and RGB-D cameras, provide more information about the environment compared to RGB images \cite{lu2020deep}. However, encoding 3D point clouds with image processing neural networks is challenging because they lack the inherent spatial structure found in image data. Consequently, novel deep learning frameworks are required for processing unstructured 3D point cloud data. Notable among proposed point cloud neural networks (PCNets) are PointNet \cite{qi2017pointnet}, PointNet++ \cite{qi2017pointnet++}, and Point Transformer \cite{zhao2021point}, which combine basic neural network frameworks and permutation invariance operators (e.g., sum, max) to operate on 3D point cloud data \cite{lin2023deep}.

In real-world settings, planners often have access to partial and noisy 3D point cloud observations of the workspace. Point cloud neural networks can encode such unstructured data, where the number and ordering of points may vary across observations. These models embed each point into a feature space and then aggregate features across neighborhoods to extract meaningful workspace representations. In the context of motion planning, these models can encode obstacle geometry, free-space structure, and robot–workspace interactions directly from sensor data. As a result, they can be leveraged for end-to-end planning, to guide informed sampling in sampling-based algorithms, and to accelerate collision checking with perception-driven embeddings.
\subsection{Neural Radiance Field} \label{subsec: NeRF}
Neural Radiance Field (NeRF) frameworks learn 3D scene representations from 2D images, and provide a promising method for encoding perception and motion in robotics \cite{wang2024nerf}. These frameworks take camera rays as input and output a volumetric rendering of the scene. This volumetric rendering has been integrated into classical path planning algorithms as occupancy representations for navigation \cite{deng2025neslam}. Additionally, NeRFs' 3D structural bias makes them useful for object pose estimation \cite{wen2023bundlesdf} and integration into manipulation policies \cite{driess2023learning}.

The success of motion planning algorithms for robotic manipulators largely depends on accurate scene representation for collision checking. NeRF's ability to reconstruct 3D scenes from 2D images can be used for dynamic collision checking between the manipulator and the neural field. Also, the sampling primitive of sampling-based algorithms can be conditioned on the reconstructed 3D scene for informed sampling \cite{adamkiewicz2022vision}.
\subsection{Neural SDFs} \label{subsec: DeepSDF}
Neural SDFs are signed distance functions learned using deep neural networks \cite{park2019deepsdf}, \cite{gropp2020implicit}. They are commonly used for scene reconstruction, where the zero-level set of the neural SDF is extracted to represent clear geometric surfaces within the scene. Collision checking is the major bottleneck in motion planning for robotic manipulators. Neural SDFs also can be considered as the collision avoidance constraint in motion planning for robotic manipulators (Section \ref{subsec:DL-collisionchecking}).
\subsection{Robotic Foundation Models} \label{subsec: robotfoundation}
LLMs are trained on large-scale internet datasets and have been applied to text generation and open-vocabulary visual recognition tasks. Their semantic reasoning and visual interpretation capabilities have been leveraged for high-level robotic planning. However, since these models primarily reason about semantics and textual prompts, they require additional auxiliary components to handle low-level motion planning and control tasks \cite{firoozi2023foundation}.

Robotic foundation models, also known as vision-language-action (VLA) models, jointly process visual input, language commands, and directly output executable actions for task-conditioned control in an end-to-end manner \cite{brohan2023rt, black2024pi0visionlanguageactionflowmodel}. These models combine LLMs' ability to encode text and images with physical interactions learned from a robotic-specific dataset \cite{o2024open} to establish generalist robotic systems for manipulation.

\section{Deep Learning in Planning for Robotic Manipulators} \label{sec:dl-based_planning}
\tikzstyle{my-box}=[
    rectangle,
    draw=darkgray,
    rounded corners,
    text opacity=1,
    minimum height=1.5em,
    minimum width=5em,
    inner sep=2pt,
    align=center,
    fill opacity=.5,
    font=\small
]
\tikzstyle{leaf}=[my-box, minimum height=1.5em, 
    fill=white!80!gray, text=black,
    align=left,font=\small,
    inner xsep=2pt,
    inner ysep=4pt,
]

\begin{figure*}[htbp]
    \centering
    \resizebox{\textwidth}{!}
    {
        \begin{forest}
            forked edges,
            for tree={
                grow=east,
                reversed=true,
                anchor=base west,
                parent anchor=east,
                child anchor=west,
                base=left,
                font=\small,
                rectangle,
                draw=darkgray,
                rounded corners,
                align=left,
                minimum width=4em,
                edge={darkgray, line width=1pt},
                inner xsep=2pt,
                inner ysep=3pt},
                ver/.style={rotate=90, child anchor=north, parent anchor=south, anchor=center},
                where level=1{text width=5em,font=\small,}{},
                where level=2{text width=12em,font=\small,}{},
                where level=3{text width=12em,font=\small,}{},              
                where level=4{text width=12em,font=\small,}{},              
            [Deep Learning in Planning, ver, 
                [MLPs \cite{zhang2023dive}
                    [\textbf{End-to-end Planning} [{\textbf{End-to-end Path}: \cite{pandy2020learning}}, leaf, text width=20em]]
                    [\textbf{Sampling-based Primitives}
                        [{\textbf{Informed Sampling:} \cite{qureshi2019motion, qureshi2020motion, qureshideeply, tamizi2024end, parque2021learning, lai2022parallelised, bhardwaj2021leveraging, lyu2022motion}.}, leaf, text width=20em]
                        [{\textbf{Local Steering:}\cite{yu2024efficient,chiang2020fast, chiang2021fast, sugaya2021multitask}.}, leaf, text width=20em]
                    ]
                    [\textbf{Constrained Sampling-based} [{\textbf{Learning Constrained Manifolds: }\cite{qureshi2020neural, qureshi2021constrained,sutanto2021learning}.}, leaf, text width=20em]]
                    [\textbf{Trajectory Optimization} [{\textbf{Warm-starting Optimization Process:} \cite{ichnowski2020deep, kicki2023fast} \\ \cite{banerjee2020learning}.}, leaf, text width=20em]]
                    [\textbf{Collision Checking} [{\textbf{Distance-to-collision Estimator:} \cite{ichter2019robot, liu2025reliable, luo2023reinforcement, krawczyk2023comparison, tran2020predicting, guo2023deepcollide, rakita2018relaxedik, chase2020neural, liu2023whole}.}, leaf, text width=20em]]
                ]
                [CNNs \cite{yu2020robotic}
                    [\textbf{End-to-end Planning}[{\textbf{End-to-end Path}\cite{ota2021deep, ni2022ntfields, ni2023progressive, ni2024physics, ni2024physicsws, liu2024physics, niphysics}.}, leaf, text width=20em]]
                    [\textbf{Sampling-based Primitives}
                        [{\textbf{Informed Sampling:} \cite{shah2022using, patil2019prediction, abdi2023hybrid, chamzas2022learning, terasawa20203d}}, leaf, text width=20em]
                    ]
                    [\textbf{Collision Checking} [{\textbf{Binary Collision Estimator:}\cite{danielczuk2021object, murali2023cabinet}.}, leaf, text width=20em]]
                ]
                [PC-Nets \\ \cite{qi2017pointnet++}
                [\textbf{End-to-end Planning} [{\textbf{End-to-end Path:} \cite{fishman2023motion, fishmanavoid, dalal2024neural, yang2025deep, soleymanzadeh2025perfact}.}, leaf, text width=20em]]
                ]
                [RNNs \cite{lu2019learning}
                    [\textbf{End-to-end Planning}[{\textbf{End-to-end Path:} \cite{bency2019neural}}, leaf, text width=20em]]
                    [\textbf{Sampling-based primitives}
                        [{\textbf{Informed-sampling:} \cite{ying2021deep, hou2023data}}, leaf, text width=20em]
                    ]
                ]
                [GNNs \cite{battaglia2018relational}
                    [\textbf{Sampling-based Primitives}
                        [{\textbf{Informed-sampling:} \cite{soleymanzadeh2025simpnet, liu2024kg, liu2024integrating, yu2021reducing, huang2022hardware, zhang2022learning, zhang2023dyngmp}}, leaf, text width=20em]
                    ]
                    [\textbf{Collision Checking}[{\textbf{Distance-to-collision Estimator:} \cite{kim2022graphdistnet, song2023graph, kim2023pairwisenet}.}, leaf, text width=20em]]
                ]
                [DGMs
                    [Variational Auto-Encoders \\ (VAEs) \cite{kingma2013auto}
                        [\textbf{End-to-end Planning}[{\textbf{End-to-end Path:} \cite{hung2022reaching, yamada2023leveraging}.}, leaf, text width=20em]]
                        [\textbf{Sampling-based Primitives}
                            [{\textbf{Informed Sampling:} \cite{ichter2018learning, dastider2023damon, dastider2022sera, dastider2022reactive, dastider2024unified, ichter2019robot, kumar2019lego, jenamani2020robotic, gaebert2022learning, kobashi2023learning, lu2024neural, huang2024planning, johnson2023learning, johnson2023zero, xia2022graph}.}, leaf, text width=20em]
                        ]
                        [\textbf{Constrained Sampling-based} [{\textbf{Learning Constrained Manifolds:} \cite{ho2023lac, park2024constrained}}, leaf, text width=20em]]
                        [\textbf{Trajectory Optimization} [{\textbf{Warm-starting Optimization Process:} \cite{osa2022motion}.}, leaf, text width=20em]]
                    ]
                    [Generative Adversarial Network \\ (GAN) \cite{goodfellow2020generative}
                        [\textbf{Constrained Sampling-based}[{\textbf{Learning Constrained Manifolds:} \cite{lembono2020generative, lembono2021learning, acar2021approximating}.}, leaf, text width=20em]]
                        [\textbf{Trajectory Optimization} [{\textbf{Warm-starting Optimization Process:} \cite{ando2023learning}.}, leaf, text width=20em]]
                    ]
                    [Normalizing Flows (NFs) \cite{kingma2018glow}
                        [\textbf{Sampling-based Primitives}
                            [{\textbf{Informed Sampling:} \cite{lai2021plannerflows}.}, leaf, text width=20em]
                        ]
                    ]
                    [Energy-based Models \\ (EBMs) \cite{lecun2006tutorial}
                        [\textbf{Trajectory Optimization} [{\textbf{Warm-starting Optimization Process:} \cite{urain2022learning, zhi2023global}.}, leaf, text width=20em]]
                    ]
                    [Diffusion Models \cite{ho2020denoising}
                        [\textbf{Trajectory Optimization} [{\textbf{Planning as Inference:} \cite{huang2024diffusionseeder, carvalho2023motion, carvalho2024motion, li2024language, nikken2024denoising, yan2024m2diffuser, dastider2024apex, li2024efficient, li2024constraint, seo2024presto, power2023sampling, saha2023edmp, sharma2025cascaded, luo2024potential}.}, leaf, text width=20em]]
                    ]
                    [Flow Matching \cite{lipman2022flow}
                        [\textbf{Trajectory Optimization} [{\textbf{Planning as Inference:} \cite{nguyen2025flowmp, dai2025safe, tian2025warm}.}, leaf, text width=20em]]
                    ]
                ]
                [Transformers \\ \cite{vaswani2017attention}
                    [\textbf{Sampling-based Primitives}
                        [{\textbf{Informed Sampling:} \cite{chen2019learning, zhuang2024transformer}.}, leaf, text width=20em]
                    ]
                    [\textbf{Collision Checking}[{\textbf{Distance-to-collision Estimator:} \cite{cao2023distformer}.}, leaf, text width=20em]]
                ]
                [Foundation \\
                Models \cite{achiam2023gpt}
                    [\textbf{End-to-end Planning}[{\textbf{End-to-end Path:} \cite{mandi2024roco, kwon2024language, bucker2022reshaping, bucker2023latte}.}, leaf, text width=20em]]
                ]
                [Neural SDFs \\ \cite{park2019deepsdf}
                    [\textbf{Collision Checking}[{\textbf{Distance-to-collision Estimator:} \cite{koptev2022implicit, koptev2022neural, koptev2024reactive, liu2022regularized, liu2023collision, zhao2024perceptual, chenimplicit, quintero2024stochastic, li2024configuration, kim2024active, baxter2020deep, lee2022single, lee2024reliable, lee2023fast, michaux2023reachability, joho2024neural, kwon2024conformalized}.}, leaf, text width=20em]]
                ]
            ]
        \end{forest}
    }
    \caption{Deep learning application in manipulator planning.}
    \label{fig:sota-literature}
\end{figure*}
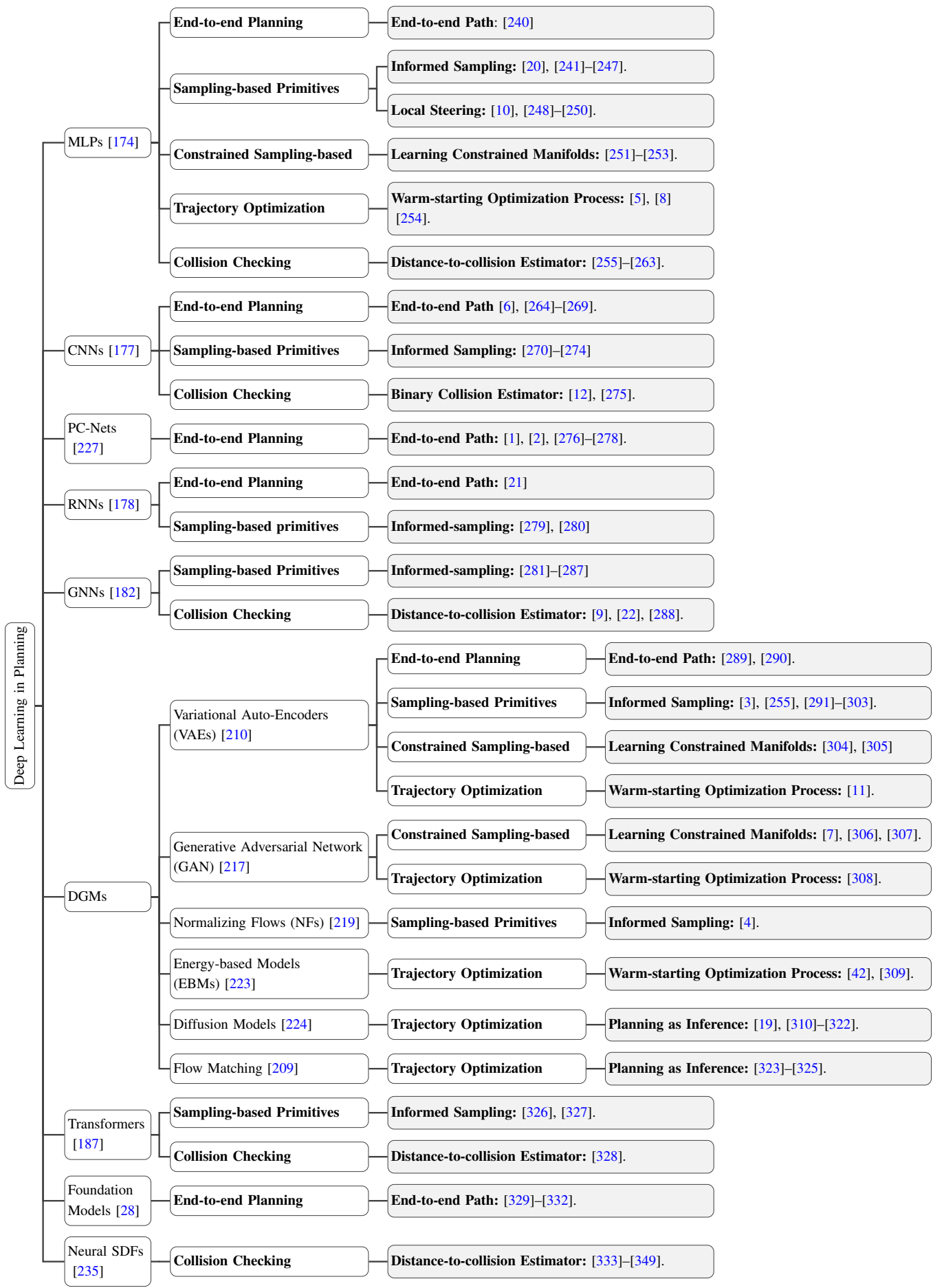

\begin{figure*}[htbp] 
\centering
\includegraphics[width=0.9\textwidth]{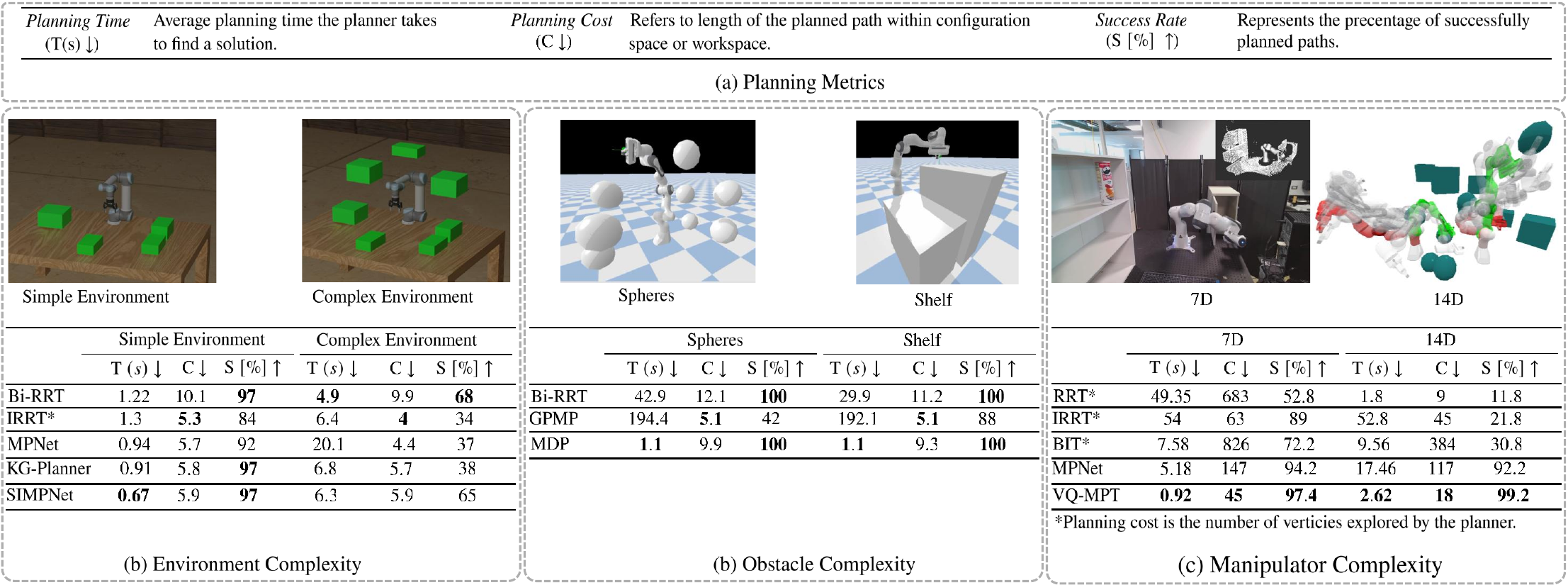}
\caption{A comparison of neural motion planners with benchmark planners across various planning complexities for out-of-distribution planning scenarios. (a) Planning metrics were used to quantify the performance of motion planners. (b) Environment complexity: SIMPNet \cite{soleymanzadeh2025simpnet} performance compared with benchmark planners (Bi-RRT \cite{kuffner2000rrt}, IRRT* \cite{gammell2014informed}, MPNet \cite{qureshi2019motion}, KG-Planner \cite{liu2024kg}) across different scenarios. (c) Degree of freedom: VQ-MPT \cite{johnson2023learning} performance compared with benchmark planners (RRT*, IRRT* \cite{gammell2014informed}, BIT* \cite{gammell2020batch}, MPNet \cite{qureshi2019motion}) across different scenarios. (d) Obstacle type: MPD \cite{carvalho2023motion} performance compared with benchmark planners (Bi-RRT \cite{kuffner2000rrt}, GPMP \cite{mukadam2016gaussian}) across different scenarios.}
\label{fig:planningexample}
\end{figure*}

\begin{figure*}[htbp] 
\centering
\includegraphics[width=\textwidth, trim={2cm, 1.8cm, 2cm, 1.5cm}, clip]{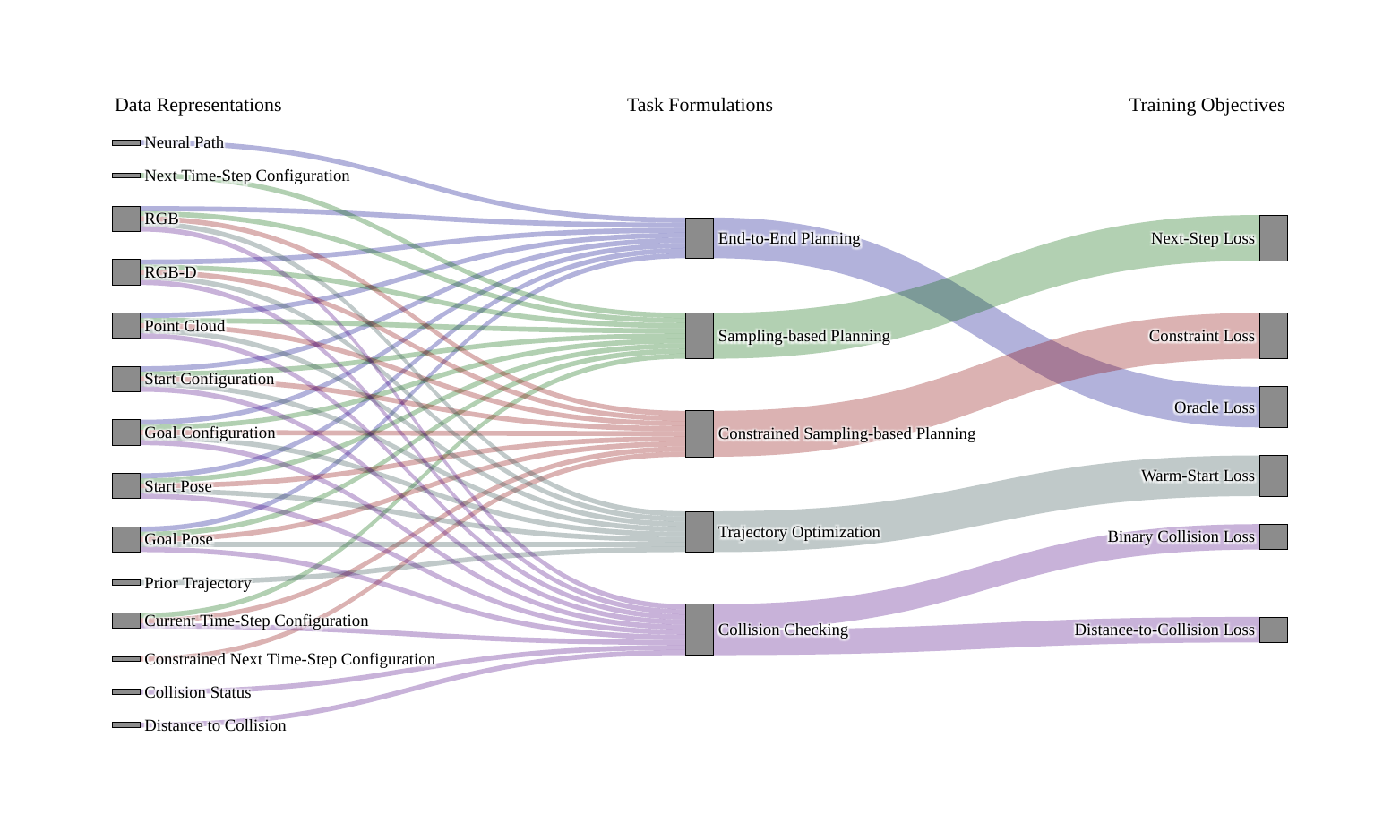}
\caption{ Data representation and training objectives of neural motion planners for robotic manipulators. The left column illustrates the input and output modalities (data representation) and the right column demonstrates the corresponding loss function (training objective) for training and deploying neural motion planners. Classical motion planners such as cuRobo \cite{sundaralingam2023curobo} and planners from OMPL \cite{sucan2012open}, along with collision checkers from FCL \cite{pan2012fcl}, are mainly used to generate oracle datasets for training neural motion planners.}
\label{fig:training}
\end{figure*}

In this section, we survey the state-of-the-art in utilizing deep learning methods for robotic manipulator motion planning. Given that robots work within similar settings and tackle similar motion planning problems, leveraging past planning experiences expedites the search for future plans. Figure \ref{fig:sota-literature} provide an overview of the state-of-the-art research that utilized deep learning frameworks for motion planning in robotic manipulators. Additionally, Figure \ref{fig:planningexample} presents the performance metrics of a neural motion planner compared to benchmark planners across varying levels of task complexity for a robotic manipulator. This section is organized based on the deep learning frameworks used to improve each component of classical planning algorithms. Figure \ref{fig:training} provides an overview of data representation, task representation, and training neural motion planners.

\subsection{End-to-end Planning} \label{subsec:DL-endtoend}
Deep learning frameworks have been utilized for end-to-end planning to capture the complex spatio-temporal dependencies and multi-modality inherent in motion planning for robotic manipulators. Table \ref{tab: endtoendref} overviews the state-of-the-art of utilizing deep learning for end-to-end planning, their contribution, and performance compared to benchmark planners.

\vspace{0.2cm}
\noindent
\textbf{Multi-Layer Perceptrons (MLPs)}: Deep neural networks (MLPs) have been utilized as universal approximations to learn the motion policy of robotic manipulators. Pandy \textit{et al.} \cite{pandy2020learning} employ an MLP consisting of a fully connected layer and ten highway layers \cite{srivastava2015highway} to learn the motion policy of the robotic manipulator. This framework processes a parameterized description of the workspace and generates an end-to-end trajectory. The network was trained using a customized cost (loss) function that aims to minimize the path length and avoid collision with workspace obstacles, such that the output waypoints satisfy the planning constraints.

\vspace{0.2cm}
\noindent
\textbf{Convolutional Neural Networks (CNNs):} Ota \textit{et al.} \cite{ota2021deep} also employ an MLP structure for short-horizon path waypoints generation. The proposed framework uses a CNN-based encoder-decoder framework to encode the depth information of the workspace and an MLP framework to generate waypoints for a downstream reactive motion generation algorithm. This module processes the depth image of the environment and the robot's state to output waypoints leading to the goal, complemented by a low-level reinforcement learning (RL)-based action controller for navigation between these waypoints.

Neural Time Fields (NTFields) \cite{ni2022ntfields}, a demonstration-free deep learning planner, utilizes a ResNet-style deep neural structure and 3D CNNs to generate the factorized time field from the start/goal configurations and workspace embeddings. Then, the speed model of the planning space is derived by solving the Eikonal equation \cite{raissi2019physics, smith2020eikonet} and used as the gradient step to move from the start configuration to the end configuration in a bi-directional manner. Progressive NTFields (P-NTFields) \textit{et al.} \cite{ni2023progressive} improve upon NTFields to address its generalizability and improve its success rate within clutter environments. The proposed framework adds a viscosity term (Laplacian of time field) to the Eikonal equation, which guarantees the smoothness of the predicted time field. Additionally, they deploy a progressive speed scheduling technique to address optimization difficulty near obstacles.

Constrained Neural Time Fields (C-NTFields) \cite{ni2024physics, ni2024physicsws} modifies the expert speed model of progressive NTFields to incorporate kinematic constraints into the estimated time fields. Task Space Regions (TSRs) \cite{berenson2011task} are leveraged to calculated the distance between robot configurations and constraint manifolds. NTFields, P-NTFields, and C-NTFileds train time fields estimator network offline and assume that the environment is known a priori. Active NTFields (A-NTFields) \cite{liu2024physics} relaxes this assumption by training the time field estimator network on the fly. The proposed framework processes the incoming sensor data to calculate the ground truth speed values, and utilizes an online learning framework for training the network.

Ni \textit{et al.} \cite{niphysics} improve NTFields' learning convergence by introducing temporal difference loss, normal alignment loss, and causality preservation alongside the original loss function. The framework also employs a PirateNet structure \cite{wang2024piratenets}, an MLP with residual gates, to estimate the time field while ensuring non-negativity, symmetry, and triangle inequality. Furthermore, it incorporates an attention mechanism to condition the field estimation network on workspace embeddings, enabling generalization to unseen workspaces.

\vspace{0.2cm}
\noindent
\textbf{Point Cloud Neural Networks (PC-Nets)}: Their ability to process unstructured 3D data and remain invariant to input permutations makes them well-suited for encoding workspace/configuration space with motion planning problems. Motion Policy Networks (M$\pi$Nets) \cite{fishman2023motion, fishmanavoid} features a point-cloud encoder that leverages PointNet++ \cite{qi2017pointnet++} to encode the input point-cloud, which consists of the manipulator's current geometry, scene geometry, and manipulator's target pose, effectively representing the workspace. Additionally, an MLP-based configuration encoder is used to capture the robot's current configuration, while another MLP serves as a decoder to generate the next planning way-point guiding the manipulator toward the goal configuration. 

Neural MP \cite{dalal2024neural} improves upon M$\pi$Nets by incorporating more realistic scenes for data generation. This framework also utilizes an LSTM-based configuration encoder and implements a stochastic learning framework based on Gaussian Mixture Models (GMMs) to address the multi-modality inherent in the planning dataset, and integrates a light-weight real-time optimization module to improve both the efficiency and success rate of the planning problem. Deep Reactive Policy (DRP) \cite{yang2025deep} extends Neural MP by integrating a point cloud encoder with an action-chunking transformer architecture for end-to-end motion planning. DPR further fine-tunes the neural planner using the Geometric Fabrics \cite{van2022geometric} framework and privileged workspace information in simulation to resolve minor collisions. In addition, it leverages point-cloud-aware Riemannian Motion Policies \cite{ratliff2018riemannian} to enable reactive planning in dynamic environments.

Motion Policy with LLM-Powered Dataset Synthesis and Fusion Action-Chunking Transformers (PerFACT) \cite{soleymanzadeh2025perfact} leverages the planning capabilities of LLMs \cite{achiam2023gpt} and procedural primitive generation \cite{dalal2024neural} to create a diverse, and semantically feasible set of workspaces for large-scale planning data collection. It then combines a point cloud encoder with a fusion action-chunking transformer architecture to intelligently attend to various planning sensing modalities for efficient planning.

\vspace{0.2cm}
\noindent
\textbf{Recurrent Neural Networks:} The recurrent connections within RNNs have been leveraged to learn the temporal dependencies within the motion planning problem. OracleNet \cite{bency2019neural} employs stacked LSTM layers to preserve the temporal information inherent in oracle paths. This framework generates waypoints directed toward the goal configuration by incorporating the goal configuration as an auxiliary input to the network.

\vspace{0.2cm}
\noindent
\textbf{Deep Generative Networks (DGNs)} are powerful frameworks for end-to-end motion planning. For instance, auto-encoders (AEs) are utilized for end-to-end planning for robotic manipulators.

\vspace{0.2cm}
\noindent
\textbf{DGNs - Auto Encoders (AE):} Latent Space Path Planning (LSPP) \cite{hung2022reaching} employs a VAE to encode both the configuration space (joint angles) and task space (end-effector positions) based on a dataset of randomly generated configurations. This specific method alleviates the need for explicit mapping between configuration space and task space. For optimization, Activation Maximization (AM) \cite{erhan2009visualizing} is used to refine the goal-reaching process by back-propagating the goal-reaching error and the output of a binary collision checker network to update the latent vector. This probabilistic approach allows the planner to exploit areas of the latent space with higher probabilities, enhancing efficiency. However, a major limitation is its requirement for predefined obstacle shapes for collision checking, which is impractical for real-world applications. To overcome this, Activation Maximization Planning in Latent Space (AMP-LS) \cite{yamada2023leveraging} integrates a neural collision predictor that operates directly on environment point clouds.

\vspace{0.2cm}
\noindent
\textbf{Transformers and Large Language Models:} The attention mechanism in transformers and LLMs enables them to capture the spatio-temporal dependencies inherent in motion planning problems. Roco \cite{mandi2024roco} utilizes an LLM agent (GPT-4) \cite{achiam2023gpt} to generate workspace waypoints for multi-robot collaboration scenarios. These waypoints are subsequently fed into a centralized RRT-based multi-arm motion planner for coordinated planning across all manipulators. The model can generate a new set of waypoints in each iteration by adjusting the temperature of the GPT-4 model, which controls the stochasticity of the language model.

Kwon \textit{et al.} \cite{kwon2024language} investigate the use of LLMs for zero-shot dense trajectory generation for robotic manipulators. They show that, with well-designed task-agnostic prompts, LLMs combined with off-the-shelf perception models can generate sequences of end-effector poses without relying on auxiliary low-level planning components \cite{vemprala2023chatgpt}. However, since these models are not trained on physical interaction data, they struggle in handling low-level planning for robotic manipulators. LATTE \cite{bucker2022reshaping, bucker2023latte} leverages pre-trained large language models to modify initial workspace manipulator trajectories given planning context. It takes in multiple data modalities - scene geometry, images, user language command, and initial trajectory -  and processes them through a transformer encoder-decoder structure to generate a modified trajectory. However, post-processing is applied to ensure the modified trajectory satisfies planning constraints since the output may not always be valid.

The main challenge in using LLMs for end-to-end manipulator motion planning is that the problem is inherently spatiotemporal rather than textual. As a result, LLM-based manipulation planning often relies on pre-defined motion primitives to carry out physical interactions within the environment and plan motions. This is due to the lack of large-scale robotic data  \cite{vemprala2023chatgpt, yu2023language,ahn2022can, liang2022code}.

\begin{table*}[htbp]
\centering
\caption{Overview of the state-of-the-art of utilizing deep learning for end-to-end planning, including the planning metrics reported relative to evaluated benchmark methods, and each approach's primary contribution to robotic manipulator motion planning.}
\label{tab: endtoendref}
\resizebox{\textwidth}{!}{%
\begin{tabular}{@{}P{0.1\textwidth}C{0.15\textwidth}C{0.15\textwidth}C{0.15\textwidth}P{0.45\textwidth}}
\arrayrulecolor{gray!80}
\hline
\rowcolor{gray!20}
\textbf{Paper}& \multicolumn{3}{c}{\textbf{Benchmarks \& Metrics}}& {\textbf{Contributions}}\\
\hline
\multicolumn{1}{c}{\phantom{Var.}}&\multicolumn{1}{c}{\cellcolor{gray!20}{T $[s]\downarrow$}}&\multicolumn{1}{c}{{\cellcolor{gray!20}C $\downarrow$}}&\multicolumn{1}{c}{\cellcolor{gray!20}S $[\%]\uparrow$}&\multicolumn{1}{c}{\phantom{Var.}} \\
\hline
\multirow{2}{*}{\begin{tabular}{l}
     Pandy \textit{et al.} \\
     \cite{pandy2020learning}
\end{tabular}}&$\mathbf{\underline{95}}~\text{ms}$&PL-W : $25.2$&$\mathbf{\underline{73}}$&\multirow{2}{*}{\begin{tabular}{l}
$\bullet$ A planning-specific cost (loss) function. \\
$\bullet$ MLPs plus highway layers \cite{srivastava2015highway} as neural motion planner. \\
\end{tabular}} \\
\cmidrule{2-4}
&$100$ ms (RRT*)&PL-W : $\mathbf{16.98}$ (RRT*)&$1.45$ (RRT*)& \\
\midrule 
Ota \textit{et al.} \cite{ota2021deep}&-&-&-& \begin{tabular}{l}
     $\bullet$ MLP-based neural waypoints generation for RL-based reactive \\
     planning.
\end{tabular} \\
\midrule
\multirow{2}{*}{\begin{tabular}{l}
     P-NTFields  \\
     \cite{ni2023progressive} 
\end{tabular}}&$\mathbf{\underline{0.03}}$&PL-C : $0.43$&$\mathbf{\underline{92}}$&\multirow{2}{*}{\begin{tabular}{l}
     $\bullet$ Utilizes ResNet-style deep network for configuration encoding. \\
     $\bullet$ Utilizes 3D CNNs for workspace encoding. 
\end{tabular}} \\
\cmidrule{2-4}
&$0.05$ (NTFields \cite{ni2022ntfields})&PL-C : $\mathbf{\underline{0.38}}$ (NTFields \cite{ni2022ntfields})&$84$ (NTFields \cite{ni2022ntfields})& \\
\midrule
\multirow{2}{*}{\begin{tabular}{l}
     C-NTFields  \\
     \cite{ni2024physics} 
\end{tabular}}&$\mathbf{\underline{0.05}}$&PL-C : $1.32$&$\mathbf{\underline{100}}$&\multirow{2}{*}{\begin{tabular}{l}
     $\bullet$ Utilizes TSR \cite{berenson2011task} to measure the distance to constraint \\
     manifolds.
\end{tabular}} \\
\cmidrule{2-4}
&$0.06$ (CBiRRT \cite{berenson2009manipulation})&PL-C : $\mathbf{\underline{1.30}}$ (CBiRRT \cite{berenson2009manipulation})&$\mathbf{\underline{100}}$ (CBiRRT \cite{berenson2009manipulation})& \\
\midrule
\multirow{2}{*}{\begin{tabular}{l}
     A-NTFields  \\
     \cite{liu2024physics} 
\end{tabular}}&$\mathbf{\underline{0.03}}$&PL-C : $\mathbf{\underline{2.25}}$&$\mathbf{\underline{91}}$&\multirow{2}{*}{\begin{tabular}{l}
     $\bullet$ Calculates ground truth speed values on the fly. \\
     $\bullet$ Estimates the time field on the fly.
\end{tabular}} \\
\cmidrule{2-4}
&$1.36$ (LazyPRM)&PL-C : $3.05$ (LazyPRM)&$87$ (LazyPRM)& \\
\midrule
\multirow{2}{*}{\begin{tabular}{l}
     Ni \textit{et al.}  \\
     \cite{niphysics} 
\end{tabular}}&$0.074$&PL-C : $1.95$&$\mathbf{\underline{87}}$&\multirow{2}{*}{\begin{tabular}{l}
     $\bullet$ Utilizes PirateNet \cite{wang2024piratenets} for the time field estimation. \\
     $\bullet$ Leverage the attention mechanism to achieve generalizability.
\end{tabular}} \\
\cmidrule{2-4}
&$\mathbf{\underline{0.063}}$ (NTFields \cite{ni2022ntfields})&PL-C : $\mathbf{\underline{1.63}}$ (NTFields \cite{ni2022ntfields})&$74$ (NTFields \cite{ni2022ntfields})& \\
\midrule
\multirow{2}{*}{OracleNet \cite{bency2019neural}}&$\mathbf{\underline{1.24}}$&PL-C : $\mathbf{\underline{0.85}}$&-&\multirow{2}{*}{\begin{tabular}{l}
     $\bullet$ Utilizes RNNs to iteratively generated end-to-end paths. \\
     $\bullet$ LSTM for encoding temporal dependencies. 
\end{tabular}} \\
\cmidrule{2-4}
&$29.32$ (RRT*)&PL-C : $1$ (RRT*)&-& \\
\midrule
\multirow{2}{*}{M$\pi$Net \cite{fishman2023motion}}&$\mathbf{\underline{0.33}}$&-&$82.78$&\multirow{2}{*}{\begin{tabular}{l}
     $\bullet$ PointNet++ \cite{qi2017pointnet} as workspace and planning space encoder. \\
     $\bullet$ Geometric, task-space loss for training. 
\end{tabular}} \\
\cmidrule{2-4}
&$16.46$ (AIT*)&-&$\mathbf{\underline{100}}$ (AIT*)& \\
\midrule
\multirow{2}{*}{Neural MP \cite{dalal2024neural}}&-&-&$\mathbf{\underline{95.83}}$&\multirow{2}{*}{\begin{tabular}{l}
     $\bullet$ PointNet++ \cite{qi2017pointnet} as workspace and planning space encoder. \\
     $\bullet$ Encoding multi-modality via GMMs. 
\end{tabular}} \\
\cmidrule{2-4}
&-&-&$16.67$ (M$\pi$Net \cite{fishman2023motion})& \\
\midrule
\multirow{2}{*}{DRP \cite{yang2025deep}}&-&-&$\mathbf{\underline{84.60}}$&\multirow{2}{*}{\begin{tabular}{l}
     $\bullet$ Geometric Fabrics \cite{van2022geometric} for fine-tuning and RMPs \cite{ratliff2018riemannian} \\ for reactive planning. 
\end{tabular}} \\
\cmidrule{2-4}
&-&-&$50.59$ (Neural MP \cite{dalal2024neural})& \\
\midrule
\multirow{2}{*}{PerFACT \cite{soleymanzadeh2025perfact}}&-&-&$\mathbf{\underline{51.2}}$&\multirow{2}{*}{\begin{tabular}{l}
     $\bullet$ LLMs \cite{achiam2023gpt} for large-scale dataset generation. \\
     $\bullet$ Planning modality-aware end-to-end planning.
\end{tabular}} \\
\cmidrule{2-4}
&-&-&$14.6$ (Neural MP \cite{dalal2024neural})& \\
\midrule
\multirow{2}{*}{LSPP \cite{hung2022reaching}}&$179$ ms&PL-W: $\mathbf{\underline{1.52}}$&$\mathbf{\underline{85}}$&\multirow{2}{*}{\begin{tabular}{l}
     $\bullet$ Planning with a latent space of a VAE. \\
     $\bullet$ Using activation maximization \cite{erhan2009visualizing} for updating latent vectors. 
\end{tabular}} \\
\cmidrule{2-4}
&$\mathbf{\underline{128}}$ ms (BiRRT)&PL-W: $2.33$ (BiRRT)&$84$ (BiRRT)& \\
\midrule
Roco \cite{mandi2024roco}&-&-&-&\begin{tabular}{l}
     $\bullet$ GPT-4 \cite{achiam2023gpt} for generating workspace intermediate goal states for \\
     multi-arm motion planning.
\end{tabular}\\
\midrule
Kwon \textit{et al.} \cite{kwon2024language}&-&-&-& \begin{tabular}{l}
    $\bullet$ LLMs for zero-shot planning for robotic manipulators.
\end{tabular}\\
\midrule
LATTE \cite{bucker2023latte}&-&-&-& \begin{tabular}{l}
    $\bullet$ Reshaping trajectories via language commands.
\end{tabular}\\
\bottomrule
\end{tabular}}
\begin{minipage}{\textwidth}
\smallskip
\footnotesize
\textbf{Note:} ``$T$'' denotes \textit{planning time}, ``$C$'' denotes \textit{planning cost}, and ``$S$'' refers to \textit{success rate} (Section \ref{subsec:planning_def-metrics}). ``$\downarrow$'' indicates lower is better, and ``$\uparrow$'' indicates higher is better. ``PL-W'' refers to path length measured in the workspace, and ``PL-C'' denotes path length in the configuration space.
\end{minipage}
\end{table*}  
\subsection{Improving Unconstrained Sampling-based Planners' Algorithmic Primitives} \label{subsec:DL-sampling-based}
This section reviews how deep learning methods have improved the performance of these types of planners. As mentioned in Section \ref{subsec: classic-sampling-based}, sampling-based planning algorithms are built upon three algorithmic primitives: sampling, steering, and collision checking. Deep learning frameworks can be employed to improve each of these primitives to increase both the efficiency and success rate of the algorithms.  Table \ref{tab: dl-sampling} overviews deep learning application in unconstrained sampling-based planning algorithms. Since collision checking is a common component across all planning algorithms, the application of deep learning frameworks for collision checking will be discussed in detail in Section \ref{subsec:DL-collisionchecking}.

\subsubsection{Sampling Primitive} \label{subsubsec: DL-sampling-based-primitive}
Deep learning frameworks can help to address the low convergence rate of classic sampling-based motion planners by leveraging historical data to generate collision-free informed samples within complex planning workspaces and narrow passages. Table \ref{tab: informedsamplingref} overviews the state-of-the-art of utilizing deep learning for informed sampling, their contribution, and performance compared to benchmark planners.

\vspace{0.2cm}
\noindent
\textbf{Multi-Layer Perceptrons:} Deep neural networks can encode the underlying distribution of planning datasets, making them well-suited for learning the underlying sampling distributions for informed sampling within sampling-based planning algorithms. Deep learning frameworks have been used as a sampling distribution to explicitly generate informed samples in place of the original sampling primitive of sampling-based planners. Motion Planning Networks (MPNet) \cite{qureshi2019motion, qureshi2020motion, qureshideeply} incorporates an MLP-based framework to serve as the sampling primitive within the structure of sampling-based planning algorithms. These frameworks maintain an active Dropout layer \cite{srivastava2014dropout} during inference (sampling), which introduces stochastic behavior in the MLP-based sampler for encoding the multi-modality within the planning problem. Contractive autoencoders (CAEs) \cite{rifai2011contractive} are used for workspace embedding, ensuring robust representation. Additionally, Path Planning and Collision Checking Network (PPCNet) \cite{tamizi2024end} consists of two neural networks: a planning network and a collision checking network for repetitive bin-picking tasks. The planning network processes the current and goal configuration to output the next time-step configuration, while the collision-checking network checks the validity of steering toward this new configuration.

MLP models have also been employed to implicitly encode and learn the planning space for informed sample generation. Parque \textit{et al.} \cite{parque2021learning} employed MLP networks to learn the linear motion planning functions of robotic manipulators. These deep networks map a given pair of start and goal configurations of the planning problem to parameters that encode the linear transition of robot joints within the configuration space. Parallelized Diffeomorphic Sampling-based Motion Planning (PDMP) \cite{lai2022parallelised} combines the benefits of sampling-based and trajectory optimization planning methods. This framework utilizes an MLP structure to implicitly learn the occupancy probability distribution of the workspace and transform it through a differentiable bijection to configuration space for informed sampling. 

Bhardwaj \textit{et al.} \cite{bhardwaj2021leveraging} formulate edge selection in lazy graph search algorithms as a Markov Decision Process (MDP) and apply imitation learning to train a deep neural network. This network is designed to map the features of each edge to the probability of its validity, enhancing the efficiency of motion planning where edge collision checking is computationally expensive. Points-Guided Sampling Net (PGSN) \cite{lyu2022motion} employs a VAE framework to encode the workspace point cloud into a linearly interpolatable latent space to enhance generalizability to unseen environments. This framework employs an MLP-based multi-modal sampling net to switch between different mode of the planning problem to encode the inherent multi-modality.

\begin{table*}
\centering
\caption{State-of-the-art literature on utilizing various deep learning frameworks to improve various components of sampling-based motion planning algorithms for robotic manipulator motion planning.}
\label{tab: dl-sampling}
\begin{tabular}{@{}P{0.08\textwidth}C{0.2\textwidth}C{0.04\textwidth}C{0.04\textwidth}C{0.04\textwidth}C{0.04\textwidth}C{0.13\textwidth}C{0.09\textwidth}C{0.13\textwidth}}
\arrayrulecolor{gray!80}
\hline
\rowcolor{gray!20} \textbf{Primitive}&\textbf{Papers}&\textbf{MLPs}&\textbf{CNNs}&\textbf{RNNs}&\textbf{GNNs}&\textbf{Generative Models}&\textbf{Transformers}&\textbf{Foundation Models}\\
\hline
\multirow{6}{*}{Sampling}& \cite{qureshi2019motion, qureshi2020motion, qureshideeply, tamizi2024end, parque2021learning, lai2022parallelised, bhardwaj2021leveraging, lyu2022motion}&\cmark&\xmark&\xmark&\xmark&\xmark&\xmark&\xmark \\
    \cmidrule{2-9}
    &\cite{shah2022using, patil2019prediction, abdi2023hybrid, chamzas2022learning, terasawa20203d}&\xmark&\cmark&\xmark&\xmark&\xmark&\xmark&\xmark \\
    \cmidrule{2-9}
    &\cite{ying2021deep, hou2023data}&\xmark&\xmark&\cmark&\xmark&\xmark&\xmark&\xmark \\
    \cmidrule{2-9}
    &\cite{soleymanzadeh2025simpnet, liu2024kg, liu2024integrating, yu2021reducing, huang2022hardware, zhang2022learning, zhang2023dyngmp}&\xmark&\xmark&\xmark&\cmark&\xmark&\xmark&\xmark \\
    \cmidrule{2-9}
    &\cite{ichter2018learning, dastider2023damon, dastider2022sera, dastider2022reactive, dastider2024unified, ichter2019robot, kumar2019lego, jenamani2020robotic, gaebert2022learning, kobashi2023learning, lu2024neural, huang2024planning, johnson2023learning, johnson2023zero, xia2022graph, lai2021plannerflows}&\xmark&\xmark&\xmark&\xmark&\cmark&\xmark&\xmark \\
    \cmidrule{2-9}
    &\cite{chen2019learning, zhuang2024transformer}&\xmark&\xmark&\xmark&\xmark&\xmark&\cmark&\xmark \\
\midrule
Steering&\cite{yu2024efficient, chiang2020fast, chiang2021fast, sugaya2021multitask}&\cmark&\xmark&\xmark&\xmark&\xmark&\xmark&\xmark \\
\bottomrule
\end{tabular}
\begin{minipage}{0.9\textwidth}
\smallskip
\footnotesize
\textbf{Note:} ``MLPs'' denotes multi-layer perceptrons, ``CNNs'' denotes convolutional neural networks, ``RNNs'' denotes recurrent neural networks, and ``GNNs'' denotes graph neural networks.
\end{minipage}
\end{table*}

\vspace{0.2cm}
\noindent
\textbf{Convolutional Neural Networks (CNNs):} Transition invariance and locality properties of CNNs allow encoding the inherent distribution within planning data for informed sample generation, and effective workspace encoding. These models account for spatial information and can capture both local and global structures. Hierarchical Abstraction guided Robot Planner (HARP) \cite{shah2022using} employs a U-net \cite{ronneberger2015u} architecture to identify critical regions (abstractions) for end-effector and other DOFs that are not defined by the end-effector's position. Then, it utilizes a customized high-level planner to generate high-level plans that are further refined by a low-level planner for trajectory generation between start and goal configurations. Similarly, Bottleneck guided RRT$^*$ \cite{patil2019prediction} utilizes a 3D CNN for encoding the bottleneck regions (narrow passages) in the workspace, biasing the sampling of the sampling-based planners towards these regions. Abdi \textit{et al.} \cite{abdi2023hybrid} leverage an object detection model (YOLO \cite{redmon2016you}) coupled with a deep neural network to get the coordinate of objects within the workspace. Then, a graph search algorithm is leveraged over a grid in the workspace to find a path between the start and goal poses.

FIRE (Fast retrIeval of Relevant Experiences) \cite{chamzas2022learning} retrieves relevant location representations from past planning instances to guide the planning problem. This method utilizes a self-supervising method to find pairs of similar-dissimilar location representations. Then, a similarity function through the latent space of Siamese network \cite{chicco2021siamese} - a CNN-based neural network architecture with two identical encoders - is trained to retrieve relevant experiences to guide the planning problem and ensure generalizability to out-of-distribution problems. Heuristic Map Network (HMNet) \cite{terasawa20203d} employs a CNN-based framework to embed and encode workspace information, to learn a heuristic map (cost-to-go) for guiding the sampling process of the planning algorithm to perform guided exploration towards the planning goal.

\vspace{0.2cm}
\noindent
\textbf{Recurrent Neural Networks (RNNs):} recognized for their capability to encode inherent temporal sequences, have been effectively employed to enhance the sampling module within sampling-based planners and to encode dynamic dependencies in dynamic task spaces. The recurrent connection within RNNs effectively encodes spatial and temporal relationships within planning datasets. LSTM-BiRRT \cite{ying2021deep} incorporates an LSTM sampler to guide the BiRRT algorithm toward the goal configuration by leveraging past planning experiences in dual-arm planning scenarios. Similarly, Hou \textit{et al.} \cite{hou2023data} utilizes an RNN-based encoder to capture the temporal dependencies within dynamic/static environments. The proposed framework also implements a deep neural network to learn the feasible solution space from past experiences, enhancing sampling efficiency.

\vspace{0.2cm}
\noindent
\textbf{Graph Neural Networks (GNNs):} The capability of GNNs to accurately characterize and learn the structure of Euclidean and non-Euclidean data makes them highly effective for encoding the planning space and biasing the sampling distribution sampling-based planning algorithms towards the planning goal. Spatial-Informed Motion Planning Network (SIMPNet) \cite{soleymanzadeh2025simpnet} and KG-Planner \cite{liu2024kg}, \cite{liu2024integrating} construct graphs representing the kinematic structure of the robotic manipulator. Then, a graph neural network is trained on the constructed graph to generate informed, kinematic aware samples towards the planning goal. Yu \textit{et al.} \cite{yu2021reducing, huang2022hardware} accelerated classical sampling-based motion planners by utilizing a trained GNN for path exploration and path smoothing. This framework generates a random graph (RGG) by randomly sampling within the configuration space and including start and goal configurations. Then, the GNN determines the validity of RGG's edges to reduce collision queries. This framework also incorporates a cross-attention mechanism to integrate the obstacle embeddings from the workspace into the configuration space for environment-aware motion planning. 

Zhang \textit{et al.} \cite{zhang2022learning} utilizes a GNN framework to encode the spatio-temporal structure within dynamic motion planning. The network uses an RGG and an attention-based temporal encoding of the dynamic obstacles to determine which edges to steer to minimize unnecessary collision queries. In this framework, the trajectory of the dynamic obstacles needs to be known a priori, which is not necessarily the case in real-world examples. In subsequent work, Zhang \textit{et al.} \cite{zhang2023dyngmp} introduces DynGMP to mitigate the need for prior knowledge of the movement of the dynamic obstacles and plan in unpredictable dynamic environments. In this framework, a GNN trained on an RGG determines edge priority. DynGMP preserves collision-free parts of the initially constructed tree for reuse in subsequent tree constructions, performing collision-checking on nodes and edges that intersect with the geometries of the obstacles and the robot to reduce the computational overhead of the replanning sub-module.

\vspace{0.2cm}
\noindent
\textbf{Deep generative networks (DGNs)} are powerful deep learning frameworks utilized to capture the multimodal nature within planning datasets. DGN models such as auto-encoders (AEs), generative adversarial networks (GANs), and normalizing flows (NFs) are commonly used to enhance the sampling primitive of sampling-based planning algorithms.

\begin{table*}[htbp]
\centering
\caption{Overview of the state-of-the-art of utilizing deep learning frameworks for informed sampling, including the planning metrics reported relative to evaluated benchmark methods, and each approach's primary contribution to robotic manipulator motion planning.}
\label{tab: informedsamplingref}
\resizebox{\textwidth}{!}{%
\begin{tabular}{@{}P{0.1\textwidth}C{0.15\textwidth}C{0.15\textwidth}C{0.15\textwidth}P{0.45\textwidth}}
\arrayrulecolor{gray!80}
\hline
\rowcolor{gray!20}
\textbf{Paper}& \multicolumn{3}{c}{\textbf{Benchmarks \& Metrics}}& {\textbf{Contributions}}\\
\hline
\multicolumn{1}{c}{\phantom{Var.}}&\multicolumn{1}{c}{\cellcolor{gray!20}{T $[s]\downarrow$}}&\multicolumn{1}{c}{{\cellcolor{gray!20}C $\downarrow$}}&\multicolumn{1}{c}{\cellcolor{gray!20}S $[\%]\uparrow$}&\multicolumn{1}{c}{\phantom{Var.}} \\
\hline
\multirow{2}{*}{MPNet \cite{qureshi2020motion}}&$\mathbf{\underline{0.81}}$&PL-C : $\mathbf{\underline{6.98}}$&$\mathbf{\underline{78.6}}$&\multirow{2}{*}{\begin{tabular}{l}
    $\bullet$ MLP framework as informed sampler. \\
    $\bullet$ Dropout \cite{srivastava2014dropout} for encoding planning multi-modality.
\end{tabular}} \\
\cmidrule{2-4}
&$9.2$ (BIT*)&$10.78$ (BIT*)&$56$ (BIT*)& \\
\hline
\multirow{2}{*}{PCCNet \cite{tamizi2024end}}&$\mathbf{\underline{0.1}}$&PL-C : $\mathbf{\underline{5.88}}$&$90$&\multirow{2}{*}{\begin{tabular}{l}
    $\bullet$ MLP-based informed sampler and collision-checker for \\
    bin-picking tasks.
\end{tabular}} \\
\cmidrule{2-4}
&$0.38$ (MPNet \cite{qureshi2020motion})&$7.3$ (MPNet \cite{qureshi2020motion})&$\mathbf{\underline{94.4}}$ (MPNet \cite{qureshi2020motion})&  \\
\midrule
\multirow{2}{*}{\begin{tabular}{l}
     Hou \textit{et al.}  \\
     \cite{hou2023data}
\end{tabular}}&$\mathbf{\underline{0.6}}$&-&-&\multirow{2}{*}{\begin{tabular}{l}
    $\bullet$ MLP-based informed sampling distribution.
\end{tabular}} \\
\cmidrule{2-4}
&$0.81$ (RRT)&-&-& \\
\midrule
\multirow{2}{*}{PDMP \cite{lai2022parallelised}}&$\mathbf{\underline{5.99}}$&-&$\mathbf{\underline{96}}$&\multirow{2}{*}{\begin{tabular}{l}
    $\bullet$ MLP to learn occupancy distribution. \\
    $\bullet$ Transforms distribution to configuration space for sampling.
\end{tabular}} \\
\cmidrule{2-4}
&$13.83$ (RRT*)&-&$43$ (RRT*)& \\
\midrule
\multirow{2}{*}{HARP \cite{shah2022using}}&-&-&$\mathbf{\underline{100}}$&\multirow{2}{*}{\begin{tabular}{l}
    $\bullet$ U-net for critical region identification for the end effector \\
    as well as other DOFs.
\end{tabular}} \\
\cmidrule{2-4}
&-&-&$\sim 50$ (BiRRT)& \\
\midrule
\multirow{2}{*}{HMNet \cite{terasawa20203d}}&$\mathbf{\underline{149}}$ ms&PL-W : $\mathbf{\underline{0.99}}$&$\mathbf{\underline{100}}$&\multirow{2}{*}{\begin{tabular}{l}
    $\bullet$ Learning heuristic map through 3D CNN.
\end{tabular}} \\
\cmidrule{2-4}
&$2466$ ms (BiRRT)&$3.8$ (BiRRT)&$76$ (BiRRT)& \\
\midrule
\multirow{2}{*}{SIMPNet \cite{terasawa20203d}}&$\mathbf{\underline{6.3}}$&PL-C : $5.9$&$\mathbf{\underline{65}}$&\multirow{2}{*}{\begin{tabular}{l}
    $\bullet$ GNNs for kinematic-aware informed sampling. \\
    $\bullet$ Cross-attention mechanism for workspace-aware sampling. 
\end{tabular}} \\
\cmidrule{2-4}
&$20$ (MPNet \cite{qureshi2020motion})&$\mathbf{\underline{4.4}}$ (MPNet \cite{qureshi2020motion})&$37$ (MPNet \cite{qureshi2020motion})& \\
\midrule
\multirow{2}{*}{SERA \cite{dastider2022sera}}&$\mathbf{\underline{9.1}}$&-&$\mathbf{\underline{95.3}}$&\multirow{2}{*}{\begin{tabular}{l}
    $\bullet$ VAE for mapping the planning space into a low-dimensional \\
    latent space.
\end{tabular}} \\
\cmidrule{2-4}
&$22.6$ (MPNet \cite{qureshi2020motion})&-&$82.5$ (MPNet \cite{qureshi2020motion})& \\
\midrule
\multirow{2}{*}{VQ-MPT \cite{johnson2023learning}}&$\mathbf{\underline{0.9}}$&\#V: $\mathbf{\underline{45}}$&$\mathbf{\underline{97}}$&\multirow{2}{*}{\begin{tabular}{l}
    $\bullet$ VQ-VAEs \cite{van2017neural} for encoding feasible sampling regions. \\
    $\bullet$ Transformer-based sampling within feasible regions.
\end{tabular}} \\
\cmidrule{2-4}
&$5.18$ (MPNet \cite{qureshi2020motion})&\#V: $147$ (MPNet \cite{qureshi2020motion})&$94$ (MPNet \cite{qureshi2020motion})& \\
\midrule
\multirow{2}{*}{G-WAE \cite{xia2022graph}}&$\mathbf{\underline{4.36}}$&PL-W : $\mathbf{\underline{1.86}}$&$\mathbf{\underline{80}}$&\multirow{2}{*}{\begin{tabular}{l}
    $\bullet$ WAEs \cite{tolstikhin2017wasserstein} for encoding free configuration space. \\
    $\bullet$ GGNNs \cite{li2015gated} for spatial-aware encoding.
\end{tabular}} \\
\cmidrule{2-4}
&$9.74$ (RRT*)&PL-W : $2.21$ (RRT*)&$37$ (RRT*)& \\
\midrule
\multirow{2}{*}{P-Flows \cite{lai2021plannerflows}}&-&\#V: $\mathbf{\underline{11.6 \times 10^3}}$&-&\multirow{2}{*}{\begin{tabular}{l}
    $\bullet$ Utilizes normalizing flows to learn sampling distribution.
\end{tabular}} \\
\cmidrule{2-4}
&-&\#V : $32.7 \times 10^3$ (MPNet \cite{qureshi2020motion})&-& \\
\midrule
\multirow{2}{*}{NEXT \cite{chen2019learning}}&-&PL-C : $\mathbf{\underline{17.5}}$&$\mathbf{\underline{70.6}}$&\multirow{2}{*}{\begin{tabular}{l}
    $\bullet$ Leverages attention mechanism to encode configuration space \\
    into a discrete latent space.
\end{tabular}} \\
\cmidrule{2-4}
&-&PL -C : $29.6$ (BIT*)&$47.5$ (BIT*)& \\
\midrule
\multirow{2}{*}{TEMP \cite{chen2019learning}}&$\mathbf{\underline{0.741}}$&\#V : $\mathbf{\underline{195}}$&-&\multirow{2}{*}{\begin{tabular}{l}
    $\bullet$ Utilizes attention mechanism for informed sample generation.
\end{tabular}} \\
\cmidrule{2-4}
&$17.87$ (RRT*)&\#V : $1486$ (RRT*)&-& \\
\bottomrule
\end{tabular}}
\begin{minipage}{\textwidth}
\smallskip
\footnotesize
\textbf{Note:} ``$T$'' denotes \textit{planning time}, ``$C$'' denotes \textit{planning cost}, and ``$S$'' refers to \textit{success rate} (Section \ref{subsec:planning_def-metrics}). ``$\downarrow$'' indicates lower is better, and ``$\uparrow$'' indicates higher is better. ``PL-W'' refers to path length measured in the workspace, ``PL-C'' denotes path length in the configuration space, and ``\#V'' indicates the number of vertices the planner explores to plan.
\end{minipage}
\end{table*}

\vspace{0.2cm}
\noindent
\textbf{DGNs - Auto-Encoders (AEs):} AEs are effective for learning sampling distributions in sampling-based planning algorithms by encoding the dataset into a latent space. Ichter \textit{et al.} \cite{ichter2018learning} employs a conditional variational autoencoder (CVAE) conditioned on planning information (i.e., obstacles, start and goal configurations) to bias the sampling process to promising regions of configuration space. This framework learns the underlying structure within the planning dataset generated by an oracle planner and generates informed samples within the framework of sampling-based planning algorithms. Dastider \textit{et al} \cite{dastider2023damon, dastider2022sera, dastider2022reactive} introduced a motion planning framework that transforms the high-dimensional configuration space into a low-dimensional latent space using a CVAE framework to generate adaptable motion policies. Then, a graph search algorithm is deployed to plan in the learned low-dimensional latent space. They further enhance their approach by implementing a diffusion variational autoencoder (D-VAE) to encode and reduce the dimensionality of the planning space for motion planning in a low-dimensional space \cite{dastider2024unified}. The D-VAE encodes the motion dynamics of the robotic manipulator and moving obstacles into a unified framework. This approach utilizes an extended Kalman filter (EKF) \cite{khodarahmi2023review} to predict the movement of dynamic obstacles.

Latent Sampling-Based Motion Planning (L-SBMP) \cite{ichter2019robot} employs an encoder-decoder framework to encode high-dimensional planning spaces into a low-dimensional latent space and performs sampling-based motion planning within it. This allows sampling, steering, and collision checking to be conducted more efficiently within the low-dimensional latent space, thus enhancing the efficiency and reducing the computational complexity of motion planning for high DOF robotic manipulators. Leveraging Experience with Graph Oracles (LEGO) \cite{kumar2019lego} trains a CVAE on an RGG to generate a roadmap which contains bottleneck nodes within the planning space. The work \cite{jenamani2020robotic}, employed LEGO-CVAE \cite{kumar2019lego} to identify and learn the critical bottleneck regions within the planning space, generating informed samples that serve as roots for RRT motion planning algorithm. 

Gaebert \textit{et al.} \cite{gaebert2022learning} employed a CVAE to encode the planning space and generate informed samples conditioned on the motion planning problem (start and goal configuration and workspace encoding). The CVAE-based sampler is incorporated into any sampling-based planning algorithm for informed sample generation. Kobashi \textit{et al.} \cite{kobashi2023learning} proposed decomposing the task space of the motion planning problem to handle the high variance implicit within such problems. This framework utilizes Binary Space Partitioning (BSP) \cite{fuchs1980visible} to partition the workplace and identify challenging regions and assign nodes to them. Then, the assigned nodes are used to train a neural network for generating nodes for new workspaces. And, a CVAE - conditioned on environment embeddings - is utilized to generate samples around key nodes for planning.

The Neural Randomized Planner (NRP) \cite{lu2024neural} trains a discriminative MLP and a generative CVAE neural local sampler to learn local sampling distributions. Then, these local samplers are used within global sampling-based motion planners for motion planning. Planning with Learned Subgoals (PLS) \cite{huang2024planning} incorporates the planning problem's temporal information to account for temporal constraints. This model feeds the temporal and spatial information of the planning problem to a CVAE to generate sub-goal configurations for planning, effectively handles time-constrained, reactive motion planning scenarios in dynamic environments.

Vector Quantized-Motion Planning Transformers (VQ-MPT) \cite{johnson2023learning, johnson2023zero} utilizes Vector Quantized-Variational Autoencoders (VQ-VAEs) \cite{van2017neural} to learn and encode the configuration space, aiming to reduce its dimensionality and identify areas likely to contain feasible paths. VQ-VAEs help to avoid the issue of posterior collapse, a common problem in traditional variational autoencoders, making them particularly effective for encoding high-dimensional configuration spaces. VQ-MPT also incorporates a cross-attention mechanism to condition the planning problem on workspace embedding, and start and goal configurations. Additionally, an auto-regressive transformer is employed to capture long-horizon correlations and sample within promising regions within the configuration space. This facilitates informed sampling in downstream sampling-based motion planning algorithms, enhancing the efficiency of these planners. 

Graph Wasserstein Autoencoder (GraphWAE) \cite{xia2022graph}, utilizes a variant of the Wasserstein Autoencoder (WAE) \cite{tolstikhin2017wasserstein} with Gated Graph Neural Networks (GGNNs) \cite{li2015gated} as encoder and decoder, to encode the collision-free region within the configuration space. The WAEs, compared to VAEs, offer enhanced stability during training and demonstrate sufficient representation capabilities, which are crucial for encoding high-dimensional and complex configuration spaces. This framework is trained on a successful path dataset, and the decoder (graph generative model) is leveraged as a neural informed sampler for a downstream sampling-based planning algorithm.

\vspace{0.2cm}
\noindent
\textbf{DGNs - Normalizing flows (NFs):} NFs have also been employed to effectively encode the configuration space for informed sampling within sampling-based planning algorithms. A distinctive property of NF is that they don't experience mode collapse, thanks to the inclusion of the mode collapse loss in the optimization problem. PlannerFlows \cite{lai2021plannerflows} learns the sampling distribution of sampling-based planning algorithms through normalizing flows. This framework is conditioned on planning information (i.e., workspace embedding, start and goal configurations) to generate informed samples towards the planning goal.

\vspace{0.2cm}
\noindent
\textbf{Transformers:} Transformers and attention mechanisms have been leveraged to encode the long-horizon spatiotemporal dependency within planning problems. Neural Exploration-Exploitation Tree (NEXT) algorithm \cite{chen2019learning} utilizes an attention-based network to embed the configuration space into a discrete latent representation. Then, a neuralized value iteration \cite{tamar2016value} is applied in the discretized latent space for informed sample generation. Transformer Enhanced Motion Planner (TEMP) \cite{zhuang2024transformer} leverages the transformer architecture's ability to encode long-horizon dependencies and inter-relationships for generating next state configuration. This framework gets workspace embeddings, planning history, and planning objectives, and outputs informed samples to expand the constructed tree towards the planning goal. This approach has enhanced both planning time and planning efficiency compared to classical planners.

\subsubsection{Steering Primitive} \label{subsubsec: DL-sampling-based-steering}
Neural network architectures can also be utilized for steering in sampling-based planning algorithms. Typically, classical sampling-based planners steer towards the sampled configuration along a straight line, which requires fine-grained collision queries. Learning-based customized steering function can be applied to minimize the early termination of the expansion and improve the efficiency of the planner. Table \ref{tab: steeringref} provides an overview of the state-of-the-art of utilizing deep learning for steering in sampling-based planning algorithms.

\begin{table*}[htbp]
\centering
\caption{Overview of the state-of-the-art of deep learning for improving the steering primitive, including the planning metrics reported relative to evaluated benchmark methods, and each approach's primary contribution to robotic manipulator motion planning.}
\label{tab: steeringref}
\resizebox{\textwidth}{!}{%
\begin{tabular}{@{}P{0.1\textwidth}C{0.15\textwidth}C{0.15\textwidth}C{0.15\textwidth}P{0.45\textwidth}}
\arrayrulecolor{gray!80}
\hline
\rowcolor{gray!20}
\textbf{Paper}& \multicolumn{3}{c}{\textbf{Benchmarks \& Metrics}}& {\textbf{Contributions}}\\
\hline
\multicolumn{1}{c}{\phantom{Var.}}&\multicolumn{1}{c}{\cellcolor{gray!20}{T $[s]\downarrow$}}&\multicolumn{1}{c}{{\cellcolor{gray!20}C $\downarrow$}}&\multicolumn{1}{c}{\cellcolor{gray!20}S $[\%]\uparrow$}&\multicolumn{1}{c}{\phantom{Var.}} \\
\hline
\multirow{2}{*}{CBF-INC \cite{yu2024efficient}}&$345.8$&\#V : $\mathbf{\underline{162.5}}$&$\mathbf{\underline{76.1}}$&\multirow{2}{*}{\begin{tabular}{l}
    $\bullet$ Utilizes a neural control barrier function for effective steering.
\end{tabular}} \\
\cmidrule{2-4}
&$\mathbf{\underline{287.6}}$ (RRT)&\#V: $252.5$ (RRT)&$62.5$ (RRT)& \\
\hline
\multirow{2}{*}{\begin{tabular}{l}
    Chiang \textit{et al.} \\
     \cite{chiang2021fast}
\end{tabular}}&-&PL - L : $\mathbf{\underline{90}}$&$\mathbf{\underline{80}}$&\multirow{2}{*}{\begin{tabular}{l}
    $\bullet$ Deep neural network for implicit swept volume estimation \\
    for efficient steering. 
\end{tabular}} \\
\cmidrule{2-4}
&-&PL - L: $100$ (PRM)&$40$ (PRM)& \\
\bottomrule
\end{tabular}}
\begin{minipage}{\textwidth}
\smallskip
\footnotesize
\textbf{Note:} ``$T$'' denotes \textit{planning time}, ``$C$'' denotes \textit{planning cost}, and ``$S$'' refers to \textit{success rate} (Section \ref{subsec:planning_def-metrics}). ``$\downarrow$'' indicates lower is better, and ``$\uparrow$'' indicates higher is better. ``PL-L'' refers to the swept volume of the robotic manipulator in liters, and ``\#V'' indicates the number of vertices the planner explores to plan. Please note that the main difference between the listed methods and respective benchmarks is the steering function within the sampling-based planning algorithm.
\end{minipage}
\end{table*}

\vspace{0.2cm}
\noindent
\textbf{Multi-Layer Perceptrons:} MLPs are utilized to learn the steering primitive in sampling-based planning algorithms. One approach to implementing customized steering functions involves using control barrier function (CBF)-based \cite{ames2016control} steering controllers. These controllers steer the robot towards the sampled configuration while avoiding collision with the obstacles. However, hand-crafted CBF-based controllers often struggle to generalize to unseen configuration spaces, and different robots, particularly those with higher DOF. Control Barrier Function-Induced Neural Controller (CBF-INC) \cite{yu2024efficient} addresses this limitation by designing a neural control barrier function for collision avoidance and informed steering within sampling-based motion planners. By reducing the number of collision checks and enhancing the steering function, CBF-INC increases the success rate and efficiency of the sampling-based planning algorithms.

Chiang \textit{et al.} \cite{chiang2020fast, chiang2021fast} utilize deep neural networks to estimate the manipulator's swept volume between two configurations to use it as the steering primitive to improve sampling-based planning algorithms. This process leverages explicit neural representations to enhance the process. This approach benefits from the local steering module having prior knowledge and proper representation of obstacle distributions within the configuration space. However, classic measures like configuration space Euclidean distance do not account for obstacle distributions, and weighted Euclidean metrics are hard to tune. To address this, Their framework utilizes an MLP-based network to approximate swept volume, though steering with the trained Deep Neural Network (DNN) remains complex due to the frequent need for inference throughout the planning process. To mitigate this, the authors suggest using a weighted Euclidean estimator (a single-layer neural network) for pre-filtering configurations before applying the trained MLP-based swept estimator model for steering, optimizing the steering in sampling-based motion planning algorithms. However, the offline data collection required for training these deep neural networks is time-consuming, as are the pre-processing steps needed for distance calculations in generating ground truth labels. To mitigate these challenges, Sugaya \textit{et al.} \cite{sugaya2021multitask} apply transfer learning \cite{weiss2016survey} to the learning of the swept volume function, aiming to capture geometrical similarities among the same type of robotic manipulators.
\subsection{Constrained Sampling-based Planning} \label{subsec:DL-constrainedsampling}
In real-world manipulation settings, the output of the motion planning algorithms must satisfy certain task-specific and manipulator-specific constraints (Section \ref{sec:planning_algortihms}). Deep learning methods have been effectively utilized to encode constraint representations in constrained motion planning. Table \ref{tab: constrainedsamplingref} provides an overview of the state-of-the-art of utilizing deep learning for constrained sampling-based planning.

\vspace{0.2cm}
\noindent
\textbf{Multi-Layer Perceptrons}: MLPs are utilized to facilitate constraint-aware sampling and projecting informed samples on constraint manifolds. Constrained Motion Planning networks (CoMPNet) \cite{qureshi2020neural} extends the MPNet framework \cite{qureshi2020motion} to handle constrained motion planning. Using a projection operator, CoMPNet uses MPNet's neural stochastic sampler to generate informed samples, which are then projected onto the constraint manifold. Constraint manifolds are defined through Task Space Regions (TSRs) \cite{berenson2011task}, and task constraints are implicitly embedded within task descriptions. CoMPNetX \cite{qureshi2021constrained} extends CoMPNet by introducing a discriminator neural network that estimates the distance between generated samples and the constraint manifolds. The discriminator network gradient is leveraged to project the configuration onto the constraint manifold if a generated configuration lies outside a defined threshold.

Also, deep learning frameworks have been utilized to learn constraint manifolds and directly sample on them during planning. Equality Constraint Manifold Neural Network (ECoMaNN) \cite{sutanto2021learning} leverages a deep learning framework to learn equality constraint manifolds from demonstrations, which can then be integrated into constrained sampling-based planners. This trained framework evaluates whether a robot configuration satisfies a given constraint and, if not, determines the distance from the constraint manifold.

\begin{table*}[htbp]
\centering
\caption{Overview of the state-of-the-art of deep learning for improving constrained sampling-based planning, including the planning metrics reported relative to evaluated benchmark methods, and each approach's primary contribution to robotic manipulator motion planning.}
\label{tab: constrainedsamplingref}
\resizebox{\textwidth}{!}{%
\begin{tabular}{@{}P{0.1\textwidth}C{0.15\textwidth}C{0.15\textwidth}C{0.15\textwidth}P{0.45\textwidth}}
\arrayrulecolor{gray!80}
\hline
\rowcolor{gray!20}
\textbf{Paper}& \multicolumn{3}{c}{\textbf{Benchmarks \& Metrics}}& {\textbf{Contributions}}\\
\hline
\multicolumn{1}{c}{\phantom{Var.}}&\multicolumn{1}{c}{\cellcolor{gray!20}{T $[s]\downarrow$}}&\multicolumn{1}{c}{{\cellcolor{gray!20}C $\downarrow$}}&\multicolumn{1}{c}{\cellcolor{gray!20}S $[\%]\uparrow$}&\multicolumn{1}{c}{\phantom{Var.}} \\
\hline
\multirow{2}{*}{\begin{tabular}{l}
     CoMPNet \\
     \cite{qureshi2020neural}
\end{tabular}}&$\mathbf{\underline{4.92}}$&-&-&\multirow{2}{*}{\begin{tabular}{l}
    $\bullet$ Utilizes MPNet's sampler for informed sampling. \\
    $\bullet$ Projects samples on constraint manifolds via projection operator.
\end{tabular}} \\
\cmidrule{2-4}
&$54.81$ (CBiRRT \cite{berenson2009manipulation})&-&-& \\
\hline
\multirow{2}{*}{\begin{tabular}{l}
     CoMPNetX \\
     \cite{qureshi2021constrained}
\end{tabular}}&$\mathbf{\underline{15.05}}$&-&-&\multirow{2}{*}{\begin{tabular}{l}
    $\bullet$ Utilizes a deterministic neural projector to project informed \\
    samples on the constraint manifolds.
\end{tabular}} \\
\cmidrule{2-4}
&$19.77$ (CoMPNet \cite{qureshi2020neural})&-&-& \\
\hline
\multirow{2}{*}{\begin{tabular}{l}
     LAC-RRT \\
     \cite{ho2023lac}
\end{tabular}}&$\mathbf{\underline{5.37}}$&\#V $203.11$&$\mathbf{\underline{100}}$&\multirow{2}{*}{\begin{tabular}{l}
    $\bullet$ Utilizes an encoder-decoder structure for configuration encoding.\\
    $\bullet$ Equality constraints are enforced within the latent space.
\end{tabular}} \\
\cmidrule{2-4}
&$47.2$ (CBiRRT \cite{berenson2009manipulation})&\#V $\mathbf{\underline{196.6}}$ (CBiRRT \cite{berenson2009manipulation})&$\mathbf{\underline{100}}$ (CBiRRT \cite{berenson2009manipulation})& \\
\hline
\multirow{2}{*}{\begin{tabular}{l}
     Park \textit{et al.} \\
     \cite{park2024constrained}
\end{tabular}}&$\mathbf{\underline{2.19}}$&-&$\mathbf{\underline{100}}$&\multirow{2}{*}{\begin{tabular}{l}
    $\bullet$ Implements constraint-aware planning in VAE's latent space.\\
    $\bullet$ Implements latent jump to address manifold discontinuities \\
    within the latent space.
\end{tabular}} \\
\cmidrule{2-4}
&$2.42$ (CoMPNetX \cite{qureshi2021constrained})&-&$\mathbf{\underline{100}}$ (CoMPNetX \cite{qureshi2021constrained})& \\
\hline
\multirow{2}{*}{\begin{tabular}{l}
     Lembono \\
     \textit{et al.} \cite{lembono2021learning}
\end{tabular}}&$\mathbf{\underline{0.74}}$&\#V: $\mathbf{\underline{59.7}}$&$99$&\multirow{2}{*}{\begin{tabular}{l}
    $\bullet$ Utilizes GANs for constraint-aware informed sampling.\\
    $\bullet$ Implements an ensemble of networks for the GANs generator \\
    to encode planning multi-modality.
\end{tabular}} \\
\cmidrule{2-4}
&$1.44$ (CBiRRT2 \cite{berenson2011task})&\#V: $116.5$ (CBiRRT2 \cite{berenson2011task})&$\mathbf{\underline{100}}$ (CBiRRT2 \cite{berenson2011task})& \\
\hline
\multirow{2}{*}{\begin{tabular}{l}
     Acar \textit{et al.} \\
     \cite{acar2021approximating}
\end{tabular}}&$\mathbf{\underline{0.116}}$&-&$\mathbf{\underline{100}}$&\multirow{2}{*}{\begin{tabular}{l}
    $\bullet$ Utilizes GANs for constraint-aware informed sampling.\\
    $\bullet$ Utilizes VAEs for constraint-aware informed sampling.
\end{tabular}} \\
\cmidrule{2-4}
&$0.57$ (CBiRRT \cite{berenson2009manipulation})&-&$\mathbf{\underline{100}}$ (CBiRRT \cite{berenson2009manipulation})& \\
\bottomrule
\end{tabular}}
\begin{minipage}{\textwidth}
\smallskip
\footnotesize
\textbf{Note:} ``$T$'' denotes \textit{planning time}, ``$C$'' denotes \textit{planning cost}, and ``$S$'' refers to \textit{success rate} (Section \ref{subsec:planning_def-metrics}). ``$\downarrow$'' indicates lower is better, and ``$\uparrow$'' indicates higher is better. ``\#V'' indicates the number of steps the planner explores on constraint manifolds to plan.
\end{minipage}
\end{table*}

\vspace{0.2cm}
\noindent
\textbf{Deep Generative Networks (DGNs):} DGNs' ability to learn the underlying distribution of datasets has been leveraged to directly learn constrained manifolds for constraint-aware sampling. DGN models such as auto-encoders (AEs), and generative adversarial networks (GANs) are commonly used for this purpose.

\vspace{0.2cm}
\noindent
\textbf{DGNs - Auto-Encoders (AEs):} AEs have been used for learning equality constraint manifolds in constrained motion planning. Learning-Assisted Constrained RRT (LAC-RRT) \cite{ho2023lac} utilizes Configuration Transfer Model (CTM) - a self-supervised, encoder-decoder architecture - to map configurations from configuration space to a feature space for constraint-aware sampling. In this feature space, a custom feature composer is utilized to impose planning equality constraints on the encoder's output. This is advantageous as handling equality constraints in the feature space becomes more straightforward. Additionally, Park \textit{et al.} \cite{park2024constrained} use a conditional variational autoencoder (CVAE) to learn constraint manifolds. Their framework maps the planning problem into a latent space conditioned on planning constraints and performs planning within this space. A validity network is employed to verify constraint satisfaction in the latent space. After planning, the path is mapped back to the configuration space, where a numerical projector projects the generated waypoints onto the constraint manifolds.

\vspace{0.2cm}
\noindent
\textbf{DGNs - Generative Adversarial networks (GANs):} GANs also have been used to learn constraint manifolds due to their scalability to high-DOF spaces and encoding conditional distributions. Lembono \textit{et al.} \cite{lembono2021learning, lembono2020generative} utilized a GANs architecture to generate samples that are already close to the constraints manifold. Their approach involves using an ensemble of neural networks in the GAN generator to encode the multi-modality inherent in the dataset and mitigate mode collapse. This framework decreases the number of projections and reduces planning time. Acar \textit{et al.} \cite{acar2021approximating} utilize two deep generative networks (DGNs)- CVAEs, and GANs - for constraint-aware sample generation. Their method learn task-specific constraint manifolds - such as end-effector pose, closed-kinematic chain, and balance - to facilitate sampling within the framework of constrained sampling-based planning algorithms.

\subsection{Global Trajectory Optimization} \label{subsec:DL-optimization}
Deep learning methods have increasingly been leveraged to learn prior trajectory distributions to guide the trajectory optimization algorithm. These frameworks combine the expressive power of deep learning frameworks with robust trajectory optimization algorithms for efficient trajectory planning. Table \ref{tab: trajectoryoptimizationref} provides an overview of the state-of-the-art of utilizing deep learning for warm-starting trajectory optimization algorithms.

\vspace{0.2cm}
\noindent
\textbf{Multi-Layer Perceptrons (MLPs):} MLPs are used to learn the initial trajectories from planning data due to their universal approximation capability. Deep-learning Jerk-limited Grasp Optimized Motion Planner (DJ-GOMP) \cite{ichnowski2020deep} incorporates an MLP-based deep-learning module to provide a robust initial approximation for the associated optimization-based motion planning algorithm. The proposed network has multiple output heads to generate initial paths with different horizon lengths. Then, a classification network predicts the optimal initial trajectory from the set of initial trajectories with various horizon lengths. Similarly, Banerjee \textit{et al.} \cite{banerjee2020learning} employed an MLP framework to warm-start their underlying trajectory optimization problem. The proposed MLP-based framework takes the start and goal states in the workspace and generates the coefficients for a polynomial parameterization of the initial trajectory.

Constrained Neural Motion Planning with B-splines (CNP-B) \cite{kicki2023fast} employs a deep neural network for constrained kinodynamic motion planning. The initial trajectory is time-parameterized using B-splines, and the network predicts the control points of the spline. The training loss is defined based on the constraint manifold loss, allowing the model to be trained with supervision using only motion planning problem instances.

\vspace{0.2cm}
\noindent
\textbf{Deep Generative Networks (DGNs):} DGNs are increasingly used in motion planning to learn distributions of successful paths from datasets, which can be used as priors in trajectory optimization problems. These learned priors can be used to generate the initial trajectories for gradient-based trajectory optimization problems, or within a maximum-a-posterior (MAP) formulation of trajectory optimization problems, typically conditioned on motion planning constraints and goals, for end-to-end trajectory optimization. 

\vspace{0.2cm}
\noindent
\textbf{DGNs - Auto-Encoders (AEs):} AEs' ability to encode the underlying distribution within datasets is utilized to generate the initial trajectory for the trajectory optimization algorithm. Motion Planning by Learning the Solution Manifold (MPSM) \cite{osa2022motion} learns the solution manifold of trajectories using a VAE architecture with a customized loss function. Trained on non-optimal trajectories sampled from a baseline proposal distribution, the framework generates trajectories that are further finetuned using the CHOMP \cite{ratliff2009chomp} algorithm.

\begin{table*}[htbp]
\centering
\caption{Overview of the state-of-the-art of deep learning for improving trajectory optimization, including the planning metrics reported relative to evaluated benchmark methods, and each approach's primary contribution to robotic manipulator motion planning.}
\label{tab: trajectoryoptimizationref}
\resizebox{\textwidth}{!}{%
\begin{tabular}{@{}P{0.1\textwidth}C{0.15\textwidth}C{0.15\textwidth}C{0.15\textwidth}P{0.45\textwidth}}
\arrayrulecolor{gray!80}
\hline
\rowcolor{gray!20}
\textbf{Paper}& \multicolumn{3}{c}{\textbf{Benchmarks \& Metrics}}& {\textbf{Contributions}}\\
\hline
\multicolumn{1}{c}{\phantom{Var.}}&\multicolumn{1}{c}{\cellcolor{gray!20}{T $[s]\downarrow$}}&\multicolumn{1}{c}{{\cellcolor{gray!20}C $\downarrow$}}&\multicolumn{1}{c}{\cellcolor{gray!20}S $[\%]\uparrow$}&\multicolumn{1}{c}{\phantom{Var.}} \\
\hline
\multirow{2}{*}{\begin{tabular}{l}
     DJ-GOMP \\
     \cite{ichnowski2020deep}
\end{tabular}}&$\mathbf{\underline{0.75}}$&-&-&\multirow{2}{*}{\begin{tabular}{l}
    $\bullet$ Utilizes an MLP framework to estimate the initial trajectory \\
    for warm-starting trajectory optimization.
\end{tabular}} \\
\cmidrule{2-4}
&$1.25$ (TrajOpt \cite{schulman2014motion})&-&-& \\
\hline
\multirow{2}{*}{\begin{tabular}{l}
     CNP-B \cite{kicki2023fast}
\end{tabular}}&$\mathbf{\underline{8}}$ ms&-&$\mathbf{\underline{100}}$&\multirow{2}{*}{\begin{tabular}{l}
    $\bullet$ Parametrizes the initial trajectory via B-spline.\\
    $\bullet$ utilizes an MLP to predict the control points of the B-spline.
\end{tabular}} \\
\cmidrule{2-4}
&$60$ ms (BIT*)&-&$\mathbf{\underline{100}}$ (BIT*)& \\
\hline
\multirow{2}{*}{\begin{tabular}{l}
     MPSM \cite{osa2022motion}
\end{tabular}}&$\mathbf{\underline{0.21}}$&-&-&\multirow{2}{*}{\begin{tabular}{l}
    $\bullet$ Encodes solution manifolds with a customized VAE structure.\\
    $\bullet$ Utilizes CHOMP \cite{ratliff2009chomp} for fine-tuning the generated trajectories.
\end{tabular}} \\
\cmidrule{2-4}
&$1.8$ (CHOMP \cite{ratliff2009chomp})&-&-& \\
\hline
\multirow{2}{*}{\begin{tabular}{l}
     Ando \textit{et al.} \\
     \cite{ando2023learning}
\end{tabular}}&$\mathbf{\underline{5.94}}$ ms&PL - W: $\mathbf{\underline{3}}$&$81.8$&\multirow{2}{*}{\begin{tabular}{l}
    $\bullet$ Encodes solution manifolds with a GAN architecture.\\
    $\bullet$ Trajectory optimization is done in the obstacle-free latent space.
\end{tabular}} \\
\cmidrule{2-4}
&$125$ ms (BiRRT)&PL - W: $3.14$ (BiRRT)&$\mathbf{\underline{100}}$ (BiRRT)& \\
\hline
\multirow{2}{*}{\begin{tabular}{l}
     GFCOP \\
     \cite{zhi2023global}
\end{tabular}}&$\mathbf{\underline{5.2}}$&PL - W: $\mathbf{\underline{1.32}}$&$98$&\multirow{2}{*}{\begin{tabular}{l}
    $\bullet$ GFs \cite{van2022geometric} for local and CMA-ES \cite{hansen2016cma} for global optimality.\\
    $\bullet$ An EBM for warm-starting CMA-ES optimization.
\end{tabular}} \\
\cmidrule{2-4}
&$6$ (BIT*)&PL - W: $3.71$ (BIT*)&$\mathbf{\underline{100}}$ (BIT*)& \\
\hline
\multirow{2}{*}{\begin{tabular}{l}
     MPD \cite{carvalho2023motion}
\end{tabular}}&$\mathbf{\underline{1.1}}$&PL - C: $9.9$&$\mathbf{\underline{100}}$&\multirow{2}{*}{\begin{tabular}{l}
    $\bullet$ Formulates motion planning as inference with a diffusion model.\\
    $\bullet$ Utilizes temporal U-net \cite{janner2022planning} for the denoising process.
\end{tabular}} \\
\cmidrule{2-4}
&$194.4$ (GPMP \cite{mukadam2016gaussian})&PL - C: $\mathbf{\underline{5.1}}$ (GPMP \cite{mukadam2016gaussian})&$42$ (GPMP \cite{mukadam2016gaussian})& \\
\hline
\multirow{2}{*}{\begin{tabular}{l}
     M$^{2}$Diffuser \\
     \cite{yan2024m2diffuser}
\end{tabular}}&$3.89$&-&$\mathbf{\underline{30.49}}$&\multirow{2}{*}{\begin{tabular}{l}
    $\bullet$ Diffusion process for trajectory distribution learning.\\
    $\bullet$ Task goals and planning costs guide the generative algorithm.
\end{tabular}} \\
\cmidrule{2-4}
&$\mathbf{\underline{0.46}}$ (M$\pi$Nets \cite{fishman2023motion})&-&$3.24$ (M$\pi$Nets \cite{fishman2023motion})& \\
\hline
\multirow{2}{*}{\begin{tabular}{l}
     EDMP \cite{saha2023edmp}
\end{tabular}}&-&-&$\mathbf{\underline{85}}$&\multirow{2}{*}{\begin{tabular}{l}
    $\bullet$ Diffusion-based trajectory generation with collision guidance.\\
    $\bullet$ Ensemble-of-collision-costs improves generalizability.
\end{tabular}} \\
\cmidrule{2-4}
&-&-&$32$ (CHOMP \cite{ratliff2009chomp})& \\
\hline
\multirow{2}{*}{\begin{tabular}{l}
     Sharma \textit{et al.} \\
     \cite{sharma2025cascaded}
\end{tabular}}&$2.74$&-&$\mathbf{\underline{85.13}}$&\multirow{2}{*}{\begin{tabular}{l}
    $\bullet$ utilizes a hierarchy of cascaded diffusion models to encode \\
    global and local information of the planning problem.
\end{tabular}} \\
\cmidrule{2-4}
&$\mathbf{\underline{1.94}}$ (EDMP \cite{saha2023edmp})&-&$80.3$ (EDMP \cite{saha2023edmp})& \\
\hline
\multirow{2}{*}{\begin{tabular}{l}
     DiffSeeder \\
     \cite{huang2024diffusionseeder}
\end{tabular}}&$42.3$&-&$\mathbf{\underline{86}}$&\multirow{2}{*}{\begin{tabular}{l}
    $\bullet$ Employs a conditional diffusion model for warm-starting \\
    the cuRobo \cite{sundaralingam2023curobo} trajectory optimization algorithm.
\end{tabular}} \\
\cmidrule{2-4}
&$\mathbf{\underline{38.6}}$ (cuRobo \cite{sundaralingam2023curobo})&-&$57$ (cuRobo \cite{sundaralingam2023curobo})& \\
\hline
\multirow{2}{*}{\begin{tabular}{l}
     Luo \textit{et al.} \\
     \cite{luo2024potential}
\end{tabular}}&$\mathbf{\underline{0.13}}$&-&$\mathbf{\underline{99.7}}$&\multirow{2}{*}{\begin{tabular}{l}
    $\bullet$ Trains EBMs of the planning space. \\
    $\bullet$ Leverages EMBs' gradient for the denoising process.
\end{tabular}} \\
\cmidrule{2-4}
&$0.22$ (M$\pi$Nets \cite{fishman2023motion})&-&$88$ (M$\pi$Nets \cite{fishman2023motion})& \\
\hline
\multirow{2}{*}{\begin{tabular}{l}
     FlowMP \\
     \cite{nguyen2025flowmp}
\end{tabular}}&$\mathbf{\underline{0.13}}$&-&-&\multirow{2}{*}{\begin{tabular}{l}
    $\bullet$ Incorporates acceleration and jerk fields for smooth \\
    and dynamically aware trajectory generation.
\end{tabular}} \\
\cmidrule{2-4}
&$5.29$ (MPD \cite{carvalho2023motion})&-&-& \\
\bottomrule
\end{tabular}}
\begin{minipage}{\textwidth}
\smallskip
\footnotesize
\textbf{Note:} ``$T$'' denotes \textit{planning time}, ``$C$'' denotes \textit{planning cost}, and ``$S$'' refers to \textit{success rate} (Section \ref{subsec:planning_def-metrics}). ``$\downarrow$'' indicates lower is better, and ``$\uparrow$'' indicates higher is better. ``PL-W'' refers to path length measured in the workspace, and ``PL-C'' denotes path length in the configuration space.
\end{minipage}
\end{table*}

\vspace{0.2cm}
\noindent
\textbf{DGNs - Generative Adversarial Networks (GANS):} GANs have been utilized to encode and learn the solution manifolds for trajectory planning. Ando \textit{et al.} \cite{ando2023learning} implemented a conditional GAN framework conditioned on obstacle information and using an RGG (comprising both in-collision and collision-free configurations). The framework encodes the free space of the configuration space into a latent space, which is entirely collision-free. This allows planning to be performed directly in the latent space, where additional constraints can also be applied. The planned path is then mapped back to the configuration space using the trained GAN.

\vspace{0.2cm}
\noindent
\textbf{DGNs - Energy-based Models (EBMs):} EBMs' ability to implicitly represent un-normalized distributions makes them capable of encoding datasets' multimodality. Urain \textit{et al.} \cite{urain2022learning} utilized EBMs , trained on demonstration trajectories, to generate prior distribution in a MAP inference framework to solve the planning problem. Instead of relying on a single monolithic prior, this framework employs a factored distribution to exploit composability to learn various aspects of the planning problem. Geometric Fabrics Command Optimization Problem (GFCOP) \cite{zhi2023global} uses Geometric Fabrics (GFs) \cite{van2022geometric} to generate locally optimal motion trajectories and applies Covariance Matrix Adaptation Evolution Strategy (CMA-ES) \cite{hansen2016cma} over trajectory waypoints for reducing global planning cost. An EBM is used to warm-start the CMA-ES global optimization. The EBM is trained in a self-supervision manner, where solutions from CMA-ES are continuously added to a buffer, enabling a unified loop of learning and optimization.

\vspace{0.2cm}
\noindent
\textbf{DGNs - Diffusion Models:} Diffusion models' ability in encoding multi-modal and high-dimensional data, as well as stable training, made them a good choice for learning motion policies from an expert dataset. One application of these models in trajectory optimization is generating the initial seed for the downstream optimization problem. DiffusionSeeder (DiffSeeder) \cite{huang2024diffusionseeder} also utilizes a conditional diffusion model, conditioned on environment embedding, for seeding cuRobo \cite{sundaralingam2023curobo} motion optimization algorithm.

Motion Planning Diffusion (MPD) \cite{carvalho2023motion, carvalho2024motion} formulates motion planning as an inference problem using a diffusion model. In this framework, a temporal U-net \cite{janner2022planning} is trained as a prior on expert demonstrations. Then, a task-conditioned posterior is defined based on task-specific costs such as collision, self-collision, joint limits violations, enabling the sampling of optimal trajectories. Language-guided Object Centric 3D Diffusion Policy (Lan-o3dp) \cite{li2024language} incorporates collision avoidance as a constraint function applied to the end-effector of the robotic manipulator. This constraint is used to guide the generation of waypoints for the end-effector. Similarly, Nikken \textit{et al.} \cite{nikken2024denoising} use a conditional diffusion model to generate trajectories for the end-effector of a robotic manipulator. Instead of relying on expert demonstrations, the framework generates training data through linear interpolation between start and goal poses and conditions on obstacle information during training.

Mobile Manipulation Diffuser (M$^{2}$Diffuser) \cite{yan2024m2diffuser} learns scene-conditioned goal-directed trajectory-level distributions \cite{huang2023diffusion} using a diffusion process trained on expert demonstrations for mobile manipulators. Then, a guided trajectory optimization is performed for finding the optimal trajectory. During inference, the trajectory optimization problem is guided by task objective energy functions (e.g., grasping, placement) as well as planning constraint functions (e.g., smoothness and collision avoidance). APEX \cite{dastider2024apex} introduces an obstacle-guided diffusion-based trajectory planner for dual-arm robotic manipulators. This platform first transforms the planning problem into a latent space with a variational autoencoder, then employs an obstacle-guidance classifier to perform planning and replanning in dynamic environments.

Li \textit{et al.} \cite{li2024efficient, li2024constraint} utilize a constraint-aware diffusion model for trajectory optimization. In this framework, planning constraints are incorporated into the training loss function for accurate trajectory representation. Planning with Environment Representation, Sampling, and Trajectory Optimization (PRESTO) \cite{seo2024presto} employs a conditional diffusion process to generate initial trajectories for the downstream trajectory optimization. The framework constructs an environmental embedding by identifying key configurations, which are then input into a Diffusion Transformer (DiT) \cite{peebles2023scalable} for the denoising process. Planning constraints such as collision avoidance, and smoothness are incorporated into the training loss to enhance generalizability to new planning instances.

Ensemble-of-costs-guided Diffusion for Motion Planning (EDMP) \cite{saha2023edmp} runs multiple guided diffusion models in parallel for trajectory generation. Each diffusion process is guided by a specific cost function and corresponding guidance hyperparameter. The use of an ensemble of collision costs improves the planner's generalization across different planning environments. Sharma \textit{et al.} \citenum{sharma2025cascaded} propose a hierarchical cascaded diffusion planner for global planning in complex environments. The framework employs a high-level diffusion model to generate a coarse plan from the start to the goal configuration, and uses lower-level diffusion models to satisfy local constraints.

Power \textit{et al.} \cite{power2023sampling} propose a constraint-composable diffusion model to generate initial trajectories for warm-starting the Constrained Stein Variational Trajectory Optimization (CSVTO) \cite{power2024constrained} algorithm. The model is a classifier-free diffusion model trained on a dataset generated by CSVTO for individual constraints. During inference, planning constraints are composed to improve generalization to new planning scenarios. Luo \textit{et al.} \cite{luo2024potential} introduce a diffusion potential field for trajectory generation, where EBMs are trained on successful paths to encode promising regions of the planning space. The gradients of learned energy functions are utilized to guide the denoising process during trajectory sampling. In this framework, different EBMs are associated with different environments and can be combined to enable generalization to out-of-distribution trajectory generation scenarios.

\vspace{0.2cm}
\noindent
\textbf{DGNs - Flow Matching (FM):} Flow matching's ability to directly learn a time-dependent transport from a source distribution to a target distribution makes it well-suited for trajectory optimization through sampling. FlowMP \cite{nguyen2025flowmp} improves upon MPD \cite{carvalho2023motion} by utilizing flow matching to learn the trajectory prior. It extends the flow matching framework by incorporating acceleration and jerk fields, enabling the generation of smoother and more dynamically feasible trajectories. Safe Flow Matching (SafeFlow)\cite{dai2025safe} combines the flexibility of flow matching methods with control barrier functions (CBFs) to introduce Flow Matching Barrier Functions (FMBFs), which provide formal safety guarantees for the generated trajectories. FMBFs incorporate dynamic control inputs as a regularization term to guide the flow toward the safe manifold (i.e., collision-free planning space). Tian \textit{et al.} \cite{tian2025warm} Leverages a flow matching framework to learn the planning distribution of feasible motion plans to warm-start cuRobo \cite{sundaralingam2023curobo} for efficient motion planning.
\subsection{Collision and Proximity Querying} \label{subsec:DL-collisionchecking}
Collision checking is the main bottleneck in motion planning algorithms, accounting for up to 90\% of the computation time \cite{kleinbort2020collision}. Deep neural networks have been utilized as proxy collision checkers to mitigate this limitation. Table \ref{tab: collisioncheckingref} provides an overview of the state-of-the-art of utilizing deep learning for improving collision checking.

\begin{table*}[htbp]
\centering
\caption{Overview of the state-of-the-art of deep learning for improving collision checking, including the metrics reported relative to evaluated benchmark methods, and each approach's primary contribution to robotic manipulator motion planning.}
\label{tab: collisioncheckingref}
\resizebox{\textwidth}{!}{%
\begin{tabular}{@{}P{0.1\textwidth}C{0.15\textwidth}C{0.15\textwidth}C{0.15\textwidth}P{0.45\textwidth}}
\arrayrulecolor{gray!80}
\hline
\rowcolor{gray!20}
\textbf{Paper}& \multicolumn{3}{c}{\textbf{Benchmarks \& Metrics}}& {\textbf{Contributions}}\\
\hline
\multicolumn{1}{c}{\phantom{Var.}}&\multicolumn{1}{c}{\cellcolor{gray!20}{T $[s]\downarrow$}}&\multicolumn{1}{c}{{\cellcolor{gray!20}C $\downarrow$}}&\multicolumn{1}{c}{\cellcolor{gray!20}S $[\%]\uparrow$}&\multicolumn{1}{c}{\phantom{Var.}} \\
\hline
\multirow{2}{*}{\begin{tabular}{l}
     SLIP+$^*$ \\
     \cite{luo2023reinforcement}
\end{tabular}}&$\mathbf{\underline{0.13}}$&-&$\mathbf{\underline{96}}$&\multirow{2}{*}{\begin{tabular}{l}
    $\bullet$ Utilizes an MLP framework to determine the collision status \\
    of robotic configurations.
\end{tabular}} \\
\cmidrule{2-4}
&$1.02$ (PRM)&-&$77$ (PRM)& \\
\hline
\multirow{2}{*}{\begin{tabular}{l}
     Krawczyk \\
     \textit{et al.} \cite{krawczyk2023comparison}
\end{tabular}}&$\mathbf{\underline{0.16}}$ ms&-&-&\multirow{2}{*}{\begin{tabular}{l}
    $\bullet$ Utilizes an MLP framework to determine the collision status \\
    of robotic configurations.
\end{tabular}} \\
\cmidrule{2-4}
&$0.31$ ms (FCL \cite{pan2012fcl})&-&-& \\
\hline
\multirow{2}{*}{\begin{tabular}{l}
     CN-RRT$^*$ \\
     \cite{chase2020neural}
\end{tabular}}&$\mathbf{\underline{2.0}}$&\#V: $\mathbf{\underline{964}}$&$\mathbf{\underline{97}}$&\multirow{2}{*}{\begin{tabular}{l}
    $\bullet$ Utilizes an MLP framework to determine the separation \\
    distance between the manipulator and the surrounding environment.
\end{tabular}} \\
\cmidrule{2-4}
&$2.7$ (GJK-RRT \cite{gilbert1988fast})&\#V: $1214$ (GJK-RRT \cite{gilbert1988fast})&$95.2$ (GJK-RRT \cite{gilbert1988fast})& \\
\hline
\multirow{2}{*}{\begin{tabular}{l}
     SCN \cite{danielczuk2021object} \\
\end{tabular}}&$\mathbf{\underline{0.01}}$ ms&-&$\mathbf{\underline{93.2}}$&\multirow{2}{*}{\begin{tabular}{l}
    $\bullet$ Utilizes 3D CNN to process workspace point cloud. \\
    $\bullet$ Processes target object through PointNet++ \cite{qi2017pointnet++}.
\end{tabular}} \\
\cmidrule{2-4}
&$0.49$ ms (FCL \cite{pan2012fcl})&-&$75.4$ (FCL \cite{pan2012fcl})& \\
\hline
\multirow{2}{*}{\begin{tabular}{l}
     CabiNet \cite{murali2023cabinet} \\
\end{tabular}}&$\mathbf{\underline{6.41}}$ $\mu$s&-&$\mathbf{\underline{89}}$&\multirow{2}{*}{\begin{tabular}{l}
    $\bullet$ Improves upon SceneCollisionNet \cite{danielczuk2021object} by considering diverse \\
    workspaces.
\end{tabular}} \\
\cmidrule{2-4}
&$7.03$ $\mu$s (SCN \cite{danielczuk2021object})&-&$69.9$ (SCN \cite{danielczuk2021object})& \\
\hline
\multirow{2}{*}{\begin{tabular}{l}
     GraphDistNet \\
     \cite{kim2022graphdistnet}
\end{tabular}}&$3.2$&-&$\mathbf{\underline{70}}$&\multirow{2}{*}{\begin{tabular}{l}
    $\bullet$ utilizes two graphs to encode the geometrical relation between \\
    the manipulator and workspace obstacles.
\end{tabular}} \\
\cmidrule{2-4}
&$\mathbf{\underline{1.9}}$ (ClearanceNet \cite{chase2020neural})&-&$30$ (ClearanceNet \cite{chase2020neural})& \\
\hline
\multirow{2}{*}{\begin{tabular}{l}
     GDN-R \cite{song2023graph}
\end{tabular}}&$\textbf{\underline{0.18}}$&-&-&\multirow{2}{*}{\begin{tabular}{l}
    $\bullet$ utilizes Gumbel top-k relaxation to identify highly probable \\
    interconnections between graph geometries.
\end{tabular}} \\
\cmidrule{2-4}
&$0.38$ (GraphDistNet \cite{kim2022graphdistnet})&-&-& \\
\hline
\multirow{2}{*}{\begin{tabular}{l}
     PairwiseNet \\
     \cite{kim2023pairwisenet}
\end{tabular}}&-&-&$\mathbf{\underline{99}}$&\multirow{2}{*}{\begin{tabular}{l}
    $\bullet$ Focuses on pairwise collision distance estimation.\\
    $\bullet$ Utilizes DGCNN \cite{wang2019dynamic} for encoding the point cloud of objects.
\end{tabular}} \\
\cmidrule{2-4}
&-&-&$96$ (ClearanceNet \cite{chase2020neural})& \\
\hline
\multirow{2}{*}{\begin{tabular}{l}
     DistFormer \\
     \cite{cao2023distformer}
\end{tabular}}&$31458$ ms&-&$\mathbf{\underline{99.1}}$&\multirow{2}{*}{\begin{tabular}{l}
    $\bullet$ Utilizes bounding sequence to retain geometrical properties.\\
    $\bullet$ Leverages transformers to estimate distance to collision.
\end{tabular}} \\
\cmidrule{2-4}
&$\mathbf{\underline{14797.2}}$ ms (GraphDistNet \cite{kim2022graphdistnet})&-&$97.76$ (GraphDistNet \cite{kim2022graphdistnet})& \\
\hline
\multirow{2}{*}{\begin{tabular}{l}
     CompositeSDF$^*$ \\
     \cite{liu2023collision}
\end{tabular}}&$\mathbf{\underline{4.78}}$&-&-&\multirow{2}{*}{\begin{tabular}{l}
    $\bullet$ Utilizes MLP frameworks to learn an SDF network for \\
    each link.
\end{tabular}} \\
\cmidrule{2-4}
&$241$ (GJK-RRT \cite{gilbert1988fast})&-&-& \\
\hline
\multirow{2}{*}{\begin{tabular}{l}
     SE3NN \\
     \cite{kim2024active}
\end{tabular}}&$0.189$&-&$\mathbf{\underline{90}}$&\multirow{2}{*}{\begin{tabular}{l}
    $\bullet$ Utilizes a link SE(3) representation for planning embedding.\\
    $\bullet$ Leverages continual learning to handle dynamic environments.
\end{tabular}} \\
\cmidrule{2-4}
&$\mathbf{\underline{0.148}}$ (ClearanceNet \cite{chase2020neural})&-&$73$ (ClearanceNet \cite{chase2020neural})& \\
\bottomrule
\end{tabular}}
\begin{minipage}{\textwidth}
\smallskip
\footnotesize
\textbf{Note:} Please note that $^*$ indicates that the collision checking framework is embedded within a planning algorithm for benchmark comparison. ``$T$'' denotes \textit{planning time} for rows marked with $^*$, and \textit{average collision query time for the others}, ``$C$'' denotes \textit{planning cost}, and ``$S$'' refers to \textit{success rate} (Section \ref{subsec:planning_def-metrics}). ``$\downarrow$'' indicates lower is better, and ``$\uparrow$'' indicates higher is better. ``\#V'' indicates the number of steps the planner explores on constraint manifolds to plan.
\end{minipage}
\end{table*}

\vspace{0.2cm}
\noindent
\textbf{Multi-Layer Perceptrons (MLPs):} One line of research has used MLP frameworks for binary collision checking. Ichter \textit{et al.} \cite{ichter2019robot} employ an MLP-based binary collision checker within a sampling-based planning framework. Trained on data generated by a classical collision checker, the model acts as a binary classifier that takes consecutive states and a workspace embedding as input to determine whether the path between the two states is in collision.

Liu \textit{et al.} \cite{liu2025reliable} propose an MLP-based framework for binary self-collision detection in redundant manipulators. The network is trained on randomly sampled configurations labeled with their collision status and, during inference, predicts the probability of a given configuration being in self-collision. Additionally, Self-imitation Learning by Planning Plus (SLIP+) \cite{luo2023reinforcement} trains a deep neural network to assess the collision probability of a given robot configuration. This network takes the robot's configuration and the workspace embedding as input and outputs the collision probability of the state. Also, Krawczyk \textit{et al.} \cite{krawczyk2023comparison} utilize an MLP-based framework to predict the self-collision status of a mobile manipulator. Trained on data generated by the Flexible Collision Library (FCL) \cite{pan2012fcl}, the model takes the mobile manipulator configuration as input and outputs its collision status. 

Tran \textit{et al.} \cite{tran2020predicting} integrated a contractive auto-encoder (CAE) with an MLP to assess the collision status (i.e., collision-free or in-collision) of the sampled configurations in sampling-based motion planning frameworks. The CAE encodes the robot task space, and the output, along with a robot configuration, is fed into the MLP to determine the collision status of the configuration. DeepCollide \cite{guo2023deepcollide} employs an MLP-based structure combined with a forward kinematics kernel to determine the collision status of a robotic manipulator. This framework takes joint angles as input and outputs a collision score, enabling efficient evaluation of potential collisions. 

MLPs also have been utilized to determine the collision distance for collision and self-collision queries \cite{rakita2018relaxedik}. ClearanceNet (CN) \cite{chase2020neural}  employs an MLP framework to accurately predict the minimum separation distance between a manipulator and surrounding obstacles. The model takes as input an environment embedding and the manipulator pose, and is trained on data generated using the GJK algorithm \cite{gilbert1988fast}. By leveraging the fast inference capabilities of MLPs, ClearanceNet facilitates batch collision checking, which significantly improves planning speed by reducing the computational complexity traditionally associated with geometrical collision checking. However, since it uses point cloud representations of obstacles, ClearanceNet faces challenges in accurately estimating distances for non-convex geometries and exhibits limited generalization to unseen environments. 

Liu \textit{et al.} \cite{liu2023whole} introduced an MLP-based hierarchical self-collision detection method consisting of a classifier and a regressor designed for binary collision detection and estimating the distance-to-collision for collision-free configurations. To ensure the accuracy of the predictions, a robust geometric collision detector is used to double-check collision-free states when the distance to collision falls below a pre-defined threshold.

\vspace{0.2cm}
\noindent
\textbf{Convolutional Neural Networks (CNNs)}: CNNs are well-suited for collision checking over point clouds due to their ability to encode the local and global structures of 3D scenes. SceneCollisionNet (SCN) \cite{danielczuk2021object} is a framework designed for real-time collision checking between two point clouds - the manipulator and workspace within a motion planning context - even under partial observability. The workspace is voxelized, and a shared MLP encodes the latent features within each voxel. These voxel-wise representations are then processed through 3D convolutional layers to learn an implicit 3D embedding of the workspace. Additionally, the target object is encoded using the set abstraction layer of PointNet++ \cite{qi2017pointnet++}, enabling the downstream MLP-based binary classifier to perform collision checking between the workspace embedding and the target object. CabiNet \cite{murali2023cabinet} improves upon SceneCollisionNet capabilities to scale and generalize to various clutter environments. This is achieved by augmenting the workplaces with various common objects, such as shelves and cabinets.

\vspace{0.2cm}
\noindent
\textbf{Graph Neural Networks (GNNs):} GNNs' scalability and generalizability are utilized for collision querying in planning problems. GraphDistNet \cite{kim2022graphdistnet} aims to encode geometrical and topological relationships between a manipulator and workspace obstacles to estimate both collision distance and collision gradients. The framework constructs two graphs to encode the manipulator and the workspace and employs Graph Attention Networks \cite{velivckovic2017graph} to encode interactions between them. Trained on data generated with FCL \cite{pan2012fcl}, GraphDistNet enables accurate distance and gradient estimation and generalizes to unseen workspaces without requiring retraining. However, the cost of message passing in GraphDistNet is proportional to the graph size, which can impede its application to 3D complex geometries. To address this, GDN-R \cite{song2023graph} with layer-wise probabilistic graph rewiring is proposed for distance-to-collision estimation, utilizing Gumbel top-k relaxation \cite{kool2019stochastic} to identify high probable interconnections between graph geometries, thereby, increasing connectivity between graphed objects. The input to GDN-R consists of a graphed representation of workspace geometries, which undergo iterative message passing to produce updated embeddings. 

PairwiseNet \cite{kim2023pairwisenet} focuses on estimating pairwise collision distance, instead of focusing on global collision distance. This framework employs EdgeConv layers from Dynamic Graph Convolutional Neural Networks (DGCNN) \cite{wang2019dynamic} to encode the point-cloud of workspace shape geometries into shape feature vectors. Subsequently, a fully connected neural network processes these vectors of paired geometries to determine the minimum distance between them. Then the minimum of these pairwise distances will be treated as the global collision distance. Thanks to the pairwise collision-distance estimation, this framework easily generalizes to a workspace with new but similarly shaped geometries.

\vspace{0.2cm}
\noindent
\textbf{Transformers:} Transformers, known for their ability to learn long-horizon dependencies, have also been utilized for collision querying in mnipulator motion planning. DistFormer \cite{cao2023distformer} is a distance-to-collision estimator that leverages the attention mechanism. Trained on a dataset generated by FCL \cite{pan2012fcl}, this framework utilizes bounding sequences to retain the shape of the manipulator and obstacles, employing a self-attention mechanism to transform these bounding sequences into feature sequences. A cross-attention module then fuses the feature sequences of the manipulator and obstacles, ensuring that the manipulator is implicitly aware of the obstacle's locations. The augmented feature sequence of the manipulator is ultimately used to estimate the collision distance in this framework.

\vspace{0.2cm}
\noindent
\textbf{Large Language Models (LLMs)}: LLMs can be used to encode the workspace of a robotic manipulator by generating a voxel value map, which can serve as input for downstream collision checking algorithms. VoxPoser \cite{huang2023voxposer} utilizes LLMs to generate a 3D voxel value map based on task descriptions. It then applies a greedy search over a collision avoidance map to identify collision-free end-effector positions.

\begin{table*}[htbp]
\centering
\caption{A compact and concise overview of state-of-the-art literature that leverages various deep learning frameworks to improve various components of classical planning algorithms for robotic manipulator motion planning. \colorbox{color1}{\textit{less than 5 papers}}, \colorbox{color2}{\textit{$5\sim 10$ papers}}, \colorbox{color3}{\textit{more than 10 papers}}.}
\label{tab: guide-future}
\resizebox{\textwidth}{!}{%
\begin{tabular}{@{}P{0.05\textwidth}C{0.05\textwidth}C{0.15\textwidth}C{0.15\textwidth}C{0.15\textwidth}C{0.15\textwidth}C{0.15\textwidth}C{0.15\textwidth}}
\arrayrulecolor{gray!80}
\hline
\rowcolor{gray!20}
\multicolumn{2}{c}{\phantom{Var.}} &{\textbf{E2E Planning}} & \multicolumn{2}{c}{\textbf{U-SBMP}}& {\textbf{C-SBMP}}&{\textbf{TO}}&{\textbf{Collision Checking}}\\
\hline
\multicolumn{3}{c}{\phantom{Var.}}&\multicolumn{1}{c}{\cellcolor{gray!20}\textbf{Sampling}}&\multicolumn{1}{c}{\cellcolor{gray!20}\textbf{Steering}}&\multicolumn{3}{c}{\phantom{Var.}} \\
\midrule
\multicolumn{2}{l}{MLPs}&{\cellcolor{color1}\cite{pandy2020learning}}&{\cellcolor{color2}\cite{qureshi2019motion, qureshi2020motion, qureshideeply, tamizi2024end, parque2021learning, lai2022parallelised, bhardwaj2021leveraging, lyu2022motion}}&{\cellcolor{color1}\cite{yu2024efficient,chiang2020fast, chiang2021fast, sugaya2021multitask}}&{\cellcolor{color1}\cite{qureshi2020neural, qureshi2021constrained,sutanto2021learning}}&{\cellcolor{color1}\cite{ichnowski2020deep, banerjee2020learning, kicki2023fast}}&{\cellcolor{color2}\cite{ichter2019robot, liu2025reliable, luo2023reinforcement, krawczyk2023comparison, tran2020predicting, guo2023deepcollide, rakita2018relaxedik, chase2020neural, liu2023whole}} \\
\hline
\multicolumn{2}{l}{CNNs}&{\cellcolor{color2}\cite{ota2021deep, ni2022ntfields, ni2023progressive, ni2024physics, ni2024physicsws, liu2024physics, niphysics}}&{\cellcolor{color2}\cite{shah2022using, patil2019prediction, abdi2023hybrid, chamzas2022learning, terasawa20203d}}&{-}&{-}&{-}&{\cellcolor{color1}\cite{danielczuk2021object, murali2023cabinet}} \\
\hline
\multicolumn{2}{l}{PC-Nets}&{\cellcolor{color2}\cite{fishman2023motion, fishmanavoid, dalal2024neural, yang2025deep, soleymanzadeh2025perfact}}&{-}&{-}&{-}&{-}&{-} \\
\hline 
\multicolumn{2}{l}{RNNs}&{\cellcolor{color1}\cite{bency2019neural}}&{\cellcolor{color1}\cite{ying2021deep, hou2023data}}&{-}&{-}&{-}&{-} \\
\hline 
\multicolumn{2}{l}{GNNs}&{-}&{\cellcolor{color2}\cite{soleymanzadeh2025simpnet, liu2024kg, liu2024integrating, yu2021reducing, huang2022hardware, zhang2022learning, zhang2023dyngmp}}&{-}&{-}&{-}&{\cellcolor{color1}\cite{kim2022graphdistnet, song2023graph, kim2023pairwisenet}} \\
\hline 
\multirow{6}{*}{DGMs}&{VAEs}&{\cellcolor{color1} \cite{hung2022reaching, yamada2023leveraging}}&{\cellcolor{color3}\cite{ichter2018learning, dastider2023damon, dastider2022sera, dastider2022reactive, dastider2024unified, ichter2019robot, kumar2019lego, jenamani2020robotic, gaebert2022learning, kobashi2023learning, lu2024neural, huang2024planning, johnson2023learning, johnson2023zero, xia2022graph}}&{-}&{\cellcolor{color1}\cite{ho2023lac, park2024constrained}}&{\cellcolor{color1}\cite{osa2022motion}}&{-} \\
\cmidrule{2-8}
    &{GANs}&{-}&{-}&{-}&{\cellcolor{color1}\cite{lembono2020generative, lembono2021learning, acar2021approximating}}&{\cellcolor{color1} \cite{ando2023learning}}&{-} \\
    \cmidrule{2-8}
    &{NFs}&{-}&{\cellcolor{color1}\cite{lai2021plannerflows}}&{-}&{-}&{-}&{-} \\
    \cmidrule{2-8}
    &{EBMs}&{-}&{-}&{-}&{-}&{\cellcolor{color1}\cite{urain2022learning, zhi2023global}}&{-} \\
    \cmidrule{2-8}
    &{DMs}&{-}&{-}&{-}&{-}&{\cellcolor{color3}\cite{huang2024diffusionseeder, carvalho2023motion, carvalho2024motion, li2024language, nikken2024denoising, yan2024m2diffuser, dastider2024apex, li2024efficient, li2024constraint, seo2024presto, power2023sampling, saha2023edmp, sharma2025cascaded, luo2024potential}}&{-} \\
    \cmidrule{2-8}
    &{FM}&{-}&{-}&{-}&{-}&{\cellcolor{color1}\cite{nguyen2025flowmp, dai2025safe, tian2025warm}}&{-} \\
\hline
\multicolumn{2}{l}{Transformers}&{-}&{\cellcolor{color1}\cite{chen2019learning, zhuang2024transformer}}&{-}&{-}&{-}&{\cellcolor{color1}\cite{cao2023distformer}} \\
\hline 
\multicolumn{2}{l}{Foundation Models}&{\cellcolor{color1}\cite{mandi2024roco, kwon2024language, bucker2022reshaping, bucker2023latte}}&{-}&{-}&{-}&{-}&{-} \\
\hline 
\multicolumn{2}{l}{Neural SDFs}&{-}&{-}&{-}&{-}&{-}&{\cellcolor{color3} \cite{koptev2022implicit, koptev2022neural, koptev2024reactive, liu2022regularized, liu2023collision, zhao2024perceptual, chenimplicit, quintero2024stochastic, li2024configuration, kim2024active, baxter2020deep, lee2022single, lee2024reliable, lee2023fast, michaux2023reachability, joho2024neural, kwon2024conformalized}} \\
\bottomrule
\end{tabular}}
\begin{minipage}{\textwidth}
\smallskip
\footnotesize
\textbf{Note:} ``E2E Planning'' denotes end-to-end planning, ``SBMP'' denotes unconstrained sampling-based motion planning algorithms, ``C-SBMP'' denotes constrained sampling-based motion planning algorithms, and ``TO'' denotes trajectory optimization algorithms.
\end{minipage}
\end{table*}  

\vspace{0.2cm}
\noindent 
\textbf{Neural SDFs}: Neural SDFs learn a continuous signed distance function using a neural network, which can be integrated into motion planning algorithms for robotic manipulators. Neural Joint Signed Distance Field (Neural JSDF) \cite{koptev2022implicit, koptev2022neural, koptev2024reactive} is an MLP-based implicit signed distance function to estimate the distance between robot links to any point in the workspace. The network is trained on a dataset containing both collision-free and in-collision samples generated randomly. The proposed network provides a smooth and differentiable distance field that can be used as a collision constraint in optimization-based trajectory planning algorithms. Neural JSDF implicitly learns the manipulator's kinematics, which leads to the accumulation of error along the forward kinematics chain. Regularized Deep Sign Distance Field (ReDSDF) \cite{liu2022regularized} introduces a neural signed distance function for articulated objects, taking robot configuration and a 3D point in the workspace as input to the signed distance, which is then used to compute the repulsive velocities in motion planning.

Composite SDF \cite{liu2023collision}  is a neural signed distance function for predicting the minimum distance between an articulated robot and any 3D point in the workspace. Instead of learning an SDF for the whole body of manipulator, this framework learns individual SDF networks for each robot link to handle the complexity of high-dimensional configuration spaces. Composite SDF utilizes the Open3D library \cite{zhou2018open3d} to sample on/near manipulator meshes for training the proposed framework. Zhao \textit{et al.} \cite{zhao2024perceptual} enhance the Composite SDF framework by incorporating the gradient of the distance function, enabling its integration into optimization-based trajectory planning algorithms. The composite neural SDF frameworks need to follow the kinematic chain step-by-step, which leads to computational complexity. To address this issue, Robot Neural Distance Function (RNDF) \cite{chenimplicit} incorporates the manipulator's forward kinematics chain implicitly within a neural framework. The model includes a regression head for each robot link, with each head conditioned on the others based on the arm's kinematic chain. This design reduces the computational complexity of explicitly computing the forward kinematics while also minimizing error accumulation associated with implicit representation.

Quintero-Pena \textit{et al.} \cite{quintero2024stochastic} extend the Neural JSDF \textit{et al.} \cite{koptev2022neural} to stochastic environments by learning a distribution over signed distance function. An MLP-based framework is utilized to learn and predict the mean and variance of the stochastic neural signed distance function, capturing uncertainty in distance estimation. Configuration Space Distance Field (CDF) \cite{li2024configuration} directly measures the distance between joint configurations and workspace obstacles in the configuration space, where planning and control are performed. This distance field preserves the Euclidean property and ensures a unit-norm gradient, eliminating the need for repeated inverse kinematics computations to map between the workspace and configuration space.

One major limitation of neural collision checkers is their poor adaptability to minor workspace changes, such as the addition of new obstacles or robots, which often necessitates complete retraining. SE3NN \cite{kim2024active} addresses this issue by utilizing an active learning method to augment the dataset with near-boundary configurations. Instead of using joint configurations as input to the neural collision checker, the framework adopts a redundant SE(3) representation of planning space for effective collision checking. Markov Chain Monte Carlo (MCMC) sampling \cite{geyer1992practical} is then employed to continuously sample boundary data for continuous training of the network.

Another approach to collision detection involves estimating the manipulator's swept volume between start and goal configurations. Deep neural networks have been widely used to approximate the geometry of this swept volume, enabling continuous collision checking. Baxter \textit{et al.} \cite{baxter2020deep} propose an MLP-based framework that takes the start and goal configurations of a manipulator as input and outputs a voxel grid representation of the swept volume. The V-REP simulation framework \cite{rohmer2013v} is used to generate training data by collecting the robot's swept volumes. This approach, however, results in low spatial resolution and inherent discretization errors. To address this issue, Lee \textit{et al} \cite{lee2022single, lee2024reliable, lee2023fast} utilize a neural network to approximate the boundary surfaces of the manipulator's swept volume. The framework employs a high fidelity neural signed SDF \cite{gropp2020implicit} for surface reconstruction to represent the manipulator's boundary surface along the trajectory, enabling accurate construction of the swept volume.

Reachability-based Signed Distance Functions (RDFs) \cite{michaux2023reachability}, an implicit neural representation, computes the distance between manipulator's parameterized swept volume and workspace objects. This framework utilizes a polynomial zonotope-based representation of the robotic arm and box obstacles to create the training dataset, and follows the network structure of \cite{gropp2020implicit} as the implicit representation. Joho \textit{et al.} \cite{joho2024neural} propose a neural continuous implicit swept volume representation that takes as input start and goal configurations along with workspace query points, and outputs the signed distance between the query points and the manipulator's swept volume. To enhance reliability, the neural collision checker is interleaved with a geometric collision checker.

Comformalized Reachable Sets for Obstacle Avoidance with Spheres (CROWS) \cite{kwon2024conformalized} uses MLP-based frameworks to model the swept volume of a robotic manipulator. One network learns the centers and radii of Spherical Forward Occupancy (FSO) \cite{michaux2024safe} of the robotic manipulator, while another provides the derivative of the SFO with respect to the sphere centers. The SFO approximates the swept volume of the robotic manipulator using a collection of spheres, and this implicit representation is used as a collision constraint in optimization-based trajectory planning.

\section{Challenges and Future Perspectives} \label{sec:perspectives}
In this section, we explore the challenges of employing deep learning frameworks for motion planning in robotic manipulators, and explore future avenues to address some of these challenges. Table \ref{tab: guide-future} provides a guide for researchers interested in using deep learning techniques for robotic manipulator motion planning. It classifies more than 100 research papers published since 2018 according to motion planning primitives and deep learning frameworks.

\subsection{Generalizability}
\subsubsection{\textbf{Challenges}}
While various deep learning frameworks incorporate different types of inductive biases to handle data not encountered during training, they often struggle to generalize to out-of-distribution settings. This is because robot skills are limited to the movement distribution learned from planning data. This challenge is particularly pronounced in motion planning for robotic manipulators, where small changes in the workspace significantly alter the planning problem \cite{farber2003topological}. The poor scalability and generalizability of neural motion planners stem from the fact that motion planning-specific datasets are scarce.

\subsubsection{\textbf{Possible Method 1: LLMs as Generalist Motion Planners}} 
\begin{figure}[!htbp]
\centering
\includegraphics[width=0.98\linewidth]{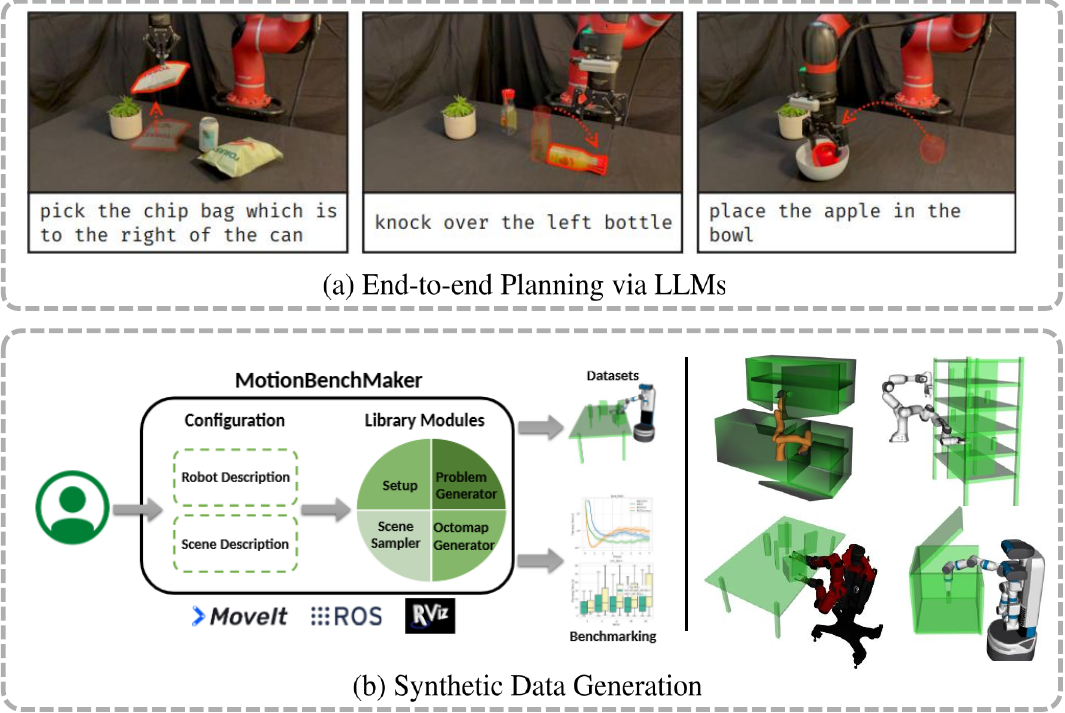}
\caption{Possible methods to improve the generalizability of neural motion planners: (a) LLMs for end-to-end planning \cite{kwon2024language}. (b) Large-scale synthetic dataset generation \cite{chamzas2021motionbenchmaker}.}
\label{fig_11- generalizability}
\end{figure}

LLMs possess a vast actionable knowledge that can be leveraged to plan motions for robotic manipulators \cite{firoozi2023foundation}. As shown in Figure \ref{fig_11- generalizability}-(a), these models can be prompted for zero-shot or few-shot motion planning for robotic manipulators. Utilizing LLMs for end-to-end motion planning is challenging because these models are not well-suited for encoding spatiotemporal dependencies within motion planning problems, as they are not trained on physical interaction datasets \cite{kwon2024language}.

A more practical application of LLMs in robotic manipulator motion planning is their potential to improve specific algorithmic primitives of classical planning algorithms. In sampling-based planning algorithms, LLMs with high temperature settings, conditioned on language instructions and visual representations of the workspace, can function as informed samplers. Moreover, these models can be instructed to learn and encode the constraint manifold, enabling constraint-aware informed sampling.

In optimization-based planning algorithms, LLMs can be used to warm-start the optimization process based on language instructions and visual representations of the workspace. For collision checking, these models can generate the workspace value map from visual and language inputs to enable real-time collision checking \cite{huang2023voxposer}.

One limitation of using LLMs within motion planning algorithms for real-time motion planning is their high inference time, as these models contain millions of parameters \cite{huber2022fast}. To mitigate this, utilizing powerful computational resources such as dedicated GPUs and TPUs can significantly reduce inference time.

\subsubsection{\textbf{Possible Method 2: Large-scale Synthetic Data Generation}} 
The generalizability of neural motion planners can be improved with large-scale planning datasets. Also, the methodology of robot foundation models \cite{brohan2023rt} can be leveraged to fine-tuning LLMs on this large-scale, planning-specific dataset.

State-of-the-art high-fidelity physics-based simulators \cite{todorov2012mujoco, makoviychuk2021isaac, coumans2016pybullet} can be leveraged to generate datasets for motion planning. However, a major limitation is the lack of variability and realism in the simulated planning workspaces. There are two possible methods to address this limitation: \textit{procedural workspace generation} and \textit{generative AI-based workspace generation}.

Procedural workspace generation methods programmatically generate cluttered workspaces for motion planning. MotionBenchMaker \cite{chamzas2021motionbenchmaker} generates diverse workspaces by procedurally generating assets with URDF sampling (Figure \ref{fig_11- generalizability}-(b)). These assets can be combined in a physics-based simulator to generate various workspaces. Classical motion planners from the Open Motion Planning Library (OMPL) \cite{sucan2012open} or advanced planners such as cuRobo \cite{sundaralingam2023curobo} can then be leveraged to generate motion planning datasets. However, MotionBenchMaker produces simple workspaces with only one major asset and large gaps between smaller ones. As a result, the generated workspaces are not realistic, and the trained neural motion planner struggle to plan in real-world settings.

Neural MP \cite{dalal2024neural} addresses MotionBenchmaker's limitations by generating more realistic and cluttered workspaces for data collection. This framework utilizes procedural assent generation and sampled everyday objects from a 3D object dataset \cite{deitke2023objaverse} to create diverse workspaces. Then, a classical planning algorithm from OMPL is utilized to collect planning data to train a generalist neural motion planner. The resulting motion planner is capable of planning within out-of-distribution scenarios, enhancing adaptability and robustness in robotic manipulator motion planning. However, randomly creating workspaces is not trivial, and the resulting workspaces may not necessarily resemble real-world scenarios.

\begin{figure*}[!htbp]
\centering
\includegraphics[width=\textwidth]{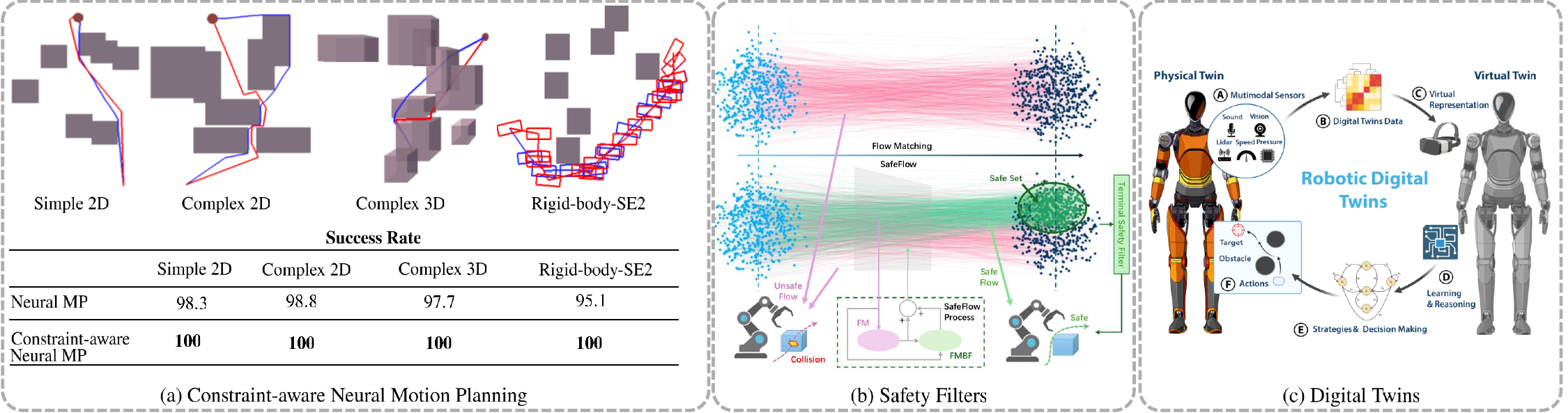}
\caption{Possible methods to enhance the safety of neural motion planners: (a) Constraint-aware neural motion planning \cite{qureshi2020motion}. (b) Safety filters \cite{dai2025safe}. (c) Digital twins \cite{li2025digital}.}
\label{fig_12- safety}
\end{figure*}

Generative AI-based workspace generation can leverage advanced AI tools to generate realistic and diverse workspaces for data collection. These methods can utilize generative methods (e.g., text-to-3D \cite{wang2023score}, 2D-to-3D \cite{chan2021pi}), or advanced NeRFs \cite{wang2024nerf} to generate articulated 3D assets from language instructions or images for workspace generation. Classical motion planners can then be utilized within these workspaces to generate a large-scale dataset. However, Current generative models struggle to generate a wide variety of articulated 3D assets with diverse configurations and articulations, which limits their effectiveness for training generalist neural motion planners. Existing 3D object datasets, such as Objaverse \cite{deitke2023objaverse} and PartNet-Mobility \cite{xiang2020sapien}, can also be used to augment generative AI workspace generation methods by directly providing articulated objects. However, the diversity of articulated objects in these datasets is limited, making them insufficient for training generalist neural motion planners.

One promising direction is to combine procedural workspace generation with generative AI methods to create realistic and diverse workspaces to collect large-scale datasets for training generalist neural motion planners.

\subsection{Safety}
\subsubsection{\textbf{Challenges}}
Neural motion planners can unlock the potential of robotic manipulators for deployment in vast real-world settings and applications. However, these planners do not provide completeness or optimality guarantees,  which may result in planning failures \cite{brunke2022safe}. On the other hand, although struggling with generalizability and scalability, classical planning algorithms do offer a certain level of completeness (e.g., probabilistic completeness of sampling-based algorithms) and theoretical optimality. Table \ref{tab: planningcomparison} provides the advantages and limitations of neural motion planners compared to classical planning algorithms.

\setlength{\tabcolsep}{3pt}
\begin{table}[htbp]
\begin{center}
\caption{A qualitative comparison between neural motion planning and classical motion planning for robotic manipulators.}
\label{tab: planningcomparison}
\begin{tabular}
{>{\raggedright\arraybackslash}p{0.08\textwidth}>{\raggedright\arraybackslash}p{0.19\textwidth}>{\raggedright\arraybackslash}p{0.18\textwidth}}
\arrayrulecolor{gray!80}
\hline
\rowcolor{gray!20} \textbf{Planners}&\textbf{Advantages}&\textbf{Limitations}\\
\hline 
\scriptsize{\textbf{SBMP}}&$\bullet$ Probabilistic completeness&$\bullet$ Non-smooth paths \\
\hline 
\scriptsize{\textbf{Global Optimization}}&$\bullet$ Smooth paths. \newline $\bullet$ Incorporates equality and non-equality constraints& $\bullet$ Susceptible to local minima \newline $\bullet$ Computationally complex \\
\hline 
\scriptsize{\textbf{Neural Planning}}&$\bullet$ Fast inference \newline $\bullet$ Encoding high-dimensional, complex distributions \newline $\bullet$ Leverages past experience for new problems& $\bullet$ Data scarcity \newline $\bullet$ Safety and reliability \\
\bottomrule
\end{tabular}
\end{center}
\begin{minipage}{0.50\textwidth}
\smallskip
\footnotesize
\textbf{Note:} ``SBMP'' denotes sampling-based motion planning, and ``Neural Planning'' refers to methods that leverage deep learning methods for planning.
\end{minipage}
\end{table}

Additionally, task-specific safety concerns require careful consideration. For example, in collaborative environments, the neural motion planner must always prioritize avoiding collisions with human operators to ensure safe operation.

Existing research on motion planning for robotic manipulators often focuses on improving efficiency and success rates using deep learning frameworks. However, many of these methods overlook critical safety aspects, even though planned trajectories must satisfy specific constraints to ensure safe operation of the robotic manipulator \cite{xu2024leto}. 

For instance, one important safety challenge arises when a robotic manipulator encounters a singularity configuration. At such configurations, the geometric Jacobian becomes ill-conditioned and loses rank, which results in the loss of one or more degrees of freedom. As a result, the end-effector movement can become unpredictable and uncontrollable \cite{spong2020robot}. However, most state-of-the-art neural motion planners overlook the singularity issue and instead focus on generalization within the workspace.

\subsubsection{\textbf{Possible Method 1: Constraint-aware Neural Motion Planning}} 
Constraint-aware motion planners combine the strengths of classical and neural motion planners for safe deployment in real-world settings \cite{qureshi2020motion}. These planners retain the completeness and optimality guarantees of classical algorithms for safe performance. At the same time, they can leverage the semantic reasoning and scene understanding capabilities of neural networks and LLMs for better generalization to domain-specific, dynamic planning scenarios (Figure \ref{fig_12- safety}-(a)).

For sampling-based planning algorithms, combining their inherent probabilistic completeness \cite{qureshi2020motion} and constraint-aware sampling \cite{qureshi2021constrained} with an implicit neural-informed sampler can enhance both the safety and adaptability of neural motion planners (Figure \ref{fig_12- safety}-(a)). Global trajectory optimization methods can integrate the safety and flexibility of optimization-based planning with the expressive power of neural networks. In these algorithms, deep generative models can capture the multi-modal distribution of the planning dataset to efficiently warm-start the optimization algorithm, while safety constraints can be incorporated as soft constraints within the optimization framework \cite{power2024constrained}. For collision checking, fast neural collision checkers can be combined with safe geometric collision checkers for safe and efficient collision checking.

Regarding singularity, one possible direction is to treat singularity as a planning cost. Prior work has mitigated the singularity through Jacobian matrix processing techniques like Damped Least Squares (DLS) \cite{mansard2009unified} and its variants \cite{guptasarma2025j}, as well as maximizing manipulability indices \cite{yoshikawa1983analysis}. Manavalan \textit{et al.} \cite{manavalan2019learning} utilize kinesthetic task demonstrations to lean the manipulator's manipulability index, given that constraints are state-dependent, and the manipulability varies during task execution. For neural motion planners, one possible approach is to identify workspace areas with low manipulability measures and guide the planner to avoid these areas. Singularity-related costs can also be incorporated into learning-based trajectory optimization methods to keep trajectories away from singular configurations. However, integrating these methods with neural motion planners for deployment in everyday unstructured environments remains challenging. Further research is necessary to combine learning-based singularity detection with neural motion planners to ensure safe operation near singular configurations.

\subsubsection{\textbf{Possible Method 2: Safety Filters}}
The generalizability and scalability of neural motion planners can be combined with model-based safety filters to avoid catastrophic failures \cite{hsu2023safety}. Safety filters monitor the operation of a robotic system and intervene when necessary \cite{hsu2021safety}.

One class of safety filters is control barrier functions (CBFs), which provide smooth intervention through real-time control optimization \cite{robey2021learning}. These methods generate a safe control signal by modifying the original task signal to satisfy the safety filter’s required decrease rate. CBFs \cite{robey2021learning} can be applied during neural path planning and/or execution to enforce safety constraints (Figure \ref{fig_12- safety}-(b)) \cite{dai2025safe}. However, designing valid CBFs for neural motion planners is challenging and requires careful consideration. Moreover, these filters mainly address geometric constraints, and additional work is needed to translate semantically defined constraints into analytical forms that can be handled by such methods \cite{brunke2025semantically}.

Another type of safety filter is trajectory optimization at runtime. For example, Model Predictive Shielding (MPS) evaluates the task control within a prediction horizon \citenum{bastani2021safe}. If the task control satisfies the safety constraints, it is executed; otherwise, a fallback policy is executed. This approach can be combined with neural motion planners to locally enforce task-specific trajectory constraints. However, MPS requires a complete dynamic model of the robotic manipulator and well-defined safety constraints, which are often difficult to obtain for real-world applications.

\subsubsection{\textbf{Possible Method 3: Digital Twins}}
High-fidelity, cost-effective simulation environments are widely used for training and optimizing embodied AI systems, including neural motion planners. However, the sim-to-real gap limits the transfer of models from simulation to the real world due to the complexity and unpredictability of real-world environments \cite{lin2025proc4gem}. Digital twins can bridge this gap by providing real-time simulations that closely mirror their physical counterparts \cite{li2025digital}.

The ability of digital twins to ensure consistency and synchronization has advanced research in waste management \cite{su2025digital} and human-robot collaboration \cite{li2024toward}. These models enable real-time monitoring, optimization, and prediction, facilitating a seamless connection between physical and virtual environments \cite{villani2024digital}. High-fidelity, task-specific digital twins have the potential to improve the safety of neural motion planners by allowing motion planning and execution to be tested in the digital twin before deployment in the real world (Figure \ref{fig_12- safety}-(c)). However, the underlying sim-to-real gap within physics engines still limits the reliability of digital twins for providing safety guarantees.

\section{Domain-specific Challenges and Potentials} \label{sec: domainspecific}
Deep learning methods hold significant promise to advance robotic manipulator planning in complex, domain-specific applications. We outline the benefits and potential risks of applying neural motion planners for robotic manipulators across various domains.

\subsection{Healthcare}
\begin{figure}[!htbp]
\centering
\includegraphics[width=0.5\textwidth]{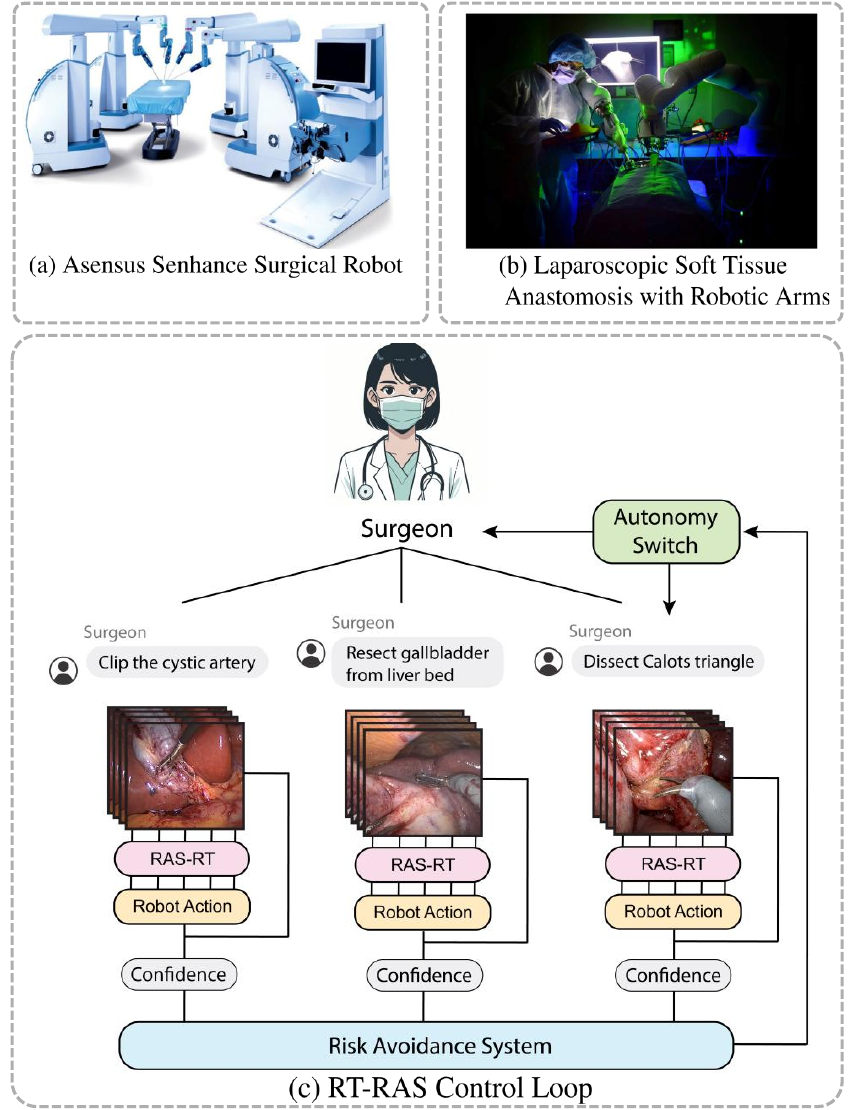}
\caption{Robot-assisted surgery. (a) Asensus Senhance surgical robot platform \cite{sarin2024upcoming}. (b) Autonomous robot for autonomous laparoscopic soft tissue anastomosis \cite{saeidi2022autonomous}. (c) Robot-assisted surgery robot transformer (RT-RAS) control loop \cite{schmidgall2024general}.}
\label{fig_13- healthcare}
\end{figure}

State-of-the-art surgical \cite{haiderbhai2024sim2real, hwang2022automating, su2021toward, su2022state, liu2024evolution} and assistive robots \cite{kidzinski2020deep, bessler2021safety, massardi2022characterization, huang2022modeling, monteiro2024human, luo2024experiment, firouzi2025biomechanical, huang2022stiffness} have achieved limited, task-specific autonomy. These systems may reduce operating time and standardize technique, potentially lowering costs and widening access, but robust generalization to complex, variable clinical conditions remains challenging. As shown in Figure \ref{fig_13- healthcare}-(a), most surgical robots still operate in master–slave teleoperation, where a surgeon commands instruments via a console or joysticks \cite{sarin2024upcoming, kantu2025portable}.

Early surgical path planning methods relied on trajectory optimizers and visual servoing under surgeon supervision, which are brittle to deformable anatomy, occlusions, and nonstationary scenes \cite{perry2013profile}. Deep learning re-frames planning as closed-loop visuo-motor proposal generation from multi-modal observations (endoscope video, robotic proprioception, and force/tactile feedback) learned via imitation, (inverse) RL, and sequence modeling \cite{schmidgall2024general, saeidi2022autonomous}. However, motion planning for surgical robots is challenging due to the complexity of surgical environments. These environments exhibit high variability across patients, continuous changes during the procedure, a limited field of view (FoV), and partial observability due to occlusions. In addition, strict safety and precision requirements further complicate the development and deployment of motion planning algorithms in surgical settings \cite{kim2025srt, su2022state}.

Recently, surgical foundation models frame planning as a closed-loop trajectory proposal conditioned on intraoperative perception. Hierarchical Surgical Transformer (SRT-H) is a hierarchical structure developed for autonomous ex vivo cholecystectomy \cite{kim2025srt}. This framework couples a language-conditioned high-level policy that issues intent instructions with a low-level controller that generates end-effector motions. The high-level policy also provides correction instructions to adjust the surgical process if needed. Experimental results show a 100$\%$ success rate across eight unseen surgical scenarios, demonstrating the potential for deploying autonomous surgical systems in clinical settings.

SRT-H robotic system \cite{kim2025srt} was evaluated in ex vivo settings, and additional considerations and risk mitigation strategies are needed for safe deployment within in vivo surgical environments. Risk avoidance methods such as conservative Q-learning \cite{chebotar2023q} and conformal methods \cite{ren2023robots} can provide the robot with confidence estimates when performing out-of-distribution tasks. Figure \ref{fig_13- healthcare}-(c) demonstrates an example of incorporating a risk avoidance system for surgical robots. Patient-specific digital twins can facilitate in vivo deployment by validating surgical trajectories on preoperative anatomy and stress-testing failure cases before real-world execution \cite{su2019mrichapter}.

Robot-assisted surgical frameworks \cite{schmidgall2024general, saeidi2022autonomous} like SRT-H can benefit from fast, collision-aware neural motion planners that propose smooth, collision-free tool paths to achieve smoother task execution and establish fully autonomous surgical robots for end-to-end procedures. Integrating safety filters into the neural motion planner \cite{dai2025safe} helps ensure safe tool movements, preventing accidental collisions with or damage to vital organs.

Surgical foundation models combined with neural motion planners necessitate large-scale surgical datasets, which are challenging to obtain due to the scarcity and variability of demonstrations. Moreover, they must undergo rigorous safety certification to comply with stringent medical regulations and clinical protocols before deployment in real-world settings.

Field of view (FoV) limitations represent a fundamental barrier to robotic-assisted surgery (RAS) automation \cite{gketsis2006laparoscopic, venkatayogi2023impaired}. These limitations impede both large-scale data collection for training generalist neural motion planners and accurate collision checking during deployment. Clinical evidence demonstrates that surgeons operate under suboptimal visual conditions for approximately 40\% of minimally invasive procedures, contributing to nearly 20\% of surgical complications \cite{dhingra2025clear}, while impaired vision affects 58\% of robotic surgery duration \cite{venkatayogi2023impaired}. These FoV constraints stem from multiple sources: trocar diameter restrictions limiting camera dimensions, rapid scene changes from camera movements, lighting fluctuations, instrument occlusion, and organ deformations \cite{kennedy2020computer, su2022mri}.

Recently, specialized frameworks have been developed for specific interventions with critical FoV challenges. SafeRPlan \cite{ao2025saferplan} implements Safe Deep Reinforcement Learning for robotic spine surgery, integrating neural networks pre-trained on preoperative images with uncertainty-aware safety filters to encode anatomical knowledge. For laparoscopic procedures, Iyama et al. \cite{iyama2025autonomous} developed a neural network combining point cloud CNN with DRL for autonomous forceps positioning, directly optimizing tissue surface planarity and visibility to address FoV limitations. In neurosurgery, Segato et al. \cite{segato2021inverse} introduced an Inverse Reinforcement Learning framework for steerable needle navigation in keyhole procedures, learning from expert demonstrations to adapt to dynamic tissue deformations in real-time, achieving exceptional precision with sub-millimeter targeting errors and 0.02-second re-planning capabilities while maintaining 100\% success rates under challenging deformable conditions.

Recent advances in neural networks have demonstrated significant progress in addressing FoV limitations. Neural radiance fields (NeRF)-based approaches have emerged as particularly promising solutions. Qin et al. \cite{qin2024neural} developed a NeRF-driven network that reconstructs highly realistic endoscopic scenes from multi-view images. To address the impracticality of extensive multi-view image collection during surgery, Neri et al. \cite{neri2025surgical} proposed a method enabling NeRF reconstruction from single intraoperative images combined with preoperative data, employing neural style transfer for efficient alignment and training. These approaches have the potential to be integrated into the neural motion planner's pipeline for efficient training and reliable deployment.

\subsection{Re-manufacturing}
Re-manufacturing involves full disassembly and reassembly of end-of-life (EOL) products to enable their sustainable recovery \cite{wang2021energy, li2020unfastening, feng2018flexible}. Human-robot collaborative (HRC) disassembly combines the robot’s strength and the human's dexterity to enable flexible and efficient disassembly of EOL products \cite{peng2025dynamic, guo2023human, qin2024multiple}. 

\begin{figure}[!htbp]
\centering
\includegraphics[width=0.5\textwidth]{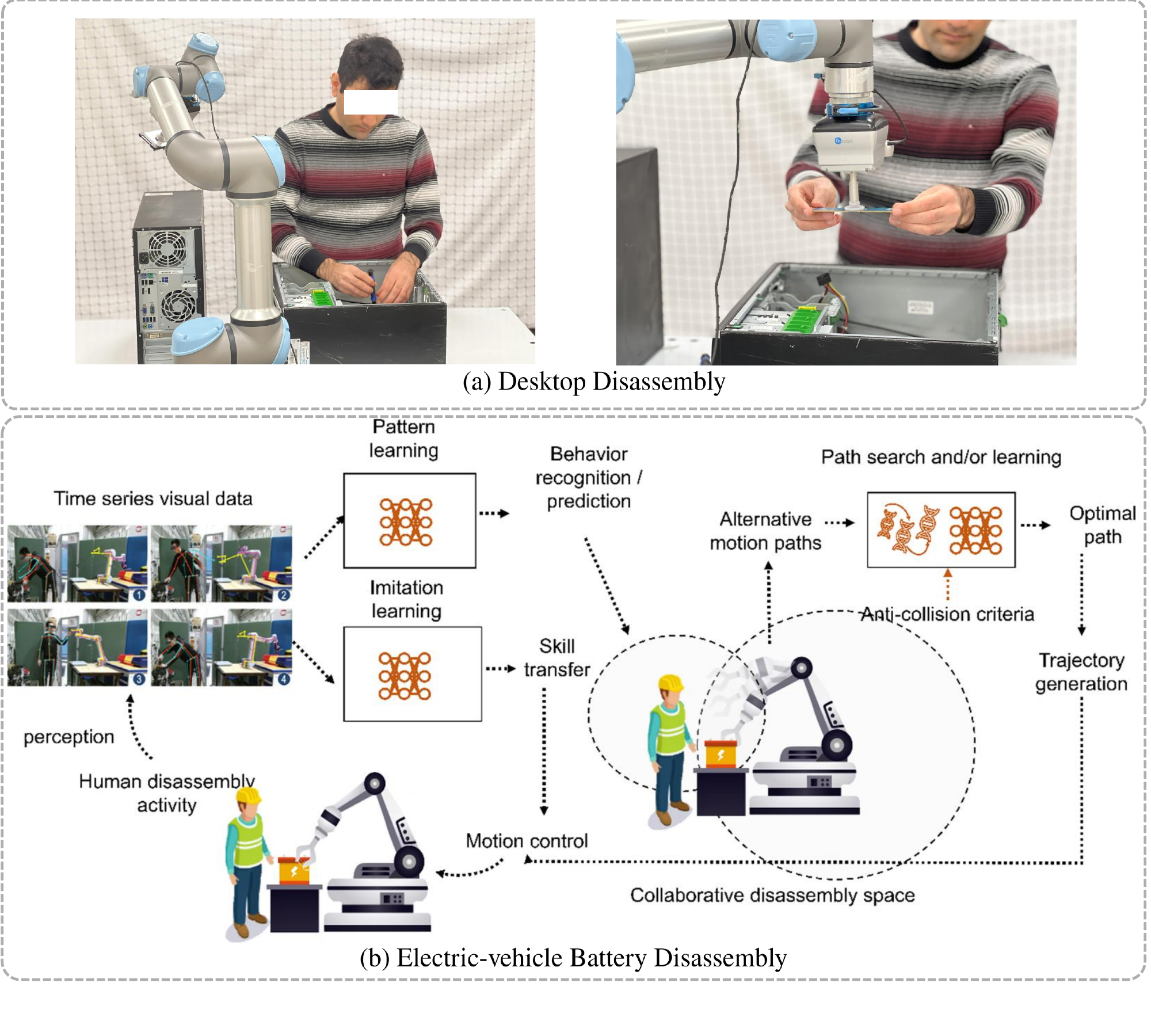}
\caption{Human-robot intelligent collaboration for disassembly. (a) Desktop disassembly. (b) Electric-vehicle battery disassembly \cite{meng2022intelligent}.}
\label{fig_14- remanufacturing}
\end{figure}

HRC disassembly requires dynamic and collision-free collaboration, which relies on efficient human motion prediction and reliable motion planning for the robotic manipulator \cite{lee2024review}. As demonstrated in Figure \ref{fig_14- remanufacturing}, ensuring collision-free collaboration is challenging due to the proximity of human operators and the complexity of the disassembly environment. Deep learning methods, have the potential to enhance the adaptability and flexibility of HRC disassembly by offering fast inference, ease of implementation, and strong generalization capabilities \cite{meng2022intelligent}.

Deep learning methods have enhanced the HRC disassembly process by facilitating reliable human motion prediction \cite{zhang2020recurrent} and efficient motion planning for robotic manipulators \cite{lee2024review}. Regarding human motion prediction, sequence modeling network architectures such as recurrent neural networks (RNNs) have been used to capture temporal patterns in human motion for efficient prediction \cite{zhang2020recurrent}. Regarding reliable motion planning, various deep learning methods have been utilized for collision-free and precise manipulator motion planning \cite{soleymanzadeh2025simpnet}. More specifically, constraint-aware neural motion planners, combined with safety filters, have the potential to generate optimal, collision-free, and safe trajectories for robotic manipulators in HRC disassembly settings. These planners are designed to efficiently handle planning constraints \cite{qureshi2020neural}, which can be deployed to handle disassembly constraints, such as managing fragile or hazardous components, and facilitate fast planning in dynamic environments. Predicted human motion could also be incorporated as an additional constraint to ensure safe collaboration. Integrating safety filters \cite{hsu2021safety} with these planners prevents collision with the human operator and guarantees safety. Validating the disassembly process in a high-fidelity digital twin further ensures safe and efficient planning execution within HRC scenarios \cite{nikolakis2019cyber}.

Large-scale re-manufacturing still relies on manual labor, while intelligent disassembly remains limited to laboratory environments \cite{lee2024review}. Deploying neural motion planners in HRC disassembly settings involves several challenges. Firstly, large-scale datasets and extensive training are required. However, EOL products lack standardized designs, and exhibit high structural variability, and have uncertain physical conditions, which complicates large-scale data collection. Moreover, task- and operator-specific safety filters are essential for safe HRC disassembly. However, the complexity of workspaces and variability among human operators make it difficult to design and integrate such safety filters into neural motion planners.

\subsection{Other Domains}
\vspace{0.2cm}
\noindent
\textbf{Manufacturing:} The deployment of robotic manipulators in existing manufacturing environments is predominantly optimized for rigid processes \cite{yu2020robotic, zhong2023real, li2023using, yang2023heuristics}. Deep learning and neural motion planners have huge potential for flexible manufacturing, which offers greater adaptability for various assembly tasks and products. However, their widespread adoption is hindered by the requirement for large-scale, real-world data collection and the stringent safety-certification procedures necessary for human-robot collaboration. Moreover, the ability to generalize learned motion policies to novel, unseen manufacturing tasks remains an open challenge.

\vspace{0.2cm}
\noindent
\textbf{Agriculture:} Precision agriculture still relies heavily on manual labor \cite{wakchaure2023application}. Deploying robotic manipulators for agriculture automation requires adaptability to environmental variations, coordination under limited connectivity, and the ability for crop-specific harvesting. Deep learning and neural motion planners offer the potential to provide end-to-end policy networks for effective harvesting in cluttered environments. However, guaranteeing the robustness of these methods is challenging due to the less-structured nature of agricultural environments.

\vspace{0.2cm}
\noindent
\textbf{Construction:} Construction environments are typically unstructured and constantly changing \cite{zada2022structure}. Deploying robotic manipulators in such environments necessitates real-time replanning and strict adherence to safety regulations. Neural motion planners can enable robotic manipulators to autonomously adapt to dynamic conditions, navigate cluttered workspaces, avoid obstacles such as scaffolding, and coordinate with other agents such as humans and cranes. However, their limitations include the scarcity of labeled on-site data, the need for real-time domain adaptation as conditions evolve, and the difficulty of certifying safety when heavy loads or human workers are involved.

\vspace{0.2cm}
\noindent
\textbf{Warehouse}: Warehouse robotics is essential for logistics and working alongside human operators \cite{prakash2020dual}. Current warehouses typically utilize conveyors and sorting machines to handle various packages. Deploying robotic manipulators within warehouses requires high mobility, manipulability, and effective human collaboration. Moreover, they often need to operate in highly constrained, confined spaces. To address these challenges, neural motion planners have the potential to generate real-time, collision-free paths to enable fast pick-and-place operations within these environments. However, these planners need to explicitly consider kinodynamic constraints and smoothness for effective deployment. Ensuring operational safety and collecting large-scale, domain-specific data are also critical for the widespread deployment of warehouse robotics.

The deployment of neural motion planners in real-world environments remains limited. Most existing studies demonstrate these methods primarily in simulation or controlled laboratory settings, with practical validation and widespread real-world adoption still underdeveloped.

\section{Conclusions} \label{sec: conclusion}
Through a comprehensive examination of state of the art, we analyzed how various deep learning architectures have improved classical motion planning algorithms for robotic manipulators. In this examination:

\begin{itemize}
    \item We delved into classical planning algorithms to identify their core components, and we discussed how various deep learning characteristics, such as fast inference, inductive biases, parallelization, and multi-modal feature encoding capabilities, have improved these components, and provided a systematic map from various deep learning frameworks to specific algorithmic primitives of classical motion planning algorithms.

    \item We also outlined the essential considerations towards developing generalist neural motion planners capable of end-to-end planning and robust deployment within unstructured real-world environments.
\end{itemize}

In addition to highlighting the improvements that deep learning methods lend to motion planning algorithms, we also identified challenges and considerations that need to be addressed before these frameworks can be safely deployed within a broad range of unstructured real-world environments. Particularly, we emphasized the need for standardized benchmarks, large-scale planning datasets, explicit handling of safety constraints, generalization to out-of-distribution scenarios, and robustness to planning uncertainties for reliable deployment within unstructured real-world environments. Additionally, we discussed and emphasized how recent large-scale foundation models can be established and leveraged to facilitate reaching this goal.

This review aims to serve as a foundational resource for researchers interested in exploring deep learning applications in motion planning for robotic manipulators.

\bibliographystyle{IEEEtran}
\bibliography{Ref/refs}

\end{document}